\documentclass{article} % For LaTeX2e
\usepackage{iclr2026_conference,times}

\usepackage{amsmath,amsfonts,bm}

\def\eqref#1{equation~\ref{#1}}
\def\1{\bm{1}}

\DeclareMathAlphabet{\mathsfit}{\encodingdefault}{\sfdefault}{m}{sl}
\SetMathAlphabet{\mathsfit}{bold}{\encodingdefault}{\sfdefault}{bx}{n}

\usepackage{hyperref}
\usepackage{url}
\usepackage{color}
\usepackage{colortbl}
\usepackage{caption}
\usepackage{array}
\usepackage{multirow}
\usepackage{booktabs}
\usepackage{pifont}
\usepackage{amssymb}
\usepackage{amsmath}
\usepackage{bm}
\usepackage{tikz}
\definecolor{Gray}{gray}{0.93}
\usepackage{listings}
\usepackage{algorithm}
\usepackage{algorithmic}
\usepackage{enumitem}
\usepackage{tcolorbox}
\tcbuselibrary{listings}
\newlength\savewidth\newcommand\shline{\noalign{\global\savewidth\arrayrulewidth
  \global\arrayrulewidth 1pt}\hline\noalign{\global\arrayrulewidth\savewidth}}
\definecolor{deemph}{gray}{0.62}
\definecolor{mgreen}{rgb}{0.19,0.80,0.19}
\newcommand{\gc}[1]{\textcolor{deemph}{#1}}
\newcommand{\veritas}{{\textbf{\textsc{Veritas}}}}
\newcommand{\veritasmini}{{\textbf{\textsc{Veritas-mini}}}}
\definecolor{selfblue}{RGB}{65,105,225} %{89,138,234}
\definecolor{lightblue}{RGB}{220,230,255}
\usepackage{fontawesome5}

\title{\textsc{Veritas}: Generalizable Deepfake Detection via Pattern-Aware Reasoning}

\author{Hao Tan$^{1,2,3}$\thanks{This work was done during the first author’s internship at Ant Group.} \quad
Jun Lan$^{3\dagger}$ \quad
Zichang Tan$^{4}$ \quad 
Senyuan Shi$^{2}$ \quad 
Ajian Liu$^{2}$ \quad \\
\textbf{Chuanbiao Song}$^{3}$ \quad
\textbf{Huijia Zhu}$^{3}$ \quad
\textbf{Weiqiang Wang}$^{3}$ \quad
\textbf{Jun Wan}$^{1,2,5\S}$ \quad
\textbf{Zhen Lei}$^{1,2,5}$ \quad
\\
$^{1}$School of Advanced Interdisciplinary Sciences (SAIS), University of Chinese Academy of Sciences\\
$^{2}$MAIS, Institute of Automation, Chinese Academy of Sciences \quad 
$^{3}$Ant Group \quad \\
$^{4}$Shenzhen Institute of Advanced Technology (SIAT), Chinese Academy of Sciences \\
$^{5}$School of Artificial Intelligence, University of Chinese Academy of Sciences \\
\texttt{\{tanhao2023, jun.wan, zhen.lei\}@ia.ac.cn} \quad
\texttt{yelan.lj@antgroup.com} \\
{\footnotesize $^\dagger$Project Lead. \, $^\S$Corresponding author. \, \faGithub\ Project Page: \href{https://github.com/EricTan7/Veritas}{https://github.com/EricTan7/Veritas}}
}

\iclrfinalcopy % Uncomment for camera-ready version, but NOT for submission.
\begin{document}

\maketitle

\begin{abstract}
Deepfake detection remains a formidable challenge due to the evolving nature of fake content in real-world scenarios.
However, existing benchmarks suffer from severe discrepancies from industrial practice, typically featuring homogeneous training sources and low-quality testing images, which hinder the practical usage of current detectors.
To mitigate this gap, we introduce \textbf{HydraFake}, a dataset that contains diversified deepfake techniques and in-the-wild forgeries, along with rigorous training and evaluation protocol, covering unseen model architectures, emerging forgery techniques and novel data domains.
Building on this resource, we propose \veritas, a multi-modal large language model (MLLM) based deepfake detector.
Different from vanilla chain-of-thought (CoT), we introduce \textit{pattern-aware reasoning} that involves critical patterns such as ``planning'' and ``self-reflection'' to emulate human forensic process.
We further propose a two-stage training pipeline to seamlessly internalize such deepfake reasoning capacities into current MLLMs.
Experiments on HydraFake dataset reveal that although previous detectors show great generalization on cross-model scenarios, they fall short on unseen forgeries and data domains.
Our \veritas $\,$ achieves significant gains across different out-of-domain (OOD) scenarios, and is capable of delivering transparent and faithful detection outputs.
\end{abstract}

\section{Introduction}
\label{sec:intro}

Recent advances in Generative AI~\citep{esser2024scaling, tian2024visual} have revolutionized our digital life, 
unprecedentedly enriching the diversity of content on social media and short-video platforms.
Though bringing immense creativity, such techniques also enable highly convincing deepfakes with minimal cost, posing significant security risks to society.
Consequently, Deepfake Detection (DFD), which aims at discerning between real and generated facial images, has become a heated research frontier, galvanizing extensive efforts.

However, current detectors mostly follow a standard evaluation, which involves training on one dataset~\citep{rossler2019faceforensics++} and testing on others~\citep{dolhansky2019deepfake, li2020celeb, dolhansky2020deepfake, zi2020wilddeepfake, zhou2021face}.
Despite its popularity, this protocol fails to align with practical industrial scenarios, where abundant training samples are available yet significant out-of-distribution (OOD) generalization challenges (e.g., brand-new forgery types and meticulously synthesized facial images) emerge during testing.
Such \textit{\textbf{discrepancy}} severely hinders the practical deployment of current detectors.
To mitigate the gap, we construct \textbf{HydraFake} dataset.
As shown in Figure~\ref{fig:data}, we systematically collect and reproduce advanced deepfake methods, covering diversified deepfake techniques and in-the-wild forgeries from social media.
To simulate potential challenges in real-world scenarios, we establish a rigorous and holistic evaluation protocol, where the training set consists of abundant samples but is restricted to three basic forgery types, and the evaluation involves hierarchical OOD testing, spanning in-domain, cross-model, cross-forgery and cross-domain scenarios, enabling fine-grained understanding of the model's capacities.
As presented in Figure~\ref{fig:data} (d), under such rigorous evaluation, current SOTA detectors show great generalization on cross-model deepfakes, but limited abilities in cross-forgery and cross-domain scenarios.

To improve the robustness on unseen forgeries and data domains, we seek to ground the generalization abilities of multi-modal large language models (MLLMs) into deepfake detection.
Recent efforts~\citep{huang2024ffaa, guo2025rethinking, peng2025mllm} have made initial attempts, while they focus on the explainability and the classification is still based on expert vision models.
In constrast, we explore to seamlessly internalize MLLMs into deepfake detection through their intrinsic reasoning abilities.
However, directly applying deep reasoning faces a critical challenge:
current MLLMs are extremely short for deepfake detection~\citep{ren2025can, tariq2025llms}.
Effective reasoning data is necessary to ground the abilities of base model.
To achieve this goal, we must answer two key questions:
\textbf{(1)} what kind of reasoning process is helpful to DFD task?
and \textbf{(2)} with sufficient data, how can we ensure the model is learning to reason for DFD rather than memorizing?

\begin{figure*}[t]
    \centering
    \includegraphics[width=0.99\linewidth]{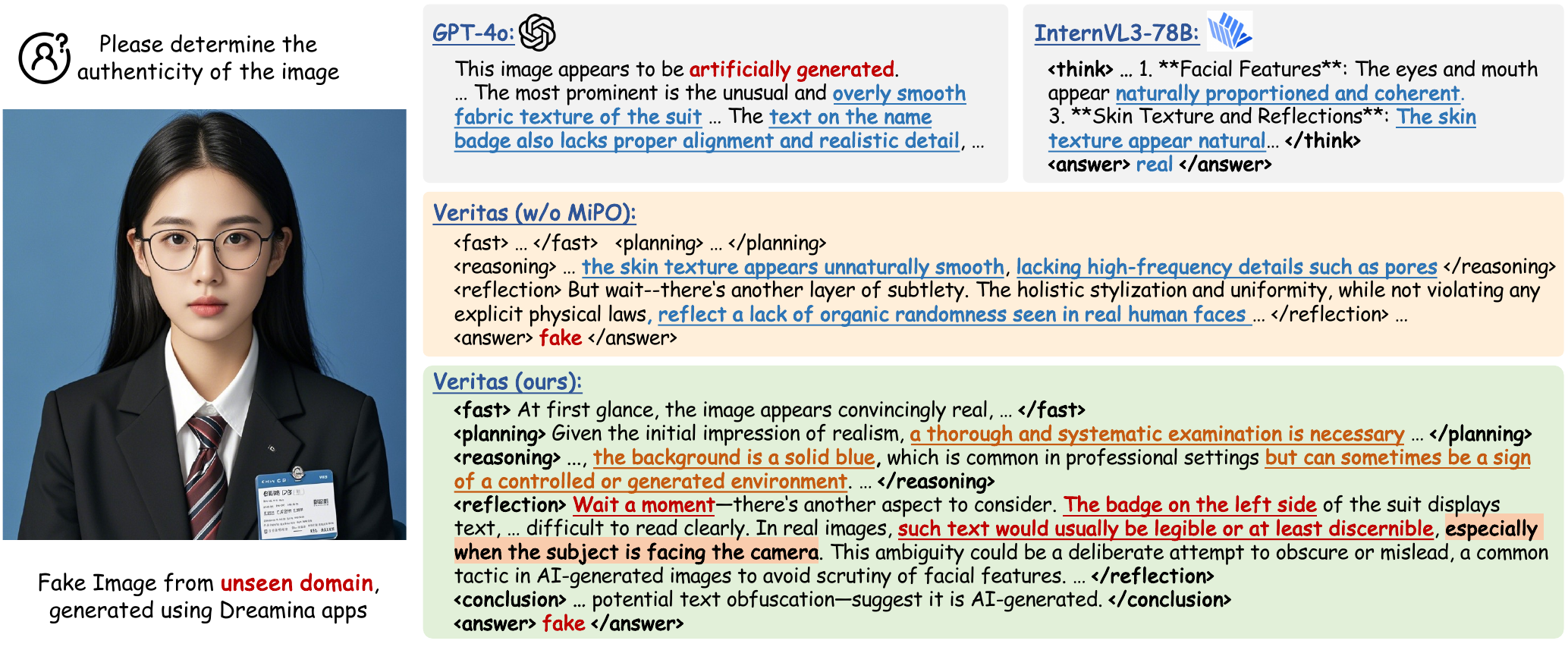}
    \vspace{-5pt}
     \caption{Comparison of the detection outputs. InternVL3-78B~\citep{zhu2025internvl3} gets incorrect answer. GPT-4o~\citep{hurst2024gpt} and our model trained without the proposed MiPO both fail to provide precise explanation.
     In contrast, our model gives transparent and faithful decision process.
     }
	\label{fig:reason_compare}
    % \vspace{-0.6cm}
\end{figure*}

\textbf{For the first question}, we introduce a pattern-aware reasoning framework.
Drawing inspiration from recent studies~\citep{zhao2025echo, muennighoff2025s1} that demonstrate critical \textit{reasoning patterns} greatly elevate the OOD performance of LLMs, we consider the human mindset for deepfake detection:
when determining the authenticity of an image, we tend to make a quick judgment based on our first impression (\textit{fast judgement}), then identify one or two prominent features (\textit{reasoning}) to draw a conclusion (\textit{conclusion}).
For more challenging samples, we may conduct a layered analysis (\textit{planning}), and may also engage in more in-depth thinking to support or overturn our initial judgement (\textit{self-reflection}).
Based on this analogy, we extract these five thinking patterns to facilitate logical and holistic reasoning.
Table~\ref{tab:pattern_mode_abl} empirically shows the benefits of such pattern-aware reasoning over vanilla Chain-of-Thought (CoT).
\textbf{For the second question}, we introduce a two-stage training pipeline consisting of pattern-guided cold-start and pattern-aware exploration, yielding our \veritas $\,$\footnote{\veritas$\,$ means ``Truth'' in Latin.}model.
During cold-start, we employ SFT to internalize thinking patterns.
Besides, we introduce a Mixed Preference Optimization (MiPO) strategy that leverages mixed non-preference data and human-annotated preference data to steer the model toward faithful and fine-grained reasoning.
As shown in Figure~\ref{fig:reason_compare}, MiPO greatly improves the reasoning quality, mitigating the memorizing behavior.
To further facilitate adaptive planning and self-reflection, we propose Pattern-aware Group Relative Policy Optimization (P-GRPO), which shapes reasoning behavior through online sampling and pattern-aware reward mechanism.
As a result, \veritas $\,$ shows great generalization on unseen forgeries and data domains, providing transparent and precise decision process (Figure~\ref{fig:reason_compare}).

To sum up, our main contributions are:
\begin{itemize}
    \item \textbf{Dataset}: We introduce \textbf{HydraFake}, a dataset that simulates real-world challenges with hierarchical generalization testing, advancing the evaluation protocol in deepfake detection and helping developers better locate the deficiencies of their detectors.
    \item \textbf{Method}: We propose a two-stage training pipeline that grounds the capabilities of MLLMs into deepfake detection through pattern-aware reasoning.
    Our model supports adaptive planning and self-reflection, delivering transparent and human-aligned decision end-to-end.
    \item \textbf{Performance}: Our \veritas $\,$ model achieves significant improvements over state-of-the-art detectors on cross-forgery and cross-domain scenarios, and our cold-start model serves as a strong reasoning foundation for further customization.
\end{itemize}

\section{Related work}
\subsection{Deepfake Detection and Datasets}
Deepfake detection aims to distinguish generated facial images from authentic human faces.
Previous efforts have explored spatial-level~\citep{ojha2023towards, yan2024transcending, tan2024rethinking, nguyen2024laa,fu2025exploring, yan2024sanity, yan2024orthogonal, yang2025all}, frequency-level~\citep{qian2020thinking, tan2024frequency, zhou2024freqblender, kashiani2025freqdebias} and sequence-level~\citep{gu2021spatiotemporal, gu2022hierarchical, gu2022delving, yan2025generalizing} approaches, achieving remarkable progress on traditional benchmarks.
To train a generalizable detector, some methods attempt to find ``bias-free'' fake images either through spatial-domain blending~\citep{li2020face, shiohara2022detecting, zhao2021learning}, frequency-domain blending~\citep{zhou2024freqblender, kashiani2025freqdebias} or feature-level augmentation~\citep{yan2024transcending}.
However, the commonly adopted protocol, i.e., training on FF++~\citep{rossler2019faceforensics++} and testing on others~\citep{dolhansky2019deepfake, li2020celeb, dolhansky2020deepfake, zi2020wilddeepfake, zhou2021face}, suffers from two problems:
(1) the training sources are overly narrow, and (2) the testing data exhibit limited forgery types and low-resolution.
Although many timely datasets~\citep{yan2024sanity, zhang2024bench, li2025artificial, huang2025so, wang2025dfbench, wen2025busterx, xia2025mirage} have been proposed for AIGC detection, the pace of deepfake detection has lagged behind.
As a result, previous methods are biased towards such settings, exhibiting degraded generalization when learning from varying sources or mixed artifacts.
To mitigate this problem, we introduce a hierarchical protocol in our HydraFake dataset, aiming to comprehensively reflect the generalization capability of the detectors.

\subsection{MLLMs for Deepfake Detection}
With the proliferation of MLLMs~\citep{liu2023visual, bai2025qwen2, zhu2025internvl3}, recent focus has shifted to explainable deepfake and AIGC detection.
However, most methods still rely on small vision models for the final decision.
For instance, M2F2-Det~\citep{guo2025rethinking} determines the authenticity purely based on CLIP models, where LLM is leveraged as a plug-in interpreter.
Similarly, DD-VQA~\citep{zhang2024common}, FFAA~\citep{huang2024ffaa} and VLF-FFD~\citep{peng2025mllm} develop post-processing system to aggregate embeddings from small vision models.
Some methods~\citep{he2025vlforgery, sun2025towards, chen2025mgffd} attempt to directly adopt the outputs from LLMs, e.g., Sun et al.~\citep{sun2025towards} construct precise forgery explanations to release the power of MLLMs.
Recent methods~\citep{huang2025sida, xu2024fakeshield, zhou2025aigi} also adopt MLLMs and curated datasets for AIGC detection.
However, these methods generate post-hoc explanations by first determining the answer.
The potential of reasoning abilities for deepfake detection is still underexplored.
The most recent methods~\citep{gao2025fakereasoning, xia2025mirage} explore the reasoning for AIGC detection, while neglecting adaptive reasoning patterns and is not tailored for facial forgery.
Different from previous methods, we introduce human-like reasoning into deepfake detection, achieving promising improvements and delivering transparent decisions end-to-end.

\begin{figure*}[ht]
    \centering
    \includegraphics[width=0.99\linewidth]{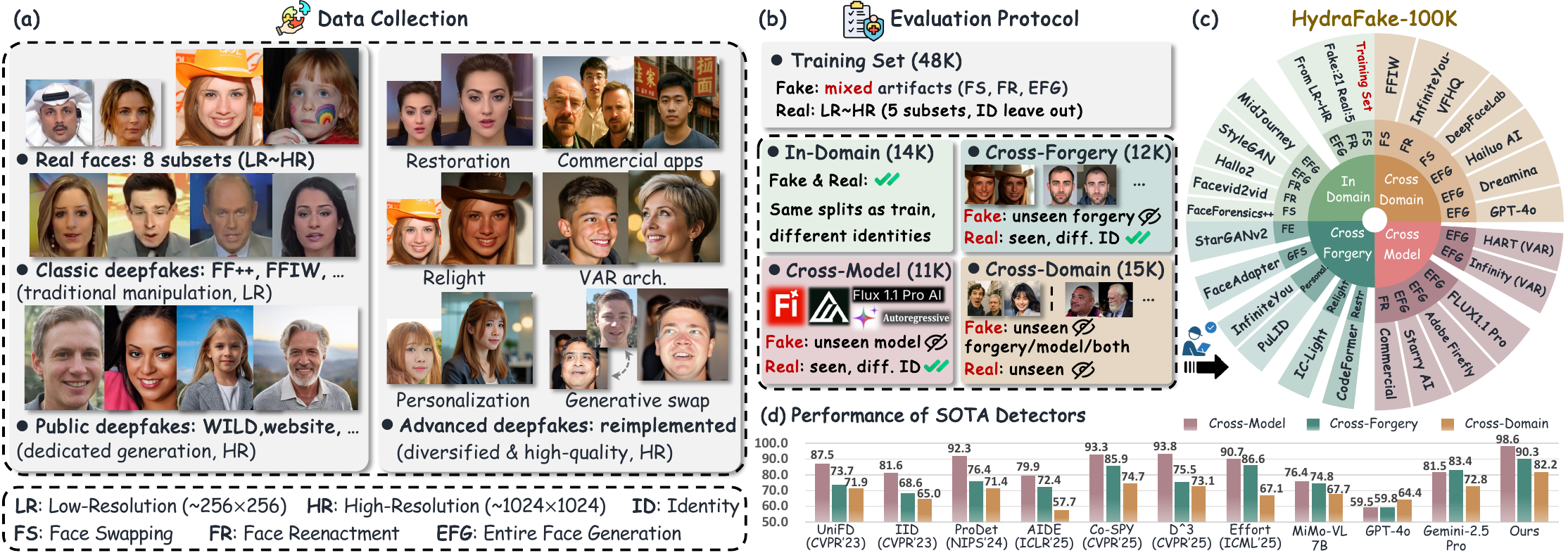}
     \caption{\textbf{Overview of HydraFake dataset}.
     \textbf{(a)} We carefully collect and reimplement advanced deepfake techniques to construct our HydraFake dataset.
     Real images are collected from $8$ datasets.
     Fake images are from classic datasets, high-quality public datasets and our self-constructed deepfake data.
     \textbf{(b)} We introduce a rigorous evaluation protocol.
     Training data contains abundant samples but limited forgery types.
     Evaluations are split into four levels.
     \textbf{(c)} Illustration of the subsets in different evaluation splits.
     \textbf{(d)} Performance of existing detectors.
     Most detectors generalize well on Cross-Model setting but perform poorly on Cross-Forgery and Cross-Domain scenarios.
     }
	\label{fig:data}
    \vspace{-0.4cm}
\end{figure*}

\section{HydraFake Dataset}
In this part, we introduce our HydraFake dataset, including the construction process and evaluation protocol.
Detailed statistics and information are provided in Appendix~\ref{supp:dataset}.

\subsection{Data Collection}
\noindent \textbf{Real Images.} As shown in Figure~\ref{fig:data} (a), the real images are collected from $8$ public datasets, containing both low-resolution (i.e., LFW~\citep{huang2008labeled}, CelebA~\citep{liu2015deep}, FaceForensics++ (FF++)~\citep{rossler2019faceforensics++}, FFIW~\citep{dolhansky2019deepfake}) and high-resolution images (i.e., FFHQ~\citep{karras2019style}, VFHQ~\citep{xie2022vfhq}, UADFV~\citep{yang2019exposing} and CelebAHQ~\citep{karras2017progressive}).
The collected images are rigorously partitioned for training and testing.

\noindent \textbf{Fake Images.} The fake images come from three sources:
\begin{itemize}[left=2pt]
    \item \noindent \textbf{Classic deepfake} data sampled from FF++~\citep{rossler2019faceforensics++} DF40~\citep{yan2024df40} and FFIW~\citep{dolhansky2019deepfake}, which mainly contain face swapping (FS) and face reenactment (FR) forgeries from $10$ generative models.
    The artifacts are mostly localized.
    \item \textbf{Public deepfake} data sampled from WILD~\citep{bongini2025wild}, seeprettyface website and TalkingHeadBench~\citep{xiong2025talkingheadbench}. This contains carefully synthesized faces from $16$ popular generators.
    However, there still exist corner cases such as fresh forgery types.
    \item \textbf{Advanced deepfake} data where we further reimplemented and crawled $10$K deepfake data from $10$ advanced generators.
    Besides traditional deepfake techniques, HydraFake dataset contains Face Restoration~\citep{zhou2022towards}, Face Relighting~\citep{zhang2025scaling}, Face Personalization~\citep{jiang2025infiniteyou,guo2024pulid}, Generative Face Swapping~\citep{han2024face} and deepfakes from Visual AutoRegressive models (VAR)~\citep{han2025infinity, tang2024hart}.
    To simulate real-world challenges, we also crawled $1$K deepfake images from social media, which include practical deepfakes generated from commercial apps, including GPT-4o~\citep{hurst2024gpt}, Dreamina~\citep{dreamina} and Hailuo AI~\citep{hailuo}.
\end{itemize}

\noindent \textbf{Quality Control.}
For classic deepfake datasets, we only select FF++~\citep{rossler2019faceforensics++} and FFIW~\citep{zhou2021face}, while not involving DFDC~\citep{dolhansky2020deepfake}, DFDCP~\citep{dolhansky2019deepfake} and WDF~\citep{zi2020wilddeepfake} due to their low quality (e.g., unexpected blurring in real images).
For our self-constructed deepfake data, we conduct strict quality control, e.g.,
for face personalization, we use Qwen2.5-VL-72B to tailor sample-specific prompts rather than using template-like prompts as in~\citep{bongini2025wild}.
For face relighting, we generate multiple lighting sources for each identity and manually select high-quality samples.
After filtering and balancing, our HydraFake dataset contains $50$K real images and $50$K fake images.

\subsection{Evaluation Protocol}
\noindent \textbf{Training.} As shown in Figure~\ref{fig:data} (b), the training set contains $48$K images.
Real images are from $5$ subsets, with other $3$ subsets left out for testing.
Fake images involves $21$ subsets while only contains $3$ forgery types (i.e., FS, FR and EFG).
This is to simulate practical setting, where abundant training images are available but various forgery types and generative models remain unseen.

\noindent \textbf{Evaluation.} 
The evaluation is divided into four distinct levels:
\begin{itemize}[left=2pt]
    \item \textbf{In-Domain} (14K): testing images share the training data source but with different identities.
    \item \textbf{Cross-Model} (11K): fake images are generated by unseen models under controlled conditions like the template-based textual prompts. 
    This includes SOTA models from recent years (e.g., FLUX1.1-Pro~\citep{FLUX}, Adobe FireFly~\citep{firefly}, Starry AI~\citep{starryai}), distinct model architectures (e.g., VAR~\citep{han2025infinity, tang2024hart} and Video AR model~\citep{magi}).
    The real images are from in-domain set but with different identities.
    \item \textbf{Cross-Forgery} (12K): fake images are generated by unseen manipulation techniques, involving attribute editing, generative face swapping, IP-preserved personalization, face relighting and face restoration.
    The real images are from in-domain set but with different identities.
    This split is to evaluate the model's capacity to detect fake images generated by unseen manipulation.
    \item \textbf{Cross-Domain} (15K): 
    fake images are either generated under controlled conditions or collected from the web, including both unseen forgeries and unseen models.
    The real images are from unseen datasets (i.e., VFHQ~\citep{xie2022vfhq}, UADFV~\citep{yang2019exposing} and FFIW~\citep{dolhansky2019deepfake}).
    The images are of different qualities, posing strong challenges.
\end{itemize}

\section{Method}
\begin{figure*}[t]
    \centering
    \includegraphics[width=0.93\linewidth]{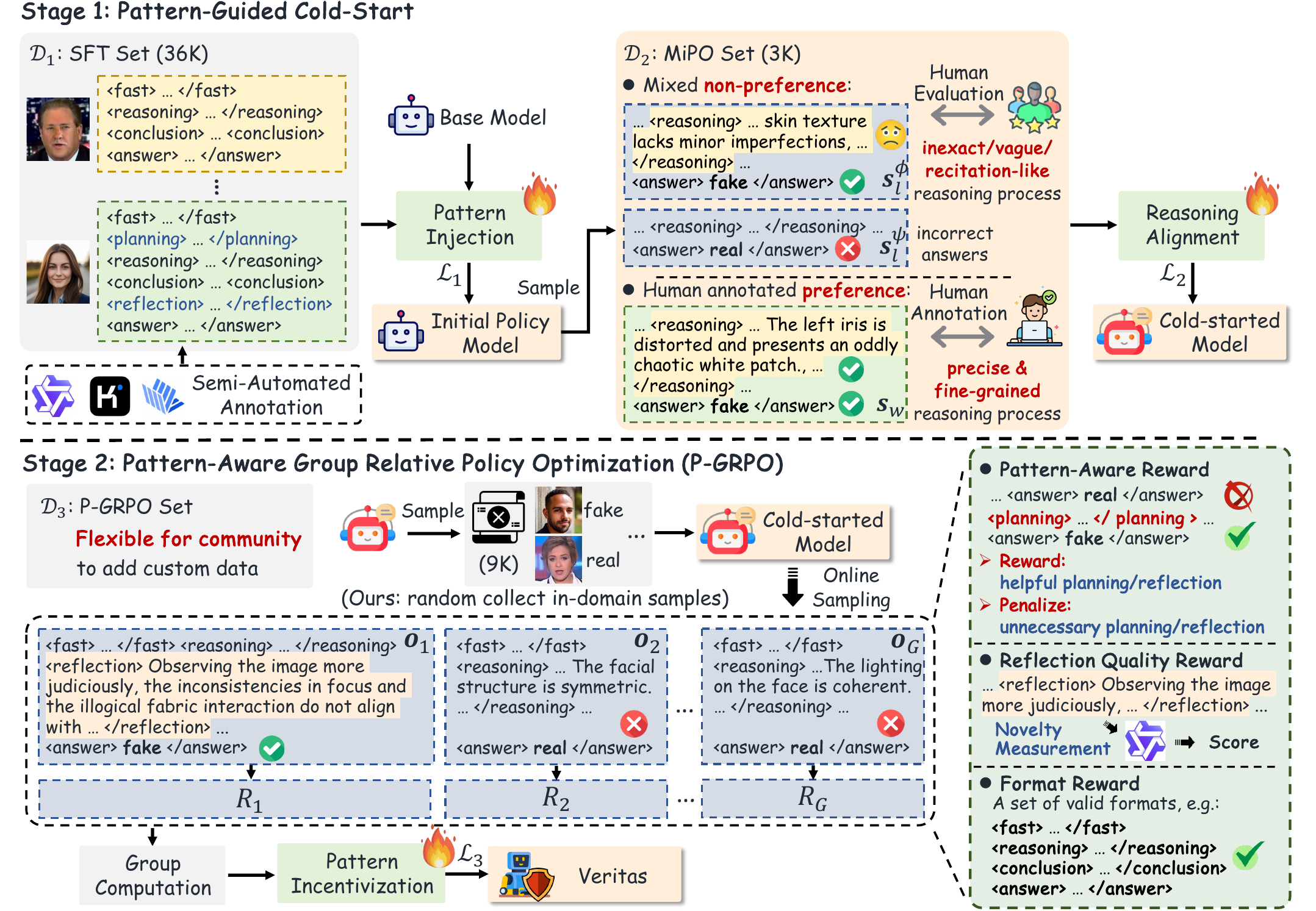}
     \caption{\textbf{Overview of two-stage training pipeline}.
     \textbf{(a)} For Pattern-Guided Cold-Start, we first employ SFT to internalize thinking patterns. Then we introduce MiPO to facilitate human-aligned reasoning. The MiPO dataset consists of mixed non-preference data, encouraging model to perform precise and fine-grained reasoning.
     \textbf{(b)} For Pattern-Aware GRPO, we introduce pattern-aware reward to incentivize adaptive reasoning ability on pattern granularity, yielding our \veritas $\,$ model.
     }
	\label{fig:method}
    \vspace{-0.4cm}
\end{figure*}

In this section, we detail the two-stage training pipeline of \veritas, including pattern-guided cold-start and pattern-aware reinforcement learning, as shown in Figure~\ref{fig:method}.

\subsection{Pattern-guided Cold-Start}
To internalize thinking patterns for deepfake detection, we first employ a pattern-guided cold-start.
Different from common practice, we involve two steps: Supervised Fine-Tuning (SFT) for format injection,
and a Mixed Preference Optimization (MiPO) strategy to align the reasoning process.

\textbf{SFT Pattern Injection.}
Suppose the SFT dataset is denoted as $\mathcal{D}_1=\{(\bm{q}, \bm{s})_i\}_{i=1}^{N_1}$, where $\bm{s}$ is the target output sequence including pattern-aware reasoning and final answer. $\bm{q}$ denotes input image and user query.
The training objective maximizes the likelihood of generating $\bm{s}$ given input $\bm{q}$:
\begin{equation}
    \mathcal{L}_{\mathrm{1}}=-\mathbb{E}_{(\bm{q},\bm{s})\thicksim\mathcal{D}_1}\sum_{t=1}^T\log\pi_\theta(\bm{s}_t\mid\bm{q},\bm{s}_{<t}),
\end{equation}
where $\pi_\theta$ denotes the token distribution from the current model.
In the following we introduce the construction process of our training data $\mathcal{D}_1$.

To minimize human costs, we use MLLMs for automated annotation, similar to recent practices~\citep{huang2024ffaa, xu2024fakeshield}.
However, this encounters two challenges in our case:
(1) The MLLMs tend to overlook some subtle artifacts like abnormal optical focusing.
(2) The model prioritizes producing logical paths than to accurately locating artifacts.
To mitigate the two issues, we construct a multi-step annotation pipeline.
We first manually inspect a subset and summarize a comprehensive artifacts taxonomy (Figure~\ref{fig:annotation} (a)):
(1) \textbf{Perceptible} structural anomalies, which are immediately visible and easy to detect.
(2) \textbf{Subtle} low-level artifacts, which require careful inspection.
(3) \textbf{Cognitive} violations of physical laws, which are implicit and require connecting to common sense or real-world knowledge.
Then, we decouple the annotation into three \textit{specialized yet coherent} steps (Figure~\ref{fig:annotation} (b)), resulting in $36$K samples for $\mathcal{D}_1$.
Detailed process and all prompt templates are provided in Appendix~\ref{supp:data_annotate}.
Annotated examples are presented in Figure~\ref{fig:annotation} (c).

\textbf{MiPO Reasoning Alignment.}
To further facilitate human-aligned reasoning, we meticulously curate a mixed preference dataset $\mathcal{D}_2=\{(\bm{q}, \bm{s}_w, \bm{s}_l^{\phi})_i\}_{i=1}^{N_2}\cup\{(\bm{q}, \bm{s}_w, \bm{s}_l^{\psi})_i\}_{i=1}^{N'_2}$.
% with $N'_2\approx3N_2$.
Specifically, we collect two types of non-preference data for the fake images: 
(1) the trajectories where the answer is correct but the reasoning content is not precise or detailed enough (i.e., $\bm{s}_l^{\phi}$).
(2) the trajectories where the answer is incorrect (i.e., $\bm{s}_l^{\psi}$).
$\bm{s}_w$ denotes preferred reasoning traces, which are precisely annotated by our human experts.
Both $\bm{s}_l^{\phi}$ and $\bm{s}_l^{\psi}$ are sampled from the outputs of the SFT model, yielding $3$K high-quality paired samples for dataset $\mathcal{D}_2$.
Note that the images in $\mathcal{D}_2$ strictly come from the in-domain training set, without introducing any OOD samples.
Suppose the SFT model is denoted as $\pi_{\theta_{\text{SFT}}}$, the training objective for MiPO is formulated as:
\begin{equation}
    \mathcal{L}_{\mathrm{2}}=-\mathbb{E}_{(\bm{q},\bm{s}_w,\bm{s}_l)\thicksim\mathcal{D}_2}
    \left[ \log\sigma\left(\beta\log\frac{\pi_{\theta}(\bm{s}_w|\bm{q})}{\pi_{\theta_{\text{SFT}}}(\bm{s}_w|\bm{q})}-\beta\log\frac{\pi_{\theta}(\bm{s}_l|\bm{q})}{\pi_{\theta_{\text{SFT}}}(\bm{s}_l|\bm{q})}
        \right)
    \right],
\end{equation}
where $\sigma(\cdot)$ denotes the sigmoid function and $\beta$ controls the strength that the model deviates from the reference model.
As shown in Figure~\ref{fig:reason_compare}, by learning from such mixed rejected traces, our model can perform more precise and fine-grained reasoning compared to pure SFT cold-start.

\subsection{Pattern-Aware Exploration}
After cold-start, the trained model possesses the fundamental reasoning capacities for deepfake detection.
However, it still fails on more challenging samples.
To mitigate this, we introduce Pattern-Aware GRPO (P-GRPO) to encourage the model to perform comprehensive reasoning and potential self-reflection.
Unlike recent approaches~\citep{tu2025learning, xiao2025fast} that encourage adaptive reasoning through length reward, we suppose that absolute reasoning length is not critical.
Instead, we incentivize appropriate thinking patterns through the pattern-aware reward mechanism.

Suppose the training data for P-GRPO is $\mathcal{D}_3 = \{(\bm{q}, \bm{a})_i \}_{i=1}^{N_3}$, where $\bm{a}$ denotes the binary answer.
We randomly sampled $9$K images from in-domain training set.
For a given query $\bm{q}$, P-GRPO samples $G$ responses $\{o_1, o_2, ..., o_G\}$ using the current policy model $\pi_{\theta_{\text{old}}}$.
The quality of each response $\{R_1, R_2, ..., R_G\}$ is evaluated through reward functions.
Suppose the cold-started model $\pi_{\theta_{\text{cold}}}$ is adopted as reference policy,
the training objective is formulated as:
\begin{equation}
\begin{aligned}
    \mathcal{L}_3\!=\!& -\mathbb{E}_{(\bm{q},\bm{a})\thicksim\mathcal{D}_3, \{\bm{o}_i\}_{i=1}^G\thicksim\pi_{\theta_{\text{old}}}(\cdot|\bm{q}) } \\
    &\frac{1}{\sum_{i=1}^G\!|\bm{o}_i|}\!\sum_{i=1}^G\!\sum_{t=1}^{|\bm{o}_i|}\left[\min\left(r_{i,t}(\theta)A_{i,t}, \text{clip}(r_{i,t}(\theta),1\!-\!\epsilon,1\!+\!\epsilon)A_{i,t}\right)-\beta^{'}\!D_{\mathrm{KL}}\!\left[\pi_\theta\|\pi_{\theta_\mathrm{cold}}\right]\right],
\end{aligned}
\end{equation}
where
\begin{equation}
    r_{i,t}(\theta)=\frac{\pi_\theta(o_{i,t}\mid \bm{I},\bm{o}_{i,<t})}{\pi_{\theta_{\text{old}}}(o_{i,t}\mid \bm{I},\bm{o}_{i,<t})},\quad A_{i,t}=\frac{R_i-\mathrm{mean}(\{R_1,\ldots,R_G\})}{\mathrm{std}(\{R_1,\ldots,R_G\})}.
\end{equation}

The reward $R_i$ for each response is evaluated from three perspectives: 

\textbf{Pattern-aware Reward.}
Suppose $\mathcal{C}\in\{0,1\}$ represents the correctness of the final answer, with $\mathcal{C}=1$ denoting the answer is right.
$\mathcal{P}\in\{0,1\}$ and $\mathcal{R}\in\{0,1\}$ represents whether the reasoning involves ``planning'' and ``self-reflection'', respectively.
The pattern-aware reward is defined as:
\begin{equation}
R_{\text{pattern}}=\left\{
\begin{aligned}
    &2.0,\quad \text{if}\ \mathcal{C}\!=\!1\ \wedge (\mathcal{P}\!=\!1\vee \mathcal{R}\!=\!1), \\
    &1.0,\quad \text{if}\ \mathcal{C}\!=\!1\ \wedge \mathcal{P}\!=\!0\wedge \mathcal{R}\!=\!0, \\
    &0.0,\quad \text{if}\ \mathcal{C}\!=\!0\ \wedge \mathcal{P}\!=\!0\wedge \mathcal{R}\!=\!0, \\
    -&0.5,\quad \text{if}\ \mathcal{C}\!=\!0\ \wedge \mathcal{P}\!=\!1\wedge \mathcal{R}\!=\!0, \\
    -&1.0,\quad \text{if}\ \mathcal{C}\!=\!0\ \wedge \mathcal{R}\!=\!1. \\
\end{aligned}
\right.
\end{equation}
Specifically, we encourage the model to reach correct answers through planning and self-reflection by assigning a larger reward (i.e., $2.0$) if they are involved in the reasoning process.
However, if these patterns lead to incorrect answers, we impose a penalty for its overthinking.
Since self-reflection is a more decisive pattern, we assign a larger penalty (i.e., $-1.0$) for errors resulting from it.

\textbf{Reflection Quality and Format Reward.}
To facilitate meaningful self-reflection, we assess the quality of reflection by an external model $\mathcal{M}$: $R_{\text{ref}}=\mathcal{M}(\bm{S})$.
The criterion is the originality of the reflection, i.e., whether it introduces new perspectives rather than restating prior discoveries.
The model only obtains $R_{\text{ref}}$ when the answer is correct.
For format reward $R_{\text{fmt}}$, we predefine some combinations of reasoning patterns and set $R_{\text{fmt}}=1$ when the response conforms to valid formats.

Suppose $\mathbb{I}(\cdot)$ is the indicator function. The final reward $R$ for each response is defined as:
\begin{equation}
    R = R_{\text{pattern}} + \lambda_1R_{\text{ref}}\cdot\mathbb{I}(\mathcal{C}\!=\!1) + \lambda_2R_{\text{fmt}}.
\end{equation}

In practice, given that only verifiable answers are required, the training data $\mathcal{D}_3$ can be freely expanded.
Our cold-start model serves as a solid reasoning foundation, upon which the community can utilize custom data with P-GRPO to achieve more powerful reasoning model for deepfake detection.

\section{Experiments}
\subsection{Experimental Setup}
\noindent \textbf{State-of-the-Art Methods.}
We trained $10$ state-of-the-art (SOTA) detectors on our dataset, including F3Net~\citep{qian2020thinking}, UniFD~\citep{ojha2023towards}, IID~\citep{huang2023implicit}, FreqNet~\citep{tan2024frequency}, ProDet~\citep{cheng2024can}, NPR~\citep{tan2024rethinking}, AIDE~\citep{yan2024sanity}, Co-SPY~\citep{cheng2025co}, D$^3$~\citep{yang2025d}, Effort~\citep{yan2024orthogonal}.
We also assess $4$ open-source MLLMs of similar size to our model, including Qwen2.5-VL-7B~\citep{bai2025qwen2}, InternVL3-8B~\citep{zhu2025internvl3}, MiMo-VL-7B~\citep{coreteam2025mimovltechnicalreport} and GLM-4.1V-9B-Thinking~\citep{hong2025glm},
along with $2$ powerful closed-source models GPT-4o~\citep{hurst2024gpt} and Gemini-2.5-Pro~\citep{comanici2025gemini}.
Besides, we evaluate recent MLLM-based forgery detectors, including FakeShield~\citep{xu2024fakeshield}, M2F2-Det~\citep{guo2025rethinking}, SIDA~\citep{huang2025sida}, FakeVLM~\citep{wen2025spot}, FFAA~\citep{huang2024ffaa}.
More details are in Appendix~\ref{supp:more_imple}.

\noindent \textbf{Metrics.}
Following previous works~\citep{zhang2024common, guo2025rethinking}, we take Accuracy (Acc) to measure the model performance.
Precision and Recall are reported in Appendix~\ref{supp:more_results}.

\noindent \textbf{Implementation Details.}
We implement \veritas $\,$ with InternVL3-8B~\citep{zhu2025internvl3}.
For the cold-start SFT, we train the model for $3$ epochs using LoRA~\citep{hu2022lora} (rank=128, $\alpha$=256).
The learning rate is set to $5\times10^{-5}$, with a batch size of $64$.
For cold-start MiPO, the model is trained for $2$ epochs with the same setting of SFT.
For P-GRPO, we further train the model for $2$ epochs with the same LoRA setting.
The learning rate is set to $1\times10^{-6}$ with a batch size of $16$.
$G$ is set to $4$, with a temperature of $1.0$.
$\beta$ and $\beta'$ are set to $0$.
We take UnifiedReward-Qwen-3B~\citep{wang2025unified} as the reward model $\mathcal{M}$.
For each stage, we directly adopt model from the last step.

\begin{table}[t]
\vspace{-0.5cm}
\caption{Performance comparison (Acc.) on HydraFake dataset. In-domain (ID) results are averaged.
To ensure fair comparisons with MLLM-based detectors, 1) we exclude ID set in their average results and 2) further restrict the training scope of our method to FF++, StyleGAN, StableDiffusion XL and FFHQ (similar to FFAA), yielding ``\veritasmini''.
The best results are \textbf{bolded} and second best are \underline{underlined}.
More metrics in Appendix~\ref{supp:more_results}.}
\vspace{-7pt}
    \centering
    \renewcommand\arraystretch{1.1}
    \scalebox{0.73}{
        \small
        \begin{tabular}{p{77pt}<{\raggedright}p{12pt}<{\centering}p{10pt}<{\centering}p{10pt}<{\centering}p{10pt}<{\centering}p{10pt}<{\centering}p{10pt}<{\centering}p{13pt}<{\centering}p{10pt}<{\centering}p{10pt}<{\centering}p{10pt}<{\centering}p{10pt}<{\centering}p{10pt}<{\centering}p{13pt}<{\centering}p{10pt}<{\centering}p{10pt}<{\centering}p{10pt}<{\centering}p{10pt}<{\centering}p{10pt}<{\centering}p{12pt}<{\centering}p{12pt}<{\centering}}
        \toprule[1pt]
        \multirow{2}{*}{\hspace{-5pt}\vspace{-2.0em}\textbf{Method}} & \multirow{2}{*}{\vspace{-2.0em}\textbf{ID}} & \multicolumn{6}{c}{\textbf{Cross-Model}} & \multicolumn{6}{c}{\textbf{Cross-Forgery}} & \multicolumn{6}{c}{\textbf{Cross-Domain}} & \multirow{2}{*}{\vspace{-2.0em}\textbf{Avg.}} \\
        % \cline{3-20}
        \cmidrule(lr){3-8} \cmidrule(lr){9-14} \cmidrule(lr){15-20}
        & & \rotatebox{60}{ADF} & \rotatebox{60}{FLUX} & \rotatebox{60}{StarryAI} & \rotatebox{60}{MAGI-1} & \rotatebox{60}{HART} & \rotatebox{60}{Infinity} & \rotatebox{60}{St.GAN2} & \rotatebox{60}{ICLight} & \rotatebox{60}{CodeF.} & \rotatebox{60}{InfiniteY.} & \rotatebox{60}{PuLID} & \rotatebox{60}{FaceAda.} & \rotatebox{60}{Deepface.} & \rotatebox{60}{InfiniteY.} & \rotatebox{60}{Dreamina} & \rotatebox{60}{HailuoAI} & \rotatebox{60}{GPT-4o} & \rotatebox{60}{FFIW} & \\
        \shline
        \rowcolor{lightblue}\textit{\textbf{Small Vision Models}} &&&&&&&&&&&&&&&&&&&&\\
        \hspace{-5pt}\rule{0pt}{7pt}F3Net (\textit{ECCV'20}) & 85.3 & 86.7 & 87.8 & 78.6 & 85.0 & 86.0 & 82.9 & 41.3 & 48.9 & 71.9 & 84.9 & 85.5 & 72.6 & 57.7 & 78.5 & 55.6 & 68.6 & 66.2 & 66.4 & 73.2 \\
        \hspace{-5pt}UniFD (\textit{CVPR'23}) & 82.7 & 90.7 & 93.8 & 82.5 & 73.0 & 94.4 & 90.7 & 61.8 & 81.9 & 75.4 & 73.7 & 68.1 & 81.3 & 67.4 & 67.3 & 80.5 & 75.2 & 73.3 & 67.5 & 78.0 \\
        \hspace{-5pt}IID (\textit{CVPR'23}) & 83.4 & 83.3 & 82.8 & 80.0 & 80.2 & 81.1 & 82.2 & 41.4 & 53.3 & 79.7 & 81.8 & 81.8 & 73.7 & 65.2 & 69.9 & 63.8 & 63.3 & 63.8 & 64.2 & 72.4 \\
        \hspace{-5pt}FreqNet (\textit{AAAI'24}) & 66.8 & 60.3 & 76.7 & 59.0 & 69.2 & 77.1 & 75.1 & 33.1 & 73.1 & 70.3 & 72.8 & 77.4 & 67.7 & 50.6 & 67.0 & 62.1 & 59.3 & 58.3 & 51.2 & 64.6 \\
        \hspace{-5pt}ProDet (\textit{NIPS'24}) & 90.5 & 92.6 & 94.2 & 88.2 & 91.9 & 93.8 & 93.1 & 56.3 & 58.6 & 80.8 & 88.1 & 91.0 & 83.3 & 58.1 & 82.9 & 71.3 & 75.6 & 66.3 & 74.1 & 80.6 \\
        \hspace{-5pt}NPR (\textit{CVPR'24}) & 75.6 & 68.8 & 91.2 & 59.5 & 82.6 & 91.3 & 84.0 & 47.7 & 67.8 & 60.6 & 79.8 & 89.0 & 67.7 & 52.6 & 73.0 & 76.6 & 62.3 & 50.2 & 46.0 & 69.8 \\
        \hspace{-5pt}AIDE (\textit{ICLR'25}) & 80.4 & 68.8 & 86.3 & 64.0 & 88.9 & 95.4 & 76.0 & 56.7 & 79.2 & 86.1 & 74.2 & 62.4 & 75.7 & 59.7 & 67.9 & 49.7 & 58.0 & 51.9 & 59.2 & 70.6 \\
        \hspace{-5pt}Co-SPY (\textit{CVPR'25}) & 86.3 & 93.5 & 95.5 & 85.3 & \underline{93.3} & 96.6 & 95.3 & 77.0 & \underline{92.5} & 88.6 & \underline{90.6} & 79.1 & 87.3 & 67.6 & 80.0 & 82.5 & 74.0 & 79.5 & 64.3 & 84.7 \\
        \hspace{-5pt}D$^3$ (\textit{CVPR'25}) & 87.3 & 93.6 & 95.6 & 91.3 & 90.7 & 95.8 & 95.5 & 62.4 & 71.6 & 82.9 & 80.0 & 82.4 & 73.7 & 69.7 & 74.6 & 78.1 & 70.9 & 80.8 & 64.3 & 81.1 \\
        \hspace{-5pt}Effort (\textit{ICML'25}) & \underline{94.7} & 82.8 & 96.5 & 78.0 & 90.5 & 97.8 & \underline{98.3} & 64.7 & \textbf{94.8} & 89.7 & 89.5 & 92.9 & \underline{88.0} & 64.8 & 82.2 & 61.5 & 66.4 & 53.8 & 74.0 & 82.2 \\
        \shline
        \rowcolor{lightblue}\textit{\textbf{Generic MLLMs}} &&&&&&&&&&&&&&&&&&&&\\
        \hspace{-5pt}Qwen2.5-VL-7B & 51.2 & 50.0 & 50.0 & 49.7 & 50.0 & 52.0 & 52.9 & 50.5 & 56.7 & 50.7 & 53.6 & 54.5 & 51.6 & 50.7 & 53.6 & 80.2 & 67.5 & 52.5 & 50.5 & 54.1 \\
        \hspace{-5pt}InternVL3-8B & 54.0 & 54.0 & 49.8 & 49.0 & 56.6 & 55.8 & 57.2 & 62.9 & 54.2 & 62.9 & 63.6 & 54.8 & 67.7 & 54.4 & 67.1 & 77.1 & 66.5 & 47.4 & 51.8 & 58.3 \\
        \hspace{-5pt}MiMo-VL-7B & 63.8 & 74.5 & 77.1 & 82.5 & 60.3 & 82.4 & 81.4 & 48.7 & 82.6 & 76.4 & 79.7 & 78.4 & 82.8 & 57.7 & 75.6 & 79.4 & 70.7 & 67.7 & 54.9 & 72.5 \\
        \hspace{-5pt}GLM-4.1V-9BThink & 56.4 & 55.2 & 52.3 & 50.5 & 51.6 & 68.4 & 60.7 & 54.3 & 68.4 & 63.3 & 65.7 & 55.1 & 81.0 & 58.7 & 72.7 & 83.7 & 69.2 & 52.0 & 53.9 & 61.7 \\
        \hspace{-5pt}GPT-4o & 53.5 & 57.7 & 52.0 & 51.4 & 59.9 & 81.2 & 54.8 & 66.4 & 58.9 & 52.5 & 64.4 & 60.9 & 55.5 & 49.4 & 62.0 & 90.7 & 73.7 & 58.0 & 52.8 & 60.8 \\
        \hspace{-5pt}Gemini-2.5-Pro & 72.2 & 64.9 & 92.4 & 82.8 & 62.5 & 93.4 & 93.2 & 73.7 & 83.3 & 87.4 & 85.5 & 84.7 & 85.6 & 67.2 & 75.6 & 87.5 & 82.4 & 70.9 & 53.0 & 78.9 \\
        \shline
        \rowcolor{lightblue}\multicolumn{3}{c}{\textit{\textbf{MLLM-based Forgery Detectors}}} &&&&&&&&&&&&&&&&&&\\
        \hspace{-5pt}M2F2-Det (\textit{CVPR'25}) & - & 56.0 & 57.7 & 59.8 & 61.8 & 61.3 & 55.4 & 78.9 & 65.5 & 80.0 & 57.4 & 57.5 & 76.3 & \underline{73.0} & 56.3 & 67.2 & 50.6 & 53.0 & 70.6 & 63.2  \\
        \hspace{-5pt}FakeShield (\textit{ICLR'25}) & - & 64.3 & 64.0 & 61.5 & 63.1 & 61.8 & 63.3 & 64.0 & 57.3 & 60.9 & 58.1 & 63.6 & 63.7 & 50.2 & 83.8 & 53.8 & 51.3 & 53.9 & 55.6 & 60.8 \\
        \hspace{-5pt}SIDA-7B (\textit{CVPR'25}) & - & \textbf{97.3} & 97.7 & 79.5 & 59.3 & \underline{98.5} & 95.0 & 59.8 & 60.6 & 62.3 & 89.7 & \underline{94.4} & 63.3 & 50.4 & 81.9 & 80.0 & 78.0 & 68.9 & 57.3 & 76.3 \\
        \hspace{-5pt}SIDA-13B (\textit{CVPR'25}) & - & 80.7 & 78.5 & 54.8 & 52.5 & 91.3 & 82.4 & 63.7 & 61.2 & 68.2 & 56.7 & 67.1 & 84.3 & 60.8 & 58.2 & 88.3 & 74.0 & 74.1 & 59.9 & 69.8 \\
        \hspace{-5pt}FFAA (\textit{Arxiv'24}) & - & 55.1 & 50.9 & 72.9 & 63.5 & 60.8 & 57.6 & \underline{82.7} & 70.9 & 71.8 & 58.4 & 62.4 & 86.0 & 67.7 & 58.4 & 55.3 & 59.2 & 49.6 & 68.3 & 64.0 \\
        \hspace{-5pt}FakeVLM (\textit{NIPS'25}) & - & 78.2 & 78.5 & 77.0 & 74.5 & 76.5 & 76.8 & 70.8 & 76.2 & 76.2 & 76.9 & 76.5 & 77.7 & \textbf{75.7} & \underline{83.6} & 81.5 & 80.8 & 78.7 & \underline{74.5} & 77.3 \\
        \rowcolor{Gray}\hspace{-5pt}\veritasmini & - & \underline{95.5} & \underline{99.1} & \textbf{97.3} & 72.8 & 97.0 & 96.1 & 82.5 & 76.3 & \underline{90.0} & 83.7 & 82.9 & 79.3 & 72.5 & 78.7 & \underline{92.0} & \textbf{93.0} & \underline{85.5} & 70.6 & \underline{85.8} \\
        \shline
        \hspace{-5pt}\rule{0pt}{8pt}\gc{\veritas $\;$(cold-start)} & \gc{96.8} & \gc{79.5} & \gc{99.6} & \gc{96.0} & \gc{99.9} & \gc{99.7} & \gc{99.9} & \gc{84.0} & \gc{65.3} & \gc{94.8} & \gc{86.2} & \gc{93.4} & \gc{86.7} & \gc{55.9} & \gc{73.5} & \gc{93.7} & \gc{89.3} & \gc{88.1} & \gc{76.4} & \gc{87.3} \\
        \rowcolor{Gray}\hspace{-7pt} \veritas $\;$(\textbf{ours}) & \textbf{97.3} & 94.8 & \textbf{99.8} & \underline{97.0} & \textbf{99.9} & \textbf{99.9} & \textbf{99.9} & \textbf{90.3} & 75.7 & \textbf{97.0} & \textbf{91.8} & \textbf{95.1} & \textbf{91.7} & 58.6 & \textbf{84.1} & \textbf{92.3} & \underline{90.2} & \textbf{89.2} & \textbf{78.5} & \textbf{90.7} \\
        \bottomrule[1pt]
        \end{tabular}
    }
    \label{tab:main}
    \vspace{-0.3cm}
\end{table}

\begin{figure}[t]
    \scriptsize
    \centering
	\begin{minipage}{0.34\linewidth}
            \captionof{table}{Effect of the proposed pattern-aware reasoning.}
            \vspace{-7pt}
    \label{tab:pattern_mode_abl}
    \renewcommand{\arraystretch}{1.04}
    \centering
    \scalebox{0.92}{
        \begin{tabular}{p{60pt}<{\raggedright}p{6pt}<{\centering}p{6pt}<{\centering}p{6pt}<{\centering}p{6pt}<{\centering}}
        \hspace{-5pt}Model & \makebox[0pt][l]{\hspace{-0.7em}ID} & \makebox[0pt][l]{\hspace{-0.8em}CM} & \makebox[0pt][l]{\hspace{-0.6em}CF} & \makebox[0pt][l]{\hspace{-0.7em}CD} \\
        \shline
        \hspace{-5pt}w/o Reasoning & \makebox[0pt][l]{\hspace{-0.8em}\textbf{97.8}} & \makebox[0pt][l]{\hspace{-0.8em}93.3} & \makebox[0pt][l]{\hspace{-0.8em}73.0} & \makebox[0pt][l]{\hspace{-0.8em}69.5} \\
        \hspace{-5pt}Post-hoc Explanation & \makebox[0pt][l]{\hspace{-0.8em}96.3} & \makebox[0pt][l]{\hspace{-0.8em}95.0} & \makebox[0pt][l]{\hspace{-0.8em}79.0} & \makebox[0pt][l]{\hspace{-0.9em}76.8} \\
        \hspace{-5pt}Flexible Reasoning & \makebox[0pt][l]{\hspace{-0.8em}96.2} & \makebox[0pt][l]{\hspace{-0.8em}94.3} & \makebox[0pt][l]{\hspace{-0.8em}81.2} & \makebox[0pt][l]{\hspace{-0.9em}76.8} \\
        \rowcolor{Gray}\hspace{-5pt}Pattern-aware Reason. & \makebox[0pt][l]{\hspace{-0.8em}96.9} & \makebox[0pt][l]{\hspace{-0.8em}\textbf{98.4}} & \makebox[0pt][l]{\hspace{-0.8em}\textbf{87.4}} & \makebox[0pt][l]{\hspace{-0.9em}\textbf{80.1}} \\
        \rowcolor{Gray}\hspace{-5pt}\textbf{$\bm{\Delta}$} Flexible Reason. & \textcolor{red}{\textbf{\makebox[0pt][l]{\hspace{-0.8em}+0.7}}} & \textcolor{red}{\textbf{\makebox[0pt][l]{\hspace{-0.8em}+4.1}}} & \textcolor{red}{\textbf{\makebox[0pt][l]{\hspace{-0.8em}+6.2}}} & \textcolor{red}{\textbf{\makebox[0pt][l]{\hspace{-0.8em}+3.3}}} \\
        \end{tabular}
    }
	\end{minipage}
    \begin{minipage}{0.34\linewidth}
            \captionof{table}{Ablations on the reward functions in P-GRPO.}
            \vspace{-7pt}
    \label{tab:pgrpo_abl}
    \renewcommand{\arraystretch}{1.04}
    \centering
    \scalebox{0.92}{
        \begin{tabular}{p{8pt}<{\centering}p{5pt}<{\centering}p{5pt}<{\centering}p{5pt}<{\centering}p{6pt}<{\centering}p{6pt}<{\centering}p{6pt}<{\centering}p{6pt}<{\centering}}
        \makebox[0pt][l]{\hspace{-1.4em}$R_{\text{pattern}}$} & \makebox[0pt][l]{\hspace{-0.8em}$R_{\text{ref}}$} & \makebox[0pt][l]{\hspace{-1em}$R_{\text{fmt}}$} & \makebox[0pt][l]{\hspace{-1em}\gc{$R_{\text{acc}}$}} & \makebox[0pt][l]{\hspace{-0.7em}ID} & \makebox[0pt][l]{\hspace{-0.8em}CM} & \makebox[0pt][l]{\hspace{-0.6em}CF} & \makebox[0pt][l]{\hspace{-0.7em}CD} \\
        \shline
         & & \checkmark & \gc{\checkmark} & \makebox[0pt][l]{\hspace{-0.8em}97.2} & \makebox[0pt][l]{\hspace{-0.8em}96.4} & \makebox[0pt][l]{\hspace{-0.8em}86.3} & \makebox[0pt][l]{\hspace{-0.8em}79.9}\\
        & \checkmark & \checkmark & \gc{\checkmark} & \makebox[0pt][l]{\hspace{-0.8em}97.0} & \makebox[0pt][l]{\hspace{-0.8em}97.3} & \makebox[0pt][l]{\hspace{-0.8em}87.0} & \makebox[0pt][l]{\hspace{-0.8em}80.7} \\
        \checkmark & & \checkmark & & \makebox[0pt][l]{\hspace{-0.8em}97.3} & \makebox[0pt][l]{\hspace{-0.8em}97.7} & \makebox[0pt][l]{\hspace{-0.8em}87.9} & \makebox[0pt][l]{\hspace{-0.8em}81.4} \\
       \rowcolor{Gray}\checkmark & \checkmark & \checkmark & & \makebox[0pt][l]{\hspace{-0.8em}97.3} & \makebox[0pt][l]{\hspace{-0.8em}98.6} & \makebox[0pt][l]{\hspace{-0.8em}90.3} & \makebox[0pt][l]{\hspace{-0.8em}82.2} \\
       \rowcolor{Gray}\multicolumn{3}{c}{\makebox[0pt][l]{\hspace{-3em}\textbf{$\bm{\Delta}$} GRPO}} & & \textcolor{red}{\textbf{\makebox[0pt][l]{\hspace{-0.8em}+0.1}}} & \textcolor{red}{\textbf{\makebox[0pt][l]{\hspace{-0.8em}+2.2}}} & \textcolor{red}{\textbf{\makebox[0pt][l]{\hspace{-0.8em}+4.0}}} & \textcolor{red}{\textbf{\makebox[0pt][l]{\hspace{-0.8em}+2.3}}} \\
        \end{tabular}
    }
	\end{minipage}
    \begin{minipage}{0.31\linewidth}
            \captionof{table}{Ablations on different base models and model sizes.}
            \vspace{-7pt}
    \label{tab:base_model}
    \renewcommand{\arraystretch}{1.04}
    \centering
    \scalebox{0.92}{
        \begin{tabular}{p{44pt}<{\raggedright}p{6pt}<{\centering}p{6pt}<{\centering}p{6pt}<{\centering}p{6pt}<{\centering}}
        \hspace{-5pt}Base Model & \makebox[0pt][l]{\hspace{-0.7em}ID} & \makebox[0pt][l]{\hspace{-0.8em}CM} & \makebox[0pt][l]{\hspace{-0.6em}CF} & \makebox[0pt][l]{\hspace{-0.7em}CD} \\
        \shline
        \hspace{-5pt}Qwen2.5-VL-7B & \makebox[0pt][l]{\hspace{-0.8em}96.8} & \makebox[0pt][l]{\hspace{-0.8em}97.7} & \makebox[0pt][l]{\hspace{-0.8em}89.0} & \makebox[0pt][l]{\hspace{-0.8em}81.4} \\
        \hspace{-5pt}MiMo-VL-7B & \makebox[0pt][l]{\hspace{-0.8em}93.0} & \makebox[0pt][l]{\hspace{-0.8em}98.6} & \makebox[0pt][l]{\hspace{-0.8em}82.6} & \makebox[0pt][l]{\hspace{-0.8em}\textbf{83.0}} \\
        \hspace{-5pt}InternVL3-2B & \makebox[0pt][l]{\hspace{-0.8em}97.4} & \makebox[0pt][l]{\hspace{-0.8em}97.4} & \makebox[0pt][l]{\hspace{-0.8em}87.3} & \makebox[0pt][l]{\hspace{-0.8em}80.4} \\
        \rowcolor{Gray}\hspace{-5pt}InternVL3-8B & \makebox[0pt][l]{\hspace{-0.8em}97.3} & \makebox[0pt][l]{\hspace{-0.8em}98.6} & \makebox[0pt][l]{\hspace{-0.8em}89.3} & \makebox[0pt][l]{\hspace{-0.9em}82.2} \\
        \hspace{-5pt}InternVL3-14B & \makebox[0pt][l]{\hspace{-0.8em}\textbf{98.5}} & \makebox[0pt][l]{\hspace{-0.8em}\textbf{99.3}} & \makebox[0pt][l]{\hspace{-0.8em}\textbf{92.2}} & \makebox[0pt][l]{\hspace{-0.8em}82.6} \\
        \end{tabular}
    }
	\end{minipage}
    \vspace{-0.4cm}
\end{figure}

\subsection{Main Results}
\textbf{Comparison to SOTA detectors.}
As shown in Table~\ref{tab:main}, our \veritas $\,$model achieves SOTA performance on four evaluation scenarios, achieving 6.0\% averaged gains over the previous best.
Existing detectors show great performance on cross-model split (over 90\% for D$^3$) but fall short on cross-forgery and cross-domain scenarios (mostly less than 85\%).
\veritas $\,$ mitigates the gap, achieving over 90.0\% accuracy on unseen forgery such as face restoration and personalization, and over 90.0\% on in-the-wild data from Dreamina and 89.2\% on GPT-4o.
The cold-start model also achieves promising results, but without incentivizing planning and self-reflection, the cross-forgery results are degraded.
More results and analyses can be found in Appendix~\ref{supp:more_results}.

\textbf{Comparison to SOTA MLLMs.}
Compared to our base model, \veritas $\,$achieves 32.4\% averaged gain, suggesting the effectiveness of our training strategy.
The models with similar sizes show limited abilities for deepfake detection, with less than 60\% accuracy.
Gemini-2.5-Pro shows the best capacities among these MLLMs, even outperforming some of the fine-tuned detectors.
\veritas$\,$ surpasses Gemini-2.5-Pro by 11.8\%, demonstrating great generalization.

\textbf{Comparisons to MLLM-based detectors.}
For fair comparisons, we restrict our training scope (Table~\ref{tab:main}).
Even with limited data scope, \veritasmini$\,$ \textit{still outperforms existing MLLM-based detectors}, indicating the effectiveness of the proposed framework.
M2F2-Det and FFAA, though targeted at deepfake detection, suffer from poor generalization on HydraFake.
SIDA-7B and FakeVLM achieve promising results by contrast.
Moreover, \veritas$\,$ exhibits certain advantages in both detection accuracy and reasoning depth (Figure~\ref{fig:mllm_comp_main}).
More cases can be found in Appendix~\ref{supp:more_qualitative_comp}.

\subsection{Ablation Studies}
\vspace{-0.2cm}
We provide primary ablations in main text.
More analyses on the training protocol (\ref{supp:cross_bench}), results on recent benchmark~\citep{li2025artificial}(\ref{supp:cross_bench}), selection of P-GRPO training data (\ref{supp:pgrpo_data}) and reward model (\ref{supp:reward_model}), hyperparameters (\ref{supp:ana_hyper}) and efficiency analysis (\ref{supp:efficiency}) can be found in Appendix.

\textbf{Effect of pattern-aware reasoning.}
As shown in Table~\ref{tab:pattern_mode_abl}, we compare different reasoning paradigms using SFT and P-GRPO training.
Although the improvements on in-domain datasets are marginal, our pattern-aware reasoning demonstrates clear advantages to flexible reasoning on OOD scenarios, achieving 6.2\% and 3.3\% gains on CF and CD testing respectively.
The post-hoc explanation adopted in recent methods exhibits degraded performance in OOD testing, further verifying the superiority of pattern-aware reasoning.

\begin{figure}[t]
\vspace{-2pt}
    \scriptsize
    \centering
\begin{minipage}{0.44\linewidth}
		\centering
    \includegraphics[height=3.3cm, width=0.94\linewidth]{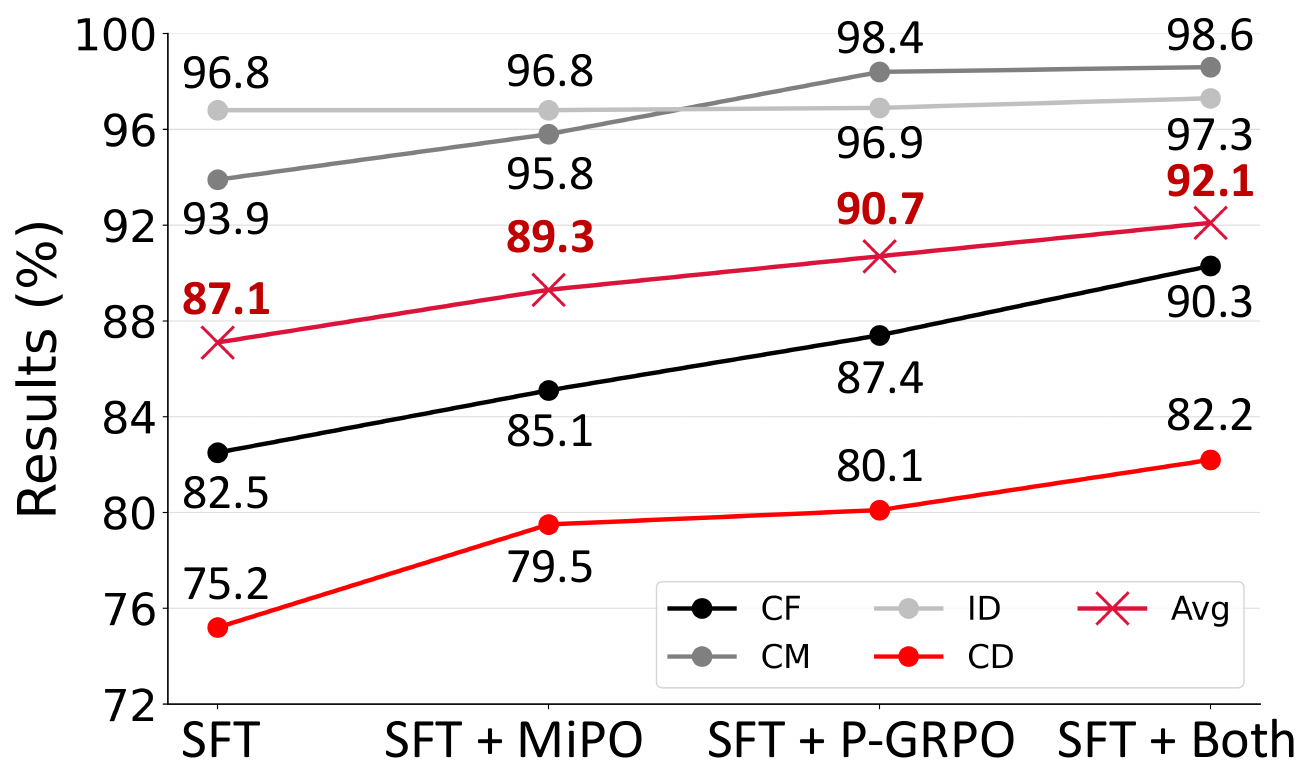}
    \vspace{-5pt}
    \caption{Ablations on the training stages.
    ``Avg'' is directly averaged across four splits.}
	\label{fig:training_stage}
\end{minipage}
\hfill
\begin{minipage}{0.52\linewidth}
		\centering
    \includegraphics[height=3.3cm, width=0.99\linewidth]{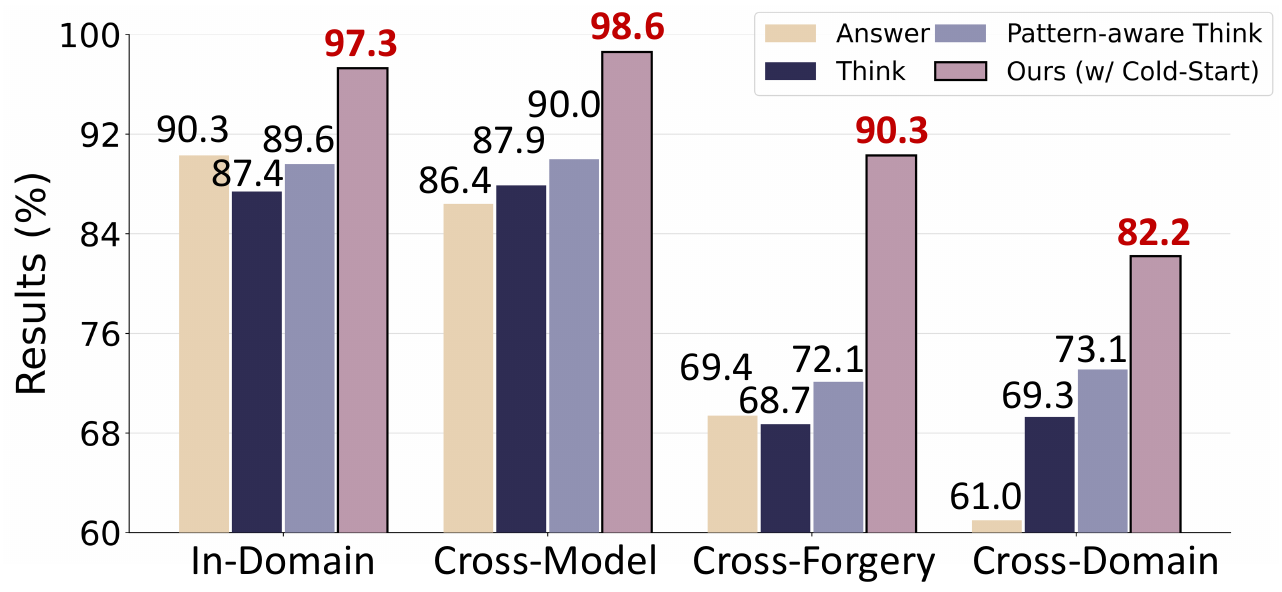}
    \vspace{-5pt}
    \caption{Effect of Cold-Start. We compare different settings of pure RL.}
	\label{fig:pure_rl}
\end{minipage}
\vspace{-0.5cm}
        
\end{figure}

\textbf{Ablations on different training stages.}
As shown in Figure~\ref{fig:training_stage}, we investigate the effect of each training stage.
Applying MiPO or P-GRPO upon SFT model both achieve significant gains, with P-GRPO performing better, which is due to the online sampling and pattern-aware incentivization.
Applying MiPO before P-GRPO yields the best performance, achieving 2.9\% and 2.1\% gains on CF and CD testing respectively.
This is because \textit{MiPO ensures high-quality rollouts in subsequent stage}, facilitating \textit{more accurate policy updates for online RL}.

\begin{figure}[t]
    \scriptsize
    \centering
	\begin{minipage}{0.47\linewidth}
            \captionof{table}{Effect of reasoning patterns. SFT and P-GRPO are performed for comparisons.}
            \vspace{-7pt}
    \label{tab:fg_pattern_abl}
    \renewcommand{\arraystretch}{1.04}
    \centering
    \scalebox{0.92}{
        \begin{tabular}{p{76pt}<{\raggedright}p{7pt}<{\centering}p{7pt}<{\centering}p{7pt}<{\centering}p{7pt}<{\centering}p{8pt}<{\centering}}
        \hspace{-5pt}Model & \makebox[0pt][l]{\hspace{-0.7em}ID} & \makebox[0pt][l]{\hspace{-0.8em}CM} & \makebox[0pt][l]{\hspace{-0.6em}CF} & \makebox[0pt][l]{\hspace{-0.7em}CD} & \makebox[0pt][l]{\hspace{-0.6em}Avg.} \\
        \shline
        \hspace{-5pt}Flexible Reasoning & 96.2 & 94.3 & 81.2 & 76.8 & 87.1 \\
        \rowcolor{Gray}\hspace{-5pt}Pattern-aware Reasoning & 96.9 & 98.4 & \textbf{87.4} & \textbf{80.1} & \textbf{90.7} \\
        \hspace{-5pt}w/o \textless fast\textgreater & \textbf{97.3} & \textbf{98.8} & 86.9 & 79.1 & 90.5 \\
        \hspace{-5pt}w/o \textless planning\textgreater & 96.7 & 96.9 & 85.0 & \textbf{80.1} & 89.7 \\
        \hspace{-5pt}w/o \textless reflection\textgreater & 97.0 & 97.2 & 82.5 & 77.3 & 88.5 \\
        \hspace{-5pt}w/o \textless conclusion\textgreater & 97.2 & 98.2 & 86.2 & 79.0 & 90.1 \\
        \end{tabular}
    }
	\end{minipage}
    \begin{minipage}{0.47\linewidth}
            \captionof{table}{Ablations on non-preference in MiPO.}
            \vspace{-5pt}
    \label{tab:fg_mipo_abl}
    \renewcommand{\arraystretch}{1.04}
    \centering
    \scalebox{0.92}{
        \begin{tabular}{p{60pt}<{\raggedright}p{7pt}<{\centering}p{7pt}<{\centering}p{7pt}<{\centering}p{7pt}<{\centering}p{8pt}<{\centering}}
        \hspace{-5pt}Model & \makebox[0pt][l]{\hspace{-0.7em}ID} & \makebox[0pt][l]{\hspace{-0.8em}CM} & \makebox[0pt][l]{\hspace{-0.6em}CF} & \makebox[0pt][l]{\hspace{-0.7em}CD} & \makebox[0pt][l]{\hspace{-0.6em}Avg.} \\
        \shline
        \rowcolor{Gray}\hspace{-5pt}\veritas & \textbf{97.3} & \textbf{98.6} & \textbf{90.3} & \textbf{82.2} & \textbf{92.1} \\
        \hspace{-5pt}w/o MiPO & 96.9 & 98.4 & 87.4 & 80.1 & 90.7 \\
        \hspace{-5pt}MiPO (w/o $\bm{s}_l^{\phi}$) & 96.9 & 98.6 & 89.2 & 81.4 & 91.5 \\
        \hspace{-5pt}MiPO (w/o $\bm{s}_l^{\psi}$) & 65.3 & 64.8 & 58.6 & 54.3 & 60.8 \\
        \end{tabular}
    }
	\end{minipage}
    \vspace{-0.2cm}
\end{figure}

\textbf{Effect of Pattern-guided Cold-Start.}
As shown in Figure~\ref{fig:pure_rl}, we investigate different RL settings without cold-start.
The training data keeps consistent with our two-stage pipeline.
Answer-only model achieves better ID results while incorporating thinking improves CM and CD performance.
However, all settings underperform the model with cold-start.
The low-quality explorations lead to unstable training.
Results in Figure~\ref{fig:training_stage} further verify the effectiveness of MiPO during cold-start.

\textbf{Effect of Pattern-aware GRPO.}
As shown in Table~\ref{tab:pgrpo_abl}, our P-GRPO achieves noticeable improvements compared to original GRPO.
Specifically, pattern-aware reward outperforms the vanilla accuracy reward especially on CF and CD scenarios.
The reflection quality reward benefits both original GRPO and our P-GRPO, which demonstrates the importance of high-quality self reflection.
In Appendix~\ref{supp:pgrpo_data}, we observe that by adding several ``unseen`` data in P-GRPO, the 
ODD performance can be further improved, demonstrating promising \textit{scalability} with only binary labels required.

\textbf{Effect of specific reasoning patterns.}
As shown in Table~\ref{tab:fg_pattern_abl},
``fast judgement'' is helpful for CF and CD, but is not critical overall.
``planning'' is more effective on CM, since the fully synthesized images require a more holistic and structured analysis.
``self-reflection'' is critical especially on CF and CD, as it incentivizes the model to discover those unseen artifacts.
``conclusion'' provides certain gains, suggesting that synthesizing separate evidence into a coherent verdict is also important.

\textbf{Ablations on the non-preference in MiPO.}
As shown in Table~\ref{tab:fg_mipo_abl},
$\bm{s}_l^{\phi}$ helps improve the performance on CF (+1.3\%) and CD (+0.8\%) scenarios.
To understand the effects, we provide a qualitative case in Figure~\ref{fig:mipo_comp}.
Without $\bm{s}_l^{\phi}$, the model still gets correct answers, but the analysis is superficial and less detailed, which causes certain failures on unseen forgeries that might require in-depth reasoning.

\subsection{Further Analyses}
\textbf{Evaluation of reasoning quality.}
To evaluate the reasoning quality, we take two types of assessments: (1) score evaluation which is based on predefined criteria (Figure~\ref{prompt:score}).
(2) Pairwise comparison which directly compares outputs from two models.
We adopt MLLM-as-a-Judge~\citep{chen2024mllm}, using GPT-4o and Gemini-2.5-Pro for evaluation.
Similar to~\citep{zhou2025aigi}, we randomly select $1$K samples for evaluation.
As shown in Table~\ref{tab:reason_quality}, our model achieves the best score and ELO rating, where MiPO greatly improves the reasoning quality.
Moreover, our MiPO outperforms DPO in raising reasoning quality, which verifies the effectiveness of mixed non-preference strategy.

\textbf{Different fine-tuned base models and model sizes.}
As shown in Table~\ref{tab:base_model}, we adopt different MLLMs as our base model.
InternVL3-8B outperforms Qwen2.5-VL-7B and MiMo-VL-7B, due to the dynamic high resolution strategy.
InternVL3-2B achieves promising performance with fewer parameters, while scaling up to 14B yields considerable gains on CM and CF scenarios.

\textbf{Robustness evaluation.}
We investigate the performance under JPEG compression and Gaussian blur.
Results in Table~\ref{tab:robust} highlight the robustness of our model.
Our model achieves consistently high performance under JPEG compression and maintains state-of-the-art results across different perturbations.
Notably, this robustness is achieved without training on corresponding data augmentations such as random Gaussian blur, which instead are commonly adopted in previous methods.

\begin{figure}[t]
    \scriptsize
    \centering
    \vspace{-0.4cm}
	\begin{minipage}{0.43\linewidth}
            \captionof{table}{Evaluation of reasoning quality. We utilize score and pairwise ELO rating.}
            \vspace{-7pt}
    \label{tab:reason_quality}
    \renewcommand{\arraystretch}{1.04}
    \centering
    \scalebox{0.92}{
        \begin{tabular}{p{60pt}<{\raggedright}p{24pt}<{\centering}p{35pt}<{\centering}p{18pt}<{\centering}}
        \multirow{2}{*}{\hspace{-5pt}\textbf{Model}} & \multicolumn{2}{c}{\textbf{Score Evaluation}} & \multirow{2}{*}{\vspace{0.5em}\textbf{ELO}} \\
        \cmidrule(lr){2-3}
        & GPT-4o & \makebox[0pt][l]{\hspace{-3.8em}Gemini-2.5-Pro} & \textbf{Rating}  \\
        \shline
        \hspace{-5pt}MiMo-VL-7B & 3.0731 & 2.5785 & 695.0 \\
        \hspace{-5pt}GPT-4o & 2.4718 & 2.1619 & 785.1 \\
        \hspace{-5pt}Gemini-2.5-Pro & 4.1681 & 4.0070 & 966.9 \\
        \cline{1-4}
        \hspace{-5pt}\rule{0pt}{7pt}\veritas $\;$(w/o MiPO) & 4.2538 & 4.1502 & 984.0 \\
        \hspace{-5pt}\veritas $\;$(w/ DPO) & 4.5077 & 4.2863 & 1210.0 \\
        \rowcolor{Gray}\hspace{-5pt}\veritas $\;$(w/ MiPO) & \textbf{4.6479} & \textbf{4.4214} & \textbf{1359.0}  \\
        \end{tabular}
    }
	\end{minipage}
    \begin{minipage}{0.55\linewidth}
            \captionof{table}{Robustness on Compression, Blur and Resize. The results are averaged across all sets.}
            \vspace{-7pt}
    \label{tab:robust}
    \renewcommand{\arraystretch}{1.04}
    \centering
    \scalebox{0.94}{
        \begin{tabular}{p{25pt}<{\raggedright}p{16pt}<{\centering}p{20pt}<{\centering}p{20pt}<{\centering}p{20pt}<{\centering}p{24pt}<{\centering}p{24pt}<{\centering}}
        \multirow{2}{*}{\hspace{-5pt}\textbf{Method}} & \multirow{2}{*}{\makebox[0pt][l]{\hspace{-2.2em}\textbf{Original}}} & \multicolumn{3}{c}{\makebox[0pt][l]{\hspace{-4.2em}\textbf{JPEG Compression}}} & \multicolumn{2}{c}{\textbf{Gaussian Blur}} \\
        \cmidrule(lr){3-5} \cmidrule(lr){6-7}
         & & QF=90 & QF=70 & QF=50 & $\sigma\!=\!1.0$ & $\sigma\!=\!2.0$ \\
        \shline
        \hspace{-5pt}UniFD & 78.0 & 77.1 & 75.4 & 72.2 & 76.5 & 75.4 \\
        \hspace{-5pt}IID & 72.4 & 71.4 & 70.1 & 69.7 & 72.1 & 71.9 \\
        \hspace{-5pt}ProDet & 80.6 & 80.1 & 79.4 & 78.7 & 79.7 & 77.8 \\
        \hspace{-5pt}Co-SPY & 84.7 & 84.5 & 83.4 & 81.0 & 82.7 & 77.0 \\
        \hspace{-5pt}Effort & 82.2 & 73.1 & 79.6 & 66.3 & 79.1 & 70.4 \\
        \rowcolor{Gray}\hspace{-5pt}\veritas & \textbf{90.7} & \textbf{90.1} & \textbf{88.7} & \textbf{87.4} & \textbf{88.8} & \textbf{84.3} \\
        \end{tabular}
    }
	\end{minipage}
    \vspace{-0.4cm}
\end{figure}

\begin{figure*}[t]
    \centering
    \includegraphics[width=0.89\linewidth]{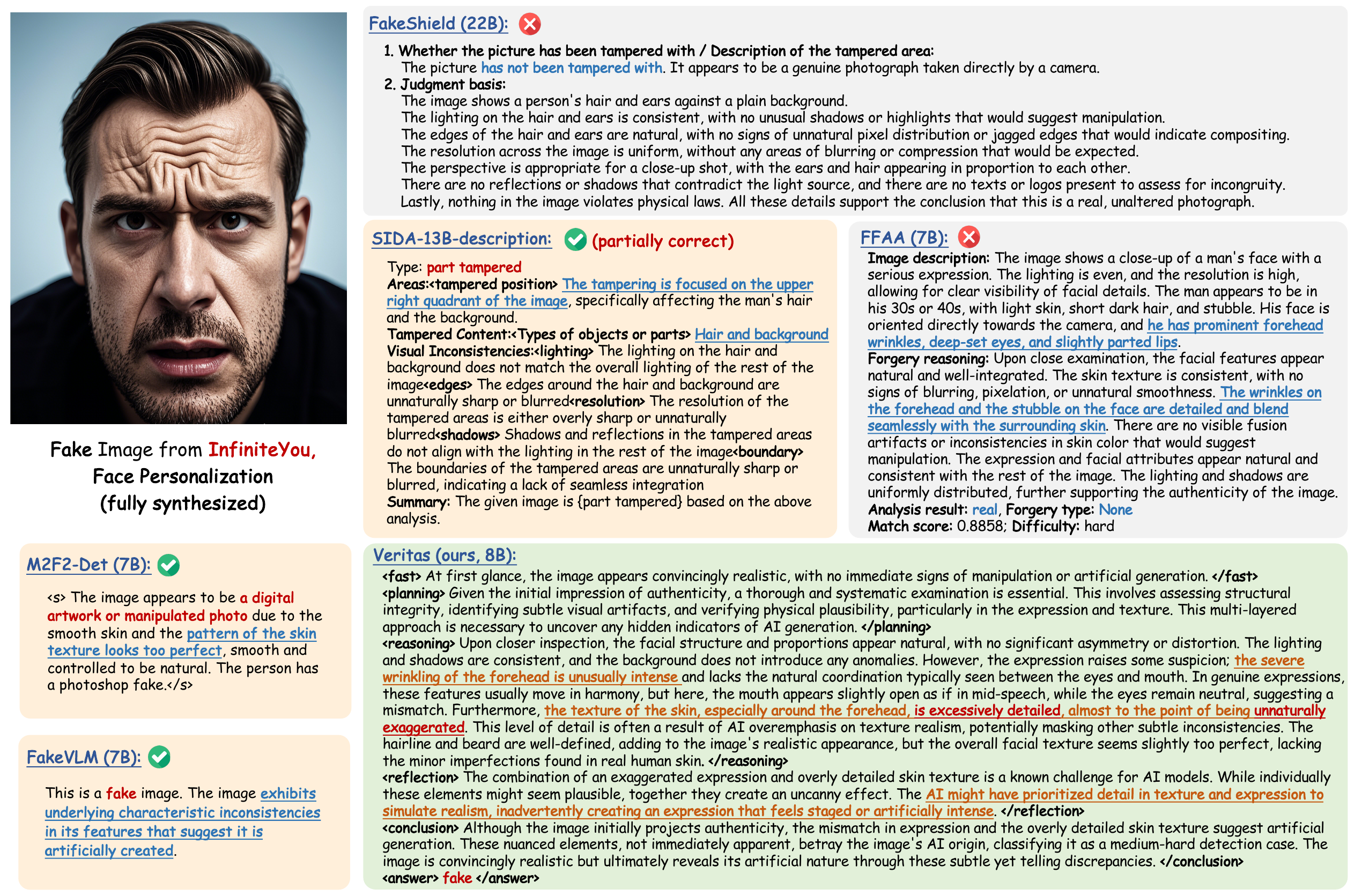}
    \vspace{-8pt}
     \caption{Reasoning comparisons between \veritas$\,$ and existing MLLM-based detectors.
     }
	\label{fig:mllm_comp_main}
    \vspace{-0.4cm}
\end{figure*}

\section{Conclusion}
\vspace{-0.2cm}
In this paper, we introduce HydraFake dataset and \veritas$\,$ model.
HydraFake introduces a holistic evaluation protocol to comprehensively measure the generalization capacities.
We then train a multi-modal large language model (MLLM) based deepfake detector trained with our two-stage pipeline.
Results on HydraFake show that current detectors struggle on cross-forgery and cross-domain scenarios, while our model greatly mitigates the gap and is capable of delivering transparent decision process.
We hope this work can inspire more generalizable and reliable deepfake detection.

\subsubsection*{Acknowledgments}
This work was supported by the Beijing Natural Science Foundation JQ23016, the Chinese National Natural Science Foundation Projects 62476273, 62406320, 62276254 and U23B2054, the Science and Technology Development Fund of Macau Project 0123/2022/A3, 0140/2024/AGJ, 0044/2024/AGJ and 0084/2024/RIB2, and Ant Group.

\bibliography{iclr2026_conference}
\bibliographystyle{iclr2026_conference}

\clearpage

\appendix
\section{Appendix}

The appendix is organized as follows:

$\bullet$ \S\ref{supp:dataset} Details of HydraFake Dataset.
\begin{itemize}
    \item \S\ref{supp:dataset_train} Training Set.
    \item \S\ref{supp:dataset_test_id} In-Domain Evaluation.
    \item \S\ref{supp:dataset_test_cm} Cross-Model Evaluation.
    \item \S\ref{supp:dataset_test_cf} Cross-Forgery Evaluation.
    \item \S\ref{supp:dataset_test_cd} Cross-Model Evaluation.
\end{itemize}
$\bullet$ \S\ref{supp:more_imple} More Implementation Details.

$\bullet$ \S\ref{supp:more_discussion} More Discussions.

$\bullet$ \S\ref{supp:data_annotate} Multi-Step Annotation Pipeline.

$\bullet$ \S\ref{supp:more_results} More Experimental Results.
\begin{itemize}
    \item \S\ref{supp:more_res_hydrafake} More Results and Analyses on HydraFake.
    \item \S\ref{supp:cross_bench} Cross Benchmark Comparison.
    \item \S\ref{supp:efficiency} Efficiency Comparison.
    \item \S\ref{supp:pgrpo_data} Effect of Training Data in P-GRPO Stage.
    \item \S\ref{supp:ana_hyper} Analysis of Hyperparameters.
    \item \S\ref{supp:reward_model} Effect of Different Reward Model.
\end{itemize}

$\bullet$ \S\ref{supp:full_prompts} Full Prompt Templates.

$\bullet$ \S\ref{supp:more_qualitative_res} More Qualitative Results.

$\bullet$ \S\ref{supp:more_qualitative_comp} More Qualitative Comparisons with Existing MLLM-based Detectors.

$\bullet$ \S\ref{supp:failure_ana} Failure Analysis of \veritas.

$\bullet$ \S\ref{supp:ethics_state} Ethics Statement.

$\bullet$ \S\ref{supp:future_work} Limitations and Future Work.

\subsection{Details of HydraFake Dataset}
\label{supp:dataset}
In this section, we provide more details about our HydraFake Dataset.
We introduce the dataset from the perspective of training and evaluation protocols.

\subsubsection{Training}
\label{supp:dataset_train}
\textbf{Real Images.}
HydraFake dataset contains real images from $8$ public datasets.
We extract $5$ subsets as the training set, containing $3$ low-quality subsets and $2$ high-quality subsets.
The low-quality images include FF++~\citep{rossler2019faceforensics++}, CelebA~\citep{liu2015deep} and LFW~\citep{huang2008labeled}.
The high-quality images include FFHQ~\citep{karras2019style} and CelebAHQ~\citep{karras2017progressive}.
This results in $24$K real images for training.

\textbf{Fake Images.}
In practical scenario, there are abundant fake images for training, while these images have two attributes: (1) the quality of the images varies greatly, and (2) the forgery types are often limited.
To mimic such setting, we extract $21$ subsets as the training set and strictly control the seen forgeries.
We only include face swapping (FS), face reenactment (FR) and entire face generation (EFG) in our training set, leaving various forgery types unseen.
Moreover, the deepfake methods in our training set are not the latest, leaving fresh methods in the evaluation.
\begin{itemize}[left=2pt]
    \item \textbf{FS}: FF++, BlendFace, FSGAN, SimSwap, FaceDancer, MobileSwap.
    \item \textbf{FR}: FF++, Facevid2vid, Hallo, Hallo2, LivePortrait, AniPortrait, EmoPortrait
    \item \textbf{EFG}: Dall-e 1, StyleGAN, StyleGAN2, VQGAN, Midjourney, Seeprettyface, Stable Cascade, Stable Diffusion XL, Attend-and-Excite.
\end{itemize}

\begin{figure*}[t]
    \centering
    \includegraphics[width=0.98\linewidth]{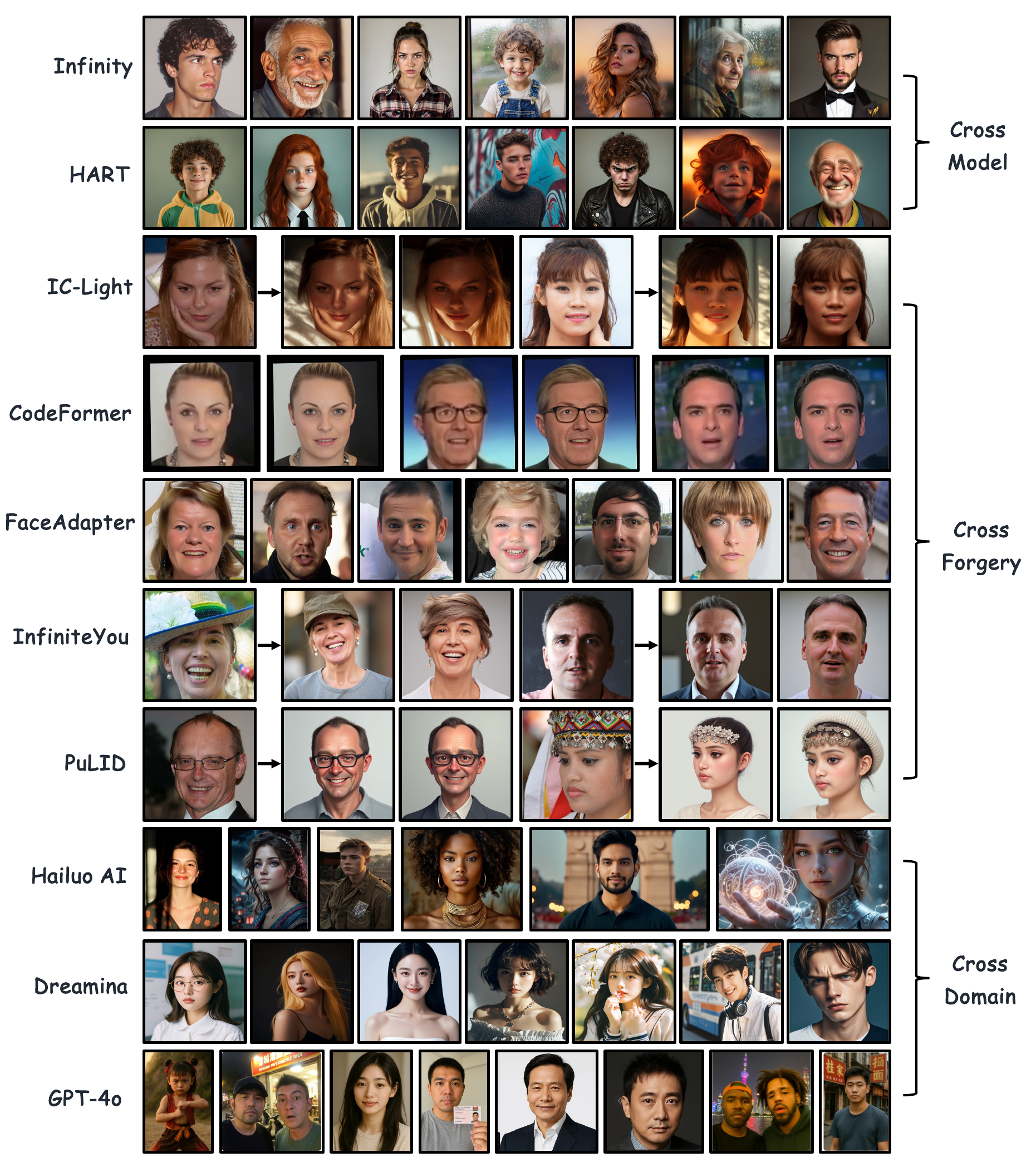}
     \caption{Examples of our self-constructed subsets in HydraFake datasets.}
	\label{fig:data_example}
\end{figure*}

\subsubsection{In-Domain Evaluation}
\label{supp:dataset_test_id}
For in-domain testing, we select $5$ subsets from the training set, and use the unseen identities as the testing samples.
Specifically, we make a balance on the image quality and forgery types, choosing a FS dataset FF++ (low-quality), two FR datasets Facevid2vid (low-quality) and Hallo2 (high-quality), and two EFG datasets StyleGAN~\citep{karras2019style} (high-quality) and Midjourney (high-quality).
For low-quality subsets, the real images are sampled from FF++.
For high-quality subsets, the real images are sampled from FFHQ.

\subsubsection{Cross-Model Evaluation}
\label{supp:dataset_test_cm}
The cross-model testing images come from deepfakes generated using unseen models.
While the difficulty varies according to the model architectures.
This includes $6$ subsets, i.e., Adobe Firefly~\citep{firefly}, MAGI-1~\citep{magi}, Flux1.1 Pro~\citep{FLUX}, StarryAI~\citep{starryai}, Infinity~\citep{han2025infinity} and HART~\citep{tang2024hart}.

\textbf{Infinity.}
Infinity is a Bitwise Visual AutoRegressive Model (VAR) capable of generating high-resolution and photorealistic images from textual prompts.
Infinity redefines visual autoregressive model under a bitwise token prediction framework with an infinite-vocabulary tokenizer and bitwise self-correction.
We reproduce the Infinity-8B model, which is capable of generating $1024\times1024$ images.
The control prompts are generated using Qwen3-32B model, which leverages a template-like sentence, balances between gender and age and avoids any semantic conflicts (e.g., wrinkles on the little girl's face) with the help of LLM.

\textbf{HART.}
Hybrid Autoregressive Transformer (HART) is an autoregressive (AR) visual generation model capable of directly generating $1024\times1024$ images, rivaling diffusion models in image generation quality.
It contains a hybrid image tokenizer to improve the fidelity of the generated images.
We reproduce HART model, which is based on Qwen2-VL-1.5B.
The textual prompts are consistent with Infinity.

\textbf{Adobe Firefly.}
Adobe Firefly is a proprietary suite of multimodal generative AI models developed by Adobe.
It is built upon a deeply customized and optimized diffusion model.
It exclusively utilizes the Adobe Stock library, open-licensed content, and public domain works, thereby designed to ensure the commercial viability and mitigate copyright risks of its generated outputs.
We collect this subset from WILD~\citep{bongini2025wild}, which is generated using template-like textual prompts.

\textbf{StarryAI.}
Starry AI is an advanced generative artificial intelligence model engineered for high-fidelity text-to-image synthesis. Its core architecture integrates a transformer-based encoder for the semantic interpretation of textual prompts with a latent diffusion model for the iterative synthesis of visual content.
The model excels at translating complex, abstract descriptions into visually coherent and stylistically nuanced imagery.
We collect this subset from WILD~\citep{bongini2025wild}, which is generated using template-like textual prompts.

\textbf{MAGI-1.}
MAGI-1 is an autoregressive denoising video generation model (Video AR) generating videos chunk-by-chunk instead of as a whole.
It excels in generating high-quality, temporally consistent videos from text or image prompts.
With support for large-scale model sizes and long context lengths, it is well-suited for a wide range of creative and generative video applications.
We collect this subset from TalkingHeadBench~\citep{bongini2025wild}.

\textbf{Flux1.1 Pro.}
Flux1.1 Pro is currently the most advanced model of Flux series, which is introduced by Black Forest Labs.
It is designed for fast, high-resolution and realistic text-to-image generation.
We collect this subset from WILD~\citep{bongini2025wild}, which is generated using template-like textual prompts.

\begin{table}[t]
\caption{Data list of the hierarchical evaluation protocol in HydraFake dataset.}
    \centering
    \scalebox{0.85}{
        \small
        \begin{tabular}{p{60pt}<{\raggedright}p{60pt}<{\centering}p{55pt}<{\centering}p{40pt}<{\centering}p{45pt}<{\centering}p{55pt}<{\centering}}
        \toprule[1pt]
        \textbf{\hspace{-5pt}Evaluation Split} & \textbf{Method} & \textbf{Sub-Type} & \textbf{Venue} & \textbf{Data Scale} & \textbf{Resolution} \\
        \shline
        \multirow{5}{*}{\hspace{-5pt}\textbf{In-Domain}} & FaceForensics++ & FS & ICCV'19 & 8,960 & $256\times 256$\\
        & Facevid2vid & FR & Arxiv'19 & 2,000 & $256\times 256$ \\
        & Hallo2 & FR & ICLR'25 & 1,660 & $256\times 256$ \\
        & StyleGAN & EFG & CVPR'19 & 600 & $1024\times 1024$ \\
        & Midjourney & EFG & None & 600 & $1024\times 1024$ \\
        \cline{1-6}
        \multirow{5}{*}{\hspace{-5pt}\textbf{Cross-Model}} & Adobe Firefly & Proprietary & None & 600 & $1024\times 1024$ \\
        & StarryAI & Proprietary & None & 600 & $1024\times 1024$ \\
        & Flux1.1 Pro & Customized & None & 600 & $1024\times 1024$ \\
        & MAGI-1 & Video AR & None & 1,048 & $256\times 256$ \\
        & HART & Image AR & Arxiv'24 & 4,200 & $1024\times 1024$ \\
        & Infinity & Image AR & CVPR'25 & 4,200 & $1024\times 1024$ \\
        \cline{1-6}
        \multirow{5}{*}{\hspace{-5pt}\textbf{Cross-Forgery}} & StarGANv2 & Editing & CVPR'20 & 2,000 & $256\times 256$ \\
        & CodeFormer & Restoration & NIPS'22 & 1,750 & $512\times 512$ \\
        & IC-Light & Relighting & ICLR'25 & 2,082 & $1536\times 1536$ \\
        & FaceAdapter & Generative FS & ECCV'24 & 300 & $1024\times 1024$ \\
        & PuLID & Personalization & NIPS'24 & 3,360 & $1024\times 1024$ \\
        & InfiniteYou & Personalization & ICCV'25 & 3,244 & $1024\times 1024$ \\
        \cline{1-6}
        \multirow{5}{*}{\hspace{-5pt}\textbf{Cross-Domain}} 
        & Hailuo AI & Commercial & None & 1,000 & $256\sim 1536$ \\
        & Dreamina & Social media & None & 952 & $1024\times 1024$ \\
        & GPT-4o & Social media & None & 630 & $159\sim1536$ \\
        & DeepFaceLab & Classic dataset & PR 2023 & 3,094 & $256\times 256$ \\
        & FFIW & Classic dataset & CVPR'21 & 6,832 & $256\times 256$ \\
        & InfiniteYou-CD & Personalization & ICCV'25 & 2,960 & $1024\times 1024$ \\
        \bottomrule[1pt]
        \end{tabular}
    }
    % \vspace{-0.25cm}
    \label{tab:data_statstics}
    % \vspace{-0.5cm}
\end{table}

\subsubsection{Cross-Forgery Evaluation}
\label{supp:dataset_test_cf}
The cross-forgery testing images come from deepfakes generated by unseen forgeries.
With the rapid development of generative techniques, novel types of forgery are constantly emerging, such as portrait relighting~\citep{zhang2025scaling} and IP-preserved personalization~\citep{jiang2025infiniteyou, guo2024pulid}.
To assess the model's generalization capacities when encountering these emerging deepfake methods, we collect $5$ representative forgery methods in our dataset, including face relighting~\citep{zhang2025scaling}, face restoration~\citep{zhou2022towards}, generative face swapping~\citep{han2024face}, facial attribute editing~\citep{choi2020stargan} and face personalization~\citep{jiang2025infiniteyou, guo2024pulid}.

\textbf{Face Relighting.}
The method is based on IC-Light~\citep{zhang2025scaling}, which is an emerging ability in the generative models and is becoming prevailing.
``IC-Light'' means ``Imposing Consistent Light'', which is capable of adjusting the lighting sources and intensity in the image while keeping the subject highly unchanged.
The condition is based on textual prompts.
We sampled real images from FFHQ~\citep{karras2019style} and reproduced IC-Light to change the lighting condition of these real images.
We implemented $10$ lighting types (e.g., ``sunshine from window'', ``soft studio lighting'' and ``neon light in city'') and $4$ lighting sources (i.e., ``left'', ``right'', ``top'' and ``bottom'').
We use multiple seeds for each condition, and then manually filter out those of low quality.

\textbf{Face Restoration.}
The method is based on CodeFormer~\citep{zhou2022towards}, which can recover low-quality (e.g., blurred) natural faces to high-quality counterparts, even when the inputs are severely degraded.
It can generate high-quality faces while maintaining the fidelity.
In fact, this is a helpful technique that has positive usage in many domains.
But considering this can be also used for low-quality deepfake images, we take this as an unseen forgery in our dataset.
Specifically, we sampled some low-quality fake images from DF40~\citep{yan2024df40} and TalkingHeadBench~\citep{xiong2025talkingheadbench}, and then employ CodeFormer to restore them into $512\times 512$ images.

\textbf{Facial Attribute Editing.}
Facial attribute editing is a common manipulation, involving altering facial attributes such as hairstyle and makeup.
In our dataset we leave this type out for testing.
We collect images generated by StarGANv2~\citep{choi2020stargan} from DF40~\citep{yan2024df40}.

\textbf{IP-preserved Face Personalization.}
IP-preserved face personalization technology enables the generation of synthetic faces that closely retain the distinctive visual attributes of original intellectual property (IP).
By producing highly realistic and IP-consistent deepfakes, it can facilitate unauthorized exploitation or impersonation of protected characters and personalities.
With the advancement of generative models, face personalization techniques are now capable of maintaining high-fidelity while following complex contextual and subject-specific instructions (e.g., transforming an ID photo into an image of the singer in the bar).
We reproduce two timely methods PuLID~\citep{guo2024pulid} and InfiniteYou~\citep{jiang2025infiniteyou}.
We sample real images from FFHQ as the source images.
To enhance the realism and \textit{semantic coherence} of face personalization, we employ Qwen2.5-VL-72B to generate customized prompts for each image.

\textbf{Generative Face Swapping.}
The face swapping data in existing datasets are often produced by conventional approaches such as graphics-based methods or GAN models.
While nowadays the generative-based methods are capable of generating high-fidelity swapped faces, which are based on Diffusion models.
Considering the latest methods such as DreamID~\citep{ye2025dreamid} and DynamicFace~\citep{wang2025dynamicface} are not open-sourced yet, we implemented FaceAdapter~\citep{han2024face}, which produces high-quality swapping data.
We will keep tracking the advancements in these methods and update our dataset.
The source faces are sampled from FFHQ.
We manually filter out those low-quality generated images, only maintaining high-fidelity samples.

\subsubsection{Cross-Domain Evaluation}
\label{supp:dataset_test_cd}
The ``domain'' in our dataset mainly refers to \textit{data source}.
For instance, the cross-forgery data are generated using in-domain real images from FFHQ, which alters the manipulation methods but keeps the data source unchanged.
But for cross-domain testing, the fake images are either generated from unseen real data srouce or entirely generated by commercial models.
And we also crawled fake images from social media, which serves as a challenging cross-domain evaluation.
Specifically, our cross-domain testing can be clustered into three types:
(1) classic datasets, including DeepFaceLab from DF40~\citep{yan2024df40} and FFIW~\citep{dolhansky2019deepfake} which is widely adopted in existing benchmarks.
(2) Reproduced deepfakes, including face personalization generated using real images from VFHQ~\citep{xie2022vfhq}.
(3) In-the-wild deepfakes, where we collect data from social media such as Xiaohongshu and TikTok.
We retrieved images through the tags of the posts and collected the images generated by GPT-4o~\citep{hurst2024gpt} and Dreamina~\citep{dreamina}, and cropped out the digital watermarks.
We further generate deepfake videos using Hailuo AI~\citep{hailuo}, and extract $8$ frames for each video.

\begin{figure*}[ht]
\centering
\begin{tcolorbox}[listing only, 
                  listing style={basicstyle=\ttfamily\footnotesize},
                  colback=gray!10, 
                  title=Valid Output Formats in P-GRPO]
Format 1 (Basic):\\
\verb|<fast>| ... \verb|</fast>|\\
\verb|<reasoning>| ... \verb|<reasoning>|\\
\verb|<conclusion>| ... \verb|<conclusion>|\\
\verb|<answer>| ... \verb|<answer>|\\
\\
Format 2 (With \textbf{Planning}):\\
\verb|<fast>| ... \verb|</fast>|\\
\verb|<planning>| ... \verb|<planning>|\\
\verb|<reasoning>| ... \verb|<reasoning>|\\
\verb|<conclusion>| ... \verb|<conclusion>|\\
\verb|<answer>| ... \verb|<answer>|\\
\\
Format 3 (With \textbf{Self-Reflection}):\\
\verb|<fast>| ... \verb|</fast>|\\
\verb|<reasoning>| ... \verb|<reasoning>|\\
\verb|<reflection>| ... \verb|<reflection>|\\
\verb|<conclusion>| ... \verb|<conclusion>|\\
\verb|<answer>| ... \verb|<answer>|\\
\\
Format 4 (With \textbf{Planning} and \textbf{Self-Reflection}):\\
\verb|<fast>| ... \verb|</fast>|\\
\verb|<planning>| ... \verb|<planning>|\\
\verb|<reasoning>| ... \verb|<reasoning>|\\
\verb|<reflection>| ... \verb|<reflection>|\\
\verb|<conclusion>| ... \verb|<conclusion>|\\
\verb|<answer>| ... \verb|<answer>|
\end{tcolorbox}

\caption{Valid output formats in $R_{fmt}$ of P-GRPO.}
\label{prompt:formats}
\end{figure*}

\subsection{More Implementation Details}
\label{supp:more_imple}
\textbf{Training resources.}
Our model is trained with $8$ PPUE GPUs based on ms-swift~\citep{zhao2024swiftascalablelightweightinfrastructure}.
The theoretical peak computational capacity (TFLOPS) of one PPUE GPU is roughly half of an NVIDIA A100 GPU, and each PPUE GPU has $96$GB VRAM.
With such infrastructure, the SFT and MiPO stage take $5.5$ hours and $2$ hours, respectively.
The P-GRPO stage takes $11$ hours on $9$K training samples.
All the inferences are conducted on a single PPUE GPU.

\textbf{Training details of previous methods.}
For previous SOTA methods, we reproduce them based on DeepfakeBench~\citep{yan2023deepfakebench}.
For F3Net~\citep{qian2020thinking}, UniFD~\citep{ojha2023towards}, IID~\citep{huang2023implicit}, FreqNet~\citep{tan2024frequency}, ProDet~\citep{cheng2024can}, NPR~\citep{tan2024rethinking} and Effort~\citep{yan2024orthogonal}, we reproduce them based on DeepfakeBench~\citep{yan2023deepfakebench}.
For AIDE~\citep{yan2024sanity}, Co-SPY~\citep{cheng2025co} and D$^3$~\citep{yang2025d}, we train the model with official codes and perform inference using DeepfakeBench.
The images are first randomly cropped into $256\times256$ and then resized to $224\times224$.
For Co-SPY~\citep{cheng2025co}, the images are resized to $384\times384$ following the official implementation.
We apply a series of data augmentations during training, including random flipping, rotation, gaussian blur, brightness and contrast alternation, color jitter and JPEG compression.
Following official guides, AIDE and D$^3$ are trained for $100$ epochs.
FreqNet and NPR are trained for $50$ epochs.
The first stage of Co-SPY (i.e., artifacts and semantic encoders) are trained for $20$ epochs, and the second stage (i.e., combination) is trained for another $10$ epochs.
Effort is trained for $10$ epochs and other methods are trained for $20$ epochs.
During testing, the images are resized to $224\times224$.
For all these methods, we curate a validation set containing 4K in-domain images for model selection.

\textbf{Other details.}
The training data from all stages are strictly sampled from HydraFake training set.
The $36$K SFT data are randomly sampled and balanced across forgery types.
The $3$K MiPO pairs are selected based on SFT models' outputs. $800$ images that the SFT model fails to reach all correct answers under $8$ rollouts are selected. Each image is paired with $4$ manually selected non-preference chains (from SFT model's outputs) and $1$ manually annotated preference chain, resulting in 3K samples for MiPO.
The $9$K P-GRPO data are randomly sampled and balanced across forgery types.
For open-sourced MLLMs, we provide prior knowledge and instruct the model to perform thinking in the prompts.
The full prompts are provided in Figure~\ref{prompt:zs_qwen}, Figure~\ref{prompt:zs_mimo} and Figure~\ref{prompt:zs_gpt}.
For Gemini-2.5-Pro, we enable thinking and searching.
$\lambda_1$ and $\lambda_2$ are set to $1.0$ and $0.25$, respectively.
The valid output formats for $R_{fmt}$ in P-GRPO are listed in Figure~\ref{prompt:formats}.

\subsection{More Discussions}
\label{supp:more_discussion}

\textbf{Difference between explainable and reasoning deepfake detection.}
In this part, we formulate different task settings of MLLM-based deepfake detection.
Given the input image $\bm{I}$, deepfake detection aims to determine its authenticity $\mathcal{Y}\in{\{0,1\}}$, where $1$ means the image is fake and vice versa.
Suppose the input image and query are collectively denoted as $\bm{q}$.
The sequential outputs of MLLM are denoted as
$\bm{s}=\{\bm{s}_1, \bm{s}_2, ..., \bm{s}_T\}$, where $T$ is sequence length.
The conditional probability of sequence $\bm{s}$ is written as $P(\bm{s}|\bm{q})=\prod_{t=1}^{T}P(\bm{s}_t|\bm{q},\bm{s}_{<t})$.

Recent works~\citep{chen2024x2, he2025vlforgery} utilize MLLM for explainable deepfake detection, where LLM first generates the answer $\bm{s}_{\mathcal{A}}$ (e.g., ``fake'' or ``this image is fake'').
Then a detailed explanation sequence $\bm{s}_{\mathcal{E}}$ is generated based on $\{\bm{q}, \bm{s}_{\mathcal{A}}\}$ (where $\mathcal{A}\ll \mathcal{E}$).
For simplicity, we suppose the answer ${\bm{s}_{\mathcal{A}}}$ is in a single word. 
The process can be decomposed into:
\begin{equation}
    P(\bm{s}|\bm{q})=\prod_{t=1}^{\mathcal{A}}P(\bm{s}_{t}|\bm{q})\cdot \prod_{t=\mathcal{A}+1}^{\mathcal{A}+\mathcal{E}}P(\bm{s}_{t}|\bm{q}, \bm{s}_{<\mathcal{A}+\mathcal{E}}),
\end{equation}
where the final decision (i.e., $\prod_{t=1}^{\mathcal{A}}P(\bm{s}_{t}|\bm{q})$) is solely conditioned on the input image.
This is fundamentally similar to the small vision models, where the distributional mapping $f: \bm{I}\rightarrow \mathcal{Y}$ is estimated directly within a single token, prone to overfitting.

In contrast, we formulate the deepfake detection as a reasoning task.
The MLLM first conduct holistic reasoning, denoted as $\bm{s}_{\mathcal{R}}$, and the final answer is then determined in $\bm{s}_{\mathcal{A}}$:
\begin{equation}
    P(\bm{s}|\bm{q})=\prod_{t=1}^{\mathcal{R}}P(\bm{s}_{t}|\bm{q}, \bm{s}_{<\mathcal{R}}) \cdot \prod_{t=\mathcal{R}+1}^{\mathcal{R}+\mathcal{A}}P(\bm{s}_{t}|\bm{q}, \bm{s}_{<\mathcal{R}+\mathcal{A}}),
\end{equation}
where the final answer (i.e., $\prod_{t=\mathcal{R}+1}^{\mathcal{R}+\mathcal{A}}P(\bm{s}_{t}|\bm{q}, \bm{s}_{<\mathcal{R}+\mathcal{A}})$) is building on inputs and reasoning process.
The mapping is altered into $f':\bm{I}\rightarrow \mathcal{R}\rightarrow \mathcal{Y}$, enabling more comprehensive and adaptive modeling.

\textbf{Large Reasoning Models.}
Large Language Models (LLMs) are inherently good reasoners for general tasks.
A simple prompt engineering could activate the reasoning behaviors of LLMs~\citep{wei2022chain, kojima2022large}, which is termed as Chain-of-Thought (CoT).
Building on the impressive capabilities of CoT, the community began to explore large-scale and structured reasoning, leading to powerful reasoning models~\citep{jaech2024openai, guo2025deepseek, xu2024llava} through tailored post-training.
For instance, DeepSeek-R1~\citep{guo2025deepseek} adopts reasoning cold-start followed by Reinforcement Learning with Verifiable Rewards (RLVR) to incentivize the general reasoning capabilities.
Inspired by the success of RLVR, recent works~\citep{liu2025visual, zhang2025r1} attempt to introduce pure rule-based RL into multimodal domain.
However, a recent study~\citep{yue2025does} points out that pure RLVR can not introduce novel abilities to base model.
We also empirically reveal the suboptimal performance achieved by pure RL.
Therefore, we first introduce a high-quality cold-start~\citep{liao2025longperceptualthoughts, chen2025advancing, team2025kimi} to internalize the thinking patterns to the base model.
Unlike general tasks that require diverse and flexible thinking patterns~\citep{zhan2025gthinker}, deepfake detection is a well-defined task.
Therefore, we establish a unified reasoning framework to facilitate efficient thinking.

\begin{figure*}[t]
    \centering
    \vspace{-0.5cm}
    \includegraphics[width=0.86\linewidth]{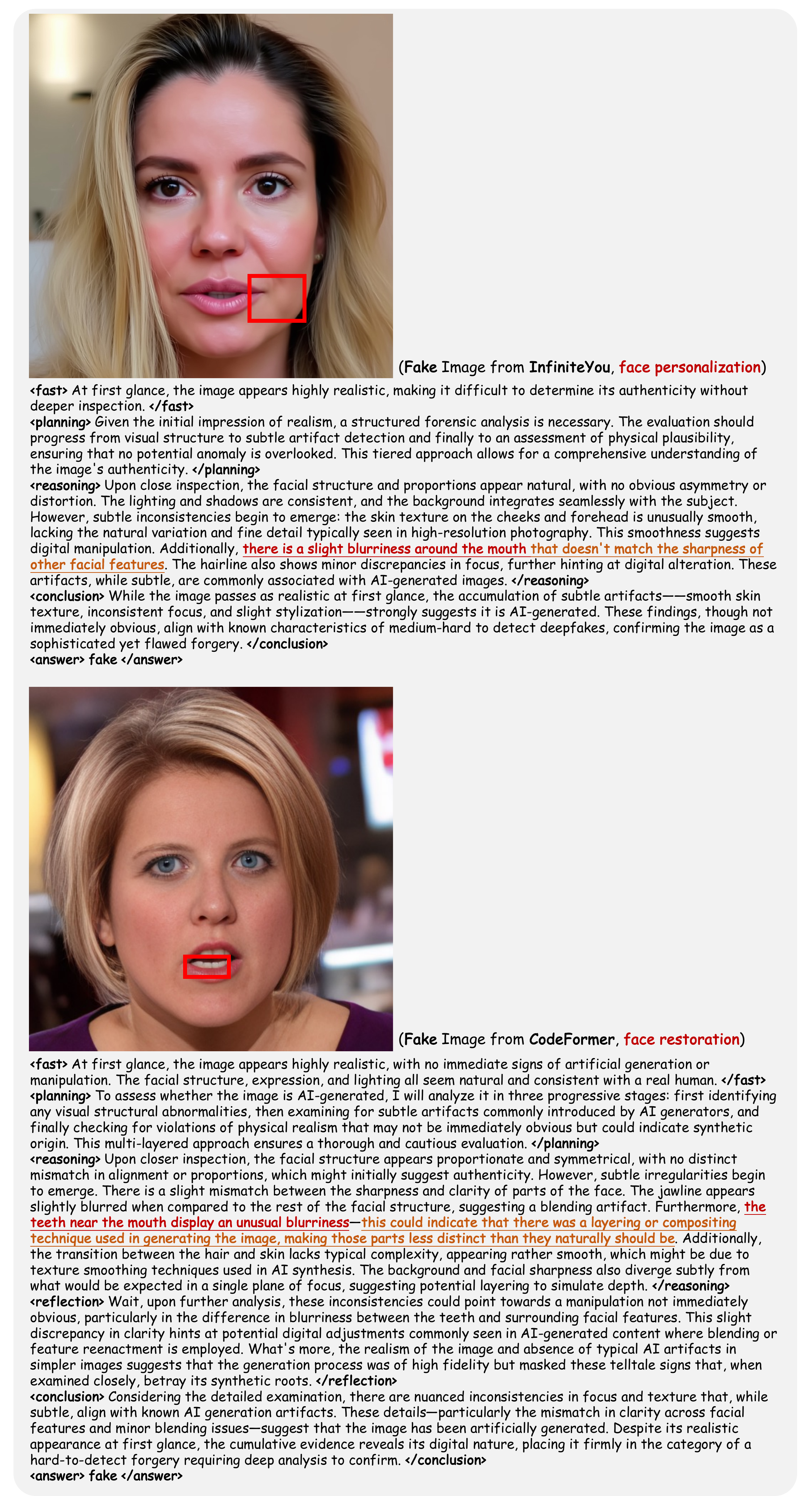}
     \caption{Illustration of model's capacity to perceive those barely noticeable artifacts. This shows certain advantages to human system.
     }
	\label{fig:deepfake_reason}
    \vspace{-0.2cm}
\end{figure*}

\textbf{Rationale for reasoning in deepfake detection.}
A possible concern is that even humans can fail on some highly realistic deepfakes.
Applying reasoning for deepfake detection also faces the same problem.
However, we point out that there is a physiological limit on human perception.
The \textbf{human} excels at high-level semantic reasoning, such as judging contextual plausibility, but is less equipped to detect subtle, low-level digital artifacts like subtle blurriness or unnatural texture patterns. 
In contrast, the \textbf{machine} can be trained to perceive these subtle artifacts with superhuman accuracy.
The primary challenge, which traditional detectors face, is not perception but generalization, i.e., they tend to overfit to specific artifact patterns. 
This is where the pattern-aware reasoning becomes crucial.
Our goal is not to mimic a human's intuitive guess, but to emulate a forensic expert's systematic investigation.
Our approach uniquely \textbf{combines} \textit{the machine's superhuman perception} with a \textit{structured and human-like reasoning framework} (e.g., planning, reasoning, self-reflection, conclusion).
As illustrated in Figure~\ref{fig:deepfake_reason}, the model can perceive subtle artifacts (e.g., barely noticeable blurriness and faint texture anomalies) that are nearly imperceptible to the human. 
Therefore, reasoning is crucial not to replicate the fallible human eye, but to provide a logical structure that effectively leverages the model's perceptual abilities for robust generalization.

\begin{figure*}[t]
    \centering
    \includegraphics[width=0.98\linewidth]{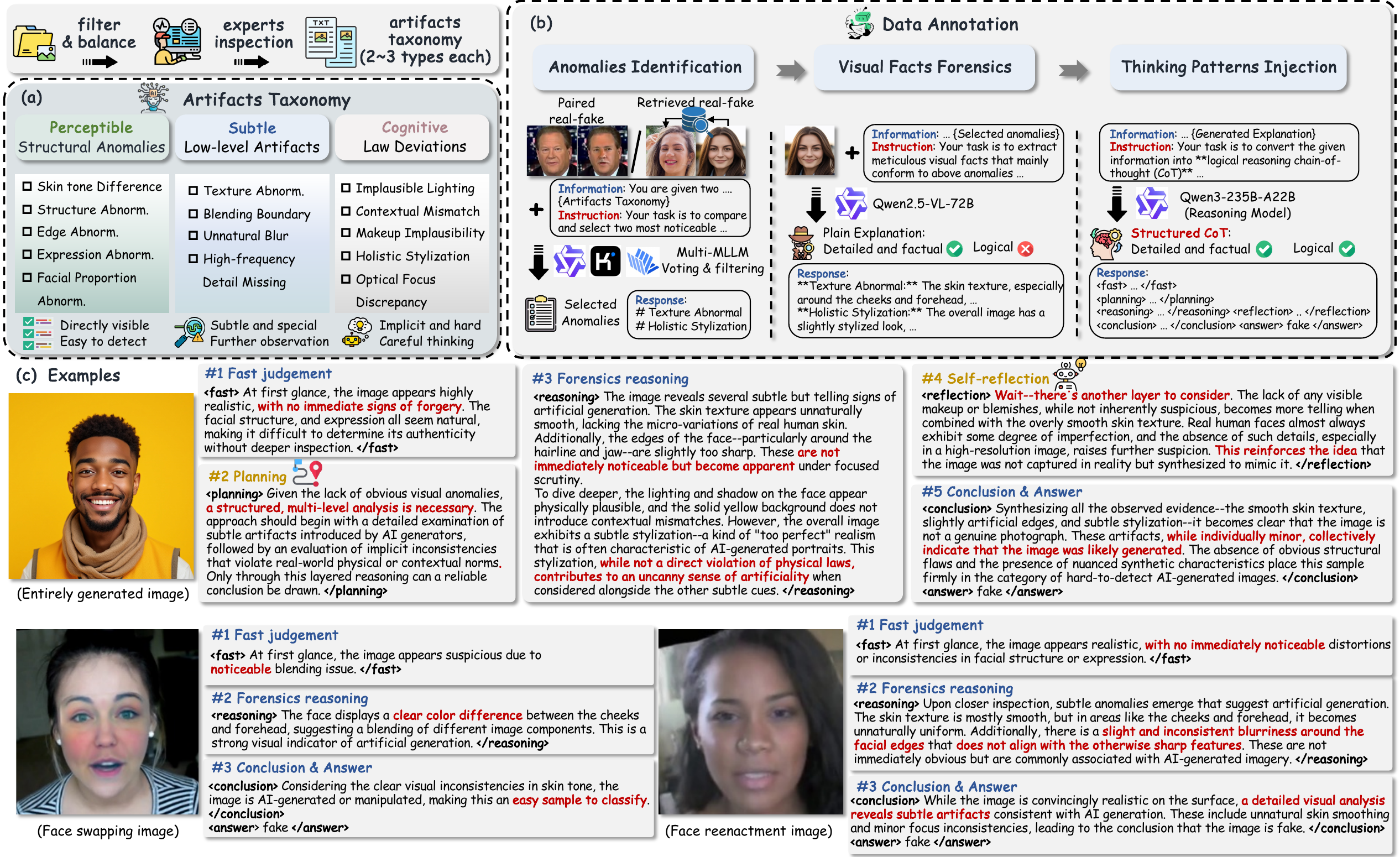}
     \caption{\textbf{Construction pipeline of Pattern-Aware SFT data}.
     \textbf{(a)} We first inspect a subset and summarize the artifacts into three clusters.
     \textbf{(b)} Then we introduce a multi-step strategy to generate pattern-aware reasoning data.
     \textbf{(c)} Annotated examples. The reasoning process evolves in complexity and depth (as highlighted in \textbf{\textcolor{red}{red}}), culminating in a final answer through synthesis of all evidence.
     }
	\label{fig:annotation}
    \vspace{-0.4cm}
\end{figure*}

\subsection{Multi-Step Annotation Pipeline of SFT data}
\label{supp:data_annotate}
As shown in Figure~\ref{fig:annotation}, for \textbf{fake images} we divide the annotation process into three steps:

\textbf{Step-\uppercase\expandafter{\romannumeral1}}: 
Based on human inspection, we found that most fake images exhibit $2$ to $3$ types of artifacts, hence we aim to find the two most prominent artifacts.
To reduce model bias, we employ an ensemble voting strategy, leveraging Qwen2.5-VL-72B~\citep{bai2025qwen2}, Kimi-VL-A3B-Thinking~\citep{team2025kimi} and InternVL3-78B~\citep{zhu2025internvl3} to sample 5 times individually.
Only the answers that receive more than 10 votes are selected, ensuring the reliability of the final results.

\textbf{Step-\uppercase\expandafter{\romannumeral2}}:
In this stage, we aim to extract visual details that conform to the identified artifacts.
We found that Qwen2.5-VL-72B performs better than GPT-4o~\citep{hurst2024gpt} here, capable of generating detailed and factual responses.
This yields concrete explanatory texts, like recent practice~\citep{wen2025spot, gao2025fakereasoning, zhang2025ivy}.
However, such \textit{plain} explanations lack human-like reasoning logic, which hinders the generalization to OOD samples.

\textbf{Step-\uppercase\expandafter{\romannumeral3}}:
To emulate human mindset, we further transform the above explanations into logical chains.
We define five ``thinking patterns'' and instruct the model to rewrite the explanations into different tags strictly based on the original meaning.
We observed that large reasoning models are inherently adept at generating highly logical content.
Therefore, we use Qwen3-235B-A22B~\citep{yang2025qwen3} for this step, yielding high-quality reasoning data.
Finally, the data undergo a filtering process, which involves rule-based filtering and balancing among different forgery types.

For \textbf{real images}, we only use the last two steps (without the need for anomalies detection).
For Step-\uppercase\expandafter{\romannumeral1} we provide the full artifacts list to Qwen2.5-VL-72B for comprehensive visual facts forensics.
For Step-\uppercase\expandafter{\romannumeral2} we adopt Qwen3-235B-A22B to convert the explanation texts into pattern-aware reasoning chain.
Specifically, for some low-quality real images, which may contain some misleading artifacts like unexpected blurriness or missing visual details, we instruct the model to point this out and put them in the ``self-reflection'' content.
However, the model is not capable of directly perceive such minor artifacts especially when told the image is authentic.
Hence, we provide a rough difficulty information based on dataset level and encourage the model to perform self-reflection on those difficult images.
This can mitigate the problem while it can not be fully addressed due to the significant loss of details in those low-resolution images.

\subsection{More Experimental Results}
\label{supp:more_results}

\subsubsection{More Results and Analyses on HydraFake}
\label{supp:more_res_hydrafake}
We provide full experimental results for all subsets.
To help better understanding the model performance, we report Precision and Recall for each subset, with fake being the positive label.

The in-domain results are shown in Table~\ref{tab:supp_id}.
Effort achieves the best results among previous detectors.
It is worth noting that, under the mixed training sources, previous methods even struggle on in-domain datasets, i.e., most methods achieve less than 90\% average performance.
This is mainly due to the degraded performance on low-resolution datasets such as FF++ and Facevid2vid.
We suppose this is a major deficiency in current deepfake detectors, which tend to bias towards image resolutions.
Our model achieves better performance, achieving over 99.5\% on those high resolution images, while the results on low resolution subsets still have room for improvement.

\begin{table}[t]
\caption{Performance comparison on the \textbf{In-Domain (ID)} subset of HydraFake dataset. The best results are \textbf{bolded} and the second best are \underline{underlined}. We report Accuracy (Acc.), Precision (P.) and Recall (R.) and the averaged results (Avg.) are reported in Accuracy.}
    \centering
    \renewcommand\arraystretch{1.1}
    \scalebox{0.76}{
        \small
        \begin{tabular}{p{75pt}<{\raggedright}p{14pt}<{\centering}p{14pt}<{\centering}p{14pt}<{\centering}p{14pt}<{\centering}p{14pt}<{\centering}p{14pt}<{\centering}p{14pt}<{\centering}p{14pt}<{\centering}p{14pt}<{\centering}p{14pt}<{\centering}p{14pt}<{\centering}p{14pt}<{\centering}p{14pt}<{\centering}p{14pt}<{\centering}p{14pt}<{\centering}p{14pt}<{\centering}}
        \toprule[1pt]
        \multirow{2}{*}{\hspace{-5pt}\textbf{Method}} & \multicolumn{3}{c}{\textbf{FaceForensics++}} & \multicolumn{3}{c}{\textbf{Facevid2vid}} & \multicolumn{3}{c}{\textbf{Hallo2}} & \multicolumn{3}{c}{\textbf{StyleGAN}} & \multicolumn{3}{c}{\textbf{Midjourney}} & \multirow{2}{*}{\textbf{Avg.}} \\
        \cmidrule(lr){2-4} \cmidrule(lr){5-7} \cmidrule(lr){8-10} \cmidrule(lr){11-13} \cmidrule(lr){14-16}
        & Acc. & P. & R. & Acc. & P. & R. & Acc. & P. & R. & Acc. & P. & R. & Acc. & P. & R. & \\
        \shline
        \hspace{-5pt}\rule{0pt}{7pt}F3Net (\textit{ECCV'20}) & 84.5 & 80.5 & 90.9 & 89.6 & 83.6 & 98.6 & 85.0 & 78.3 & 97.0 & 82.5 & 77.4 & 91.6 & 84.8 & 78.6 & 95.6 & 85.3\\
        \hspace{-5pt}UniFD (\textit{CVPR'23}) & 73.8 & 80.4 & 62.8 & 81.7 & 83.2 & 79.6 & 77.3 & 88.8 & 62.5 & 94.3 & 91.5 & 97.6 & 86.3 & 90.6 & 81.0 & 82.7 \\
        \hspace{-5pt}IID (\textit{CVPR'23}) & 83.8 & 86.6 & 79.9 & 92.8 & 89.5 & 97.0 & 79.1 & 72.2 & 94.6 & 80.7 & 72.1 & 100.0 & 80.5 & 71.9 & 100.0 & 83.4 \\
        \hspace{-5pt}FreqNet (\textit{AAAI'24}) & 52.2 & 51.2 & 94.4 & 54.5 & 52.3 & 100.0 & 71.2 & 66.7 & 84.5 & 77.5 & 69.8 & 96.6 & 78.8 & 70.8 & 98.0 & 66.8 \\
        \hspace{-5pt}ProDet (\textit{NIPS'24}) & 84.8 & 82.2 & 88.8 & 90.9 & 84.6 & 100.0 & 91.4 & 89.0 & 94.4 & 92.5 & 88.7 & 97.3 & 93.0 & 88.4 & 99.0 & 90.5 \\
        \hspace{-5pt}NPR (\textit{CVPR'24}) & 59.6 & 56.5 & 84.1 & 66.6 & 60.5 & 95.7 & 74.3 & 87.2 & 57.0 & 88.8 & 91.7 & 85.3 & 88.7 & 92.9 & 83.6 & 75.6 \\
        \hspace{-5pt}AIDE (\textit{ICLR'25}) & 58.9 & 58.7 & 59.6 & 78.6 & 70.3 & 99.0 & 84.5 & 92.4 & 75.2 & 86.8 & 93.1 & 99.0 & 93.0 & 92.7 & 93.3 & 80.4 \\
        \hspace{-5pt}Co-SPY (\textit{CVPR'25}) & 71.3 & 76.0 & 62.3 & 89.3 & 82.9 & 99.1 & 82.1 & 91.6 & 70.7 & 95.0 & 92.2 & 98.3 & 94.0 & 94.0 & 94.0 & 86.3 \\
        \hspace{-5pt}D$^3$ (\textit{CVPR'25}) & 72.5 & 70.3 & 77.8 & 82.1 & 74.7 & 96.9 & 91.1 & 91.7 & 90.3 & 96.3 & 94.2 & 98.6 & 94.7 & 91.3 & 98.6 & 87.3 \\
        \hspace{-5pt}Effort (\textit{ICML'25}) & 92.9 & 94.4 & 91.2 & 96.4 & 94.6 & 98.4 & 95.9 & 95.3 & 96.5 & 95.0 & 97.5 & 92.3 & 93.5 & 96.1 & 90.6 & \underline{94.7} \\
        \shline
        \hspace{-5pt}Qwen2.5-VL-7B & 51.3 & 93.5 & 2.8 & 51.5 & 94.3 & 3.3 & 50.0 & 50.0 & 1.1 & 51.5 & 80.0 & 4.0 & 51.5 & 76.4 & 4.3 & 51.2 \\
        \hspace{-5pt}InternVL3-8B & 55.9 & 55.5 & 59.0 & 53.7 & 53.6 & 54.6 & 56.3 & 83.8 & 15.6 & 52.0 & 66.6 & 8.0 & 52.3 & 69.4 & 8.3 & 54.0 \\
        \hspace{-5pt}MiMo-VL-7B & 56.2 & 55.3 & 64.0 & 62.2 & 59.4 & 77.1 & 60.4 & 63.9 & 47.6 & 66.4 & 69.3 & 58.6 & 73.6 & 72.0 & 77.3 & 63.8 \\
        \hspace{-5pt}GLM-4.1V-9BThink & 58.0 & 64.0 & 36.3 & 59.6 & 65.6 & 40.0 & 54.2 & 89.6 & 9.4 & 56.0 & 87.5 & 14.0 & 54.3 & 80.9 & 11.3 & 56.4 \\
        \hspace{-5pt}GPT-4o & 49.2 & 20.0 & 0.5 & 53.0 & 84.2 & 8.0 & 49.8 & 33.3 & 0.5 & 63.2 & 100.0 & 26.5 & 52.5 & 91.6 & 5.5 & 53.5 \\
        \hspace{-5pt}Gemini-2.5-Pro & 62.0 & 77.9 & 33.5 & 53.2 & 66.6 & 13.0 & 66.0 & 80.9 & 42.5 & 93.9 & 89.3 & 100.0 & 85.8 & 73.7 & 76.6 & 72.2 \\
        \rowcolor{Gray}\hspace{-7pt} \veritas $\;$(\textbf{ours}) & 90.9 & 90.2 & 91.7 & 96.1 & 92.7 & 100.0 & 99.8 & 99.6 & 100.0 & 99.8 & 99.6 & 100.0 & 100.0 & 100.0 & 100.0 & \textbf{97.3} \\
        \bottomrule[1pt]
        \end{tabular}
    }
    % \vspace{-0.25cm}
    \label{tab:supp_id}
    % \vspace{-0.5cm}
\end{table}

\begin{table}[t]
\caption{Performance comparison on the \textbf{Cross-Model (CM)} subset of HydraFake dataset. The best results are \textbf{bolded} and the second best are \underline{underlined}. We report Accuracy (Acc.), Precision (P.) and Recall (R.) and the averaged results (Avg.) are reported in Accuracy.}
    \centering
    \renewcommand\arraystretch{1.1}
    \scalebox{0.72}{
        \small
        \begin{tabular}{p{72pt}<{\raggedright}p{12pt}<{\centering}p{12pt}<{\centering}p{12pt}<{\centering}p{12pt}<{\centering}p{12pt}<{\centering}p{14pt}<{\centering}p{12pt}<{\centering}p{14pt}<{\centering}p{12pt}<{\centering}p{12pt}<{\centering}p{12pt}<{\centering}p{14pt}<{\centering}p{12pt}<{\centering}p{12pt}<{\centering}p{14pt}<{\centering}p{12pt}<{\centering}p{12pt}<{\centering}p{12pt}<{\centering}p{14pt}<{\centering}}
        \toprule[1pt]
        \multirow{2}{*}{\hspace{-5pt}\textbf{Method}} & \multicolumn{3}{c}{\textbf{Adobe Firefly}} & \multicolumn{3}{c}{\textbf{FLUX1.1Pro}} & \multicolumn{3}{c}{\textbf{StarryAI}} & \multicolumn{3}{c}{\textbf{MAGI-1}} & \multicolumn{3}{c}{\textbf{HART (VAR)}} & \multicolumn{3}{c}{\textbf{Infinity (VAR)}} & \multirow{2}{*}{\textbf{Avg.}} \\
        \cmidrule(lr){2-4} \cmidrule(lr){5-7} \cmidrule(lr){8-10} \cmidrule(lr){11-13} \cmidrule(lr){14-16} \cmidrule(lr){17-19}
        & Acc. & P. & R. & Acc. & P. & R. & Acc. & P. & R. & Acc. & P. & R. & Acc. & P. & R. & Acc. & P. & R. & \\
        \shline
        \hspace{-5pt}\rule{0pt}{7pt}F3Net (\textit{ECCV'20}) & 86.7 & 80.0 & 97.7 & 87.8 & 80.4 & 100.0 & 78.6 & 76.2 & 83.3 & 85.0 & 78.6 & 96.2 & 86.0 & 78.6 & 99.0 & 82.9 & 78.6 & 90.4 & 84.5 \\
        \hspace{-5pt}UniFD (\textit{CVPR'23}) & 90.7 & 92.9 & 88.0 & 93.8 & 91.7 & 96.3 & 82.5 & 88.8 & 74.3 & 73.0 & 84.7 & 56.1 & 94.4 & 92.2 & 97.0 & 90.7 & 92.8 & 88.2 & 87.5 \\
        \hspace{-5pt}IID (\textit{CVPR'23}) & 83.3 & 75.0 & 100.0 & 82.8 & 74.4 & 100.0 & 80.0 & 72.1 & 97.6 & 80.2 & 72.6 & 96.5 & 81.1 & 72.6 & 100.0 & 82.2 & 73.7 & 100.0 & 81.6 \\
        \hspace{-5pt}FreqNet (\textit{AAAI'24}) & 60.3 & 59.5 & 64.3 & 76.7 & 68.4 & 99.0 & 59.0 & 59.1 & 58.3 & 69.2 & 65.5 & 81.1 & 77.1 & 69.4 & 96.7 & 75.1 & 68.7 & 92.3 & 69.6 \\
        \hspace{-5pt}ProDet (\textit{NIPS'24}) & 92.6 & 88.3 & 98.3 & 94.2 & 89.5 & 100.0 & 88.2 & 88.0 & 88.3 & 91.9 & 87.5 & 97.7 & 93.8 & 89.0 & 99.8 & 93.1 & 88.7 & 98.6 & 92.3 \\
        \hspace{-5pt}NPR (\textit{CVPR'24}) & 68.8 & 86.0 & 45.0 & 91.2 & 91.5 & 90.6 & 59.5 & 78.8 & 26.0 & 82.6 & 91.9 & 71.5 & 91.3 & 93.1 & 89.3 & 84.0 & 92.6 & 73.8 & 79.6 \\
        \hspace{-5pt}AIDE (\textit{ICLR'25}) & 68.8 & 85.5 & 43.3 & 86.3 & 91.0 & 80.6 & 64.0 & 86.6 & 33.0 & 88.9 & 93.6 & 83.6 & 95.4 & 94.3 & 96.6 & 76.0 & 91.0 & 57.7 & 79.9 \\
        \hspace{-5pt}Co-SPY (\textit{CVPR'25}) & 93.5 & 93.6 & 93.3 & 95.5 & 94.1 & 97.0 & 85.3 & 93.4 & 76.0 & 93.3 & 92.2 & 72.7 & 96.6 & 94.5 & 98.9 & 95.3 & 94.1 & 96.7 & 93.3 \\
        \hspace{-5pt}D$^3$ (\textit{CVPR'25}) & 93.6 & 90.3 & 97.7 & 95.6 & 92.3 & 99.6 & 91.3 & 92.2 & 90.3 & 90.7 & 90.5 & 91.0 & 95.8 & 92.2 & 99.9 & 95.5 & 93.1 & 98.3 & \underline{93.8} \\
        \hspace{-5pt}Effort (\textit{ICML'25}) & 82.8 & 97.1 & 67.6 & 96.5 & 96.0 & 97.0 & 78.0 & 94.2 & 59.6 & 90.5 & 96.5 & 84.1 & 97.8 & 97.2 & 98.5 & 98.3 & 97.3 & 99.4 & 90.7 \\
        \shline
        \hspace{-5pt}Qwen2.5-VL-7B & 50.0 & 50.0 & 1.3 & 50.0 & 50.0 & 1.3 & 49.7 & 0.0 & 0.0 & 50.0 & 50.0 & 0.9 & 52.0 & 83.8 & 4.9 & 52.9 & 87.3 & 6.8 & 50.8 \\
        \hspace{-5pt}InternVL3-8B & 54.0 & 74.0 & 12.3 & 49.8 & 47.3 & 3.0 & 49.0 & 28.5 & 1.3 & 56.6 & 83.5 & 16.4 & 55.8 & 83.1 & 14.5 & 57.2 & 84.5 & 17.6 & 53.7 \\
        \hspace{-5pt}MiMo-VL-7B & 74.5 & 74.7 & 74.0 & 77.1 & 76.7 & 78.0 & 82.5 & 68.0 & 55.3 & 60.3 & 64.0 & 47.1 & 82.4 & 77.9 & 90.6 & 81.4 & 77.2 & 89.2 & 76.4 \\
        \hspace{-5pt}GLM-4.1V-9BThink & 55.2 & 82.9 & 13.0 & 52.3 & 76.9 & 6.6 & 50.5 & 66.6 & 2.0 & 51.6 & 81.5 & 4.2 & 68.4 & 96.1 & 38.2 & 60.7 & 95.5 & 22.5 & 56.5 \\
        \hspace{-5pt}GPT-4o & 57.7 & 96.9 & 16.0 & 52.0 & 83.3 & 5.0 & 51.4 & 85.7 & 3.0 & 59.9 & 80.0 & 26.4 & 81.2 & 92.5 & 67.9 & 54.8 & 87.5 & 10.6 & 59.5 \\
        \hspace{-5pt}Gemini-2.5-Pro & 64.9 & 78.8 & 41.2 & 92.4 & 90.3 & 94.9 & 82.8 & 92.3 & 72.0 & 62.5 & 79.0 & 34.1 & 93.4 & 88.5 & 100.0 & 93.2 & 89.1 & 98.5 & 81.5 \\
        \rowcolor{Gray}\hspace{-7pt} \veritas $\;$(\textbf{ours}) & 94.8 & 99.6 & 89.9 & 99.8 & 99.6 & 100.0 & 97.0 & 100.0 & 94.0 & 99.9 & 99.8 & 100.0 & 99.9 & 99.8 & 100.0 & 99.9 & 99.8 & 99.9 & \textbf{98.6} \\
        \bottomrule[1pt]
        \end{tabular}
    }
    % \vspace{-0.25cm}
    \label{tab:supp_cm}
    % \vspace{-0.5cm}
\end{table}

The cross-model results are shown in Table~\ref{tab:supp_cm}.
D$^{3}$ achieves the best results among previous detectors.
Lots of previous methods achieve good performance on cross-model scenarios, with averaged performance greater than 90\%, such as D$^{3}$, ProDet, Co-SPY and Effort.
These models show great performance on cross-model data, especially on VAR architectures, achieving over 95\% accuracy.
Our model demonstrates extraordinary generalization performance on cross-model data, achieving almost 99\% accuracy, while the recall capacity on proprietary models (e.g., Adobe Firefly and StarryAI) still has room for improvement.

The cross-forgery results are shown in Table~\ref{tab:supp_cf}.
Effort achieves the best performance among previous detectors.
The performance of most detectors are limited.
For instance, those detectors tailored for deepfake facial images (i.e., IID and ProDet), showing extremely limited recall capacities when encountering facial attribute editing and face relighting.
Most methods achieve moderate performance on face restoration and personalization.
These results verify that current detectors exhibit limited abilities to generalize unseen forgeries.
Besides, it is worth noting that Effort achieve excellent performance on face relighting, greatly surpassing our method.
We suppose the reason is that Effort freezes CLIP’s semantic encoder, allowing the model to focus solely on detecting whether an image has been manipulated, which is critical in detecting relighting where the identities and other semantics are largely unchanged.

\begin{table}[t]
\caption{Performance comparison on the \textbf{Cross-Forgery (CF)} subset of HydraFake dataset. The best results are \textbf{bolded} and the second best are \underline{underlined}. We report Accuracy (Acc.), Precision (P.) and Recall (R.) and the averaged results (Avg.) are reported in Accuracy.}
    \centering
    \renewcommand\arraystretch{1.1}
    \scalebox{0.72}{
        \small
        \begin{tabular}{p{72pt}<{\raggedright}p{12pt}<{\centering}p{12pt}<{\centering}p{12pt}<{\centering}p{12pt}<{\centering}p{12pt}<{\centering}p{12pt}<{\centering}p{12pt}<{\centering}p{12pt}<{\centering}p{12pt}<{\centering}p{12pt}<{\centering}p{12pt}<{\centering}p{12pt}<{\centering}p{12pt}<{\centering}p{12pt}<{\centering}p{12pt}<{\centering}p{12pt}<{\centering}p{12pt}<{\centering}p{12pt}<{\centering}p{14pt}<{\centering}}
        \toprule[1pt]
        \multirow{2}{*}{\hspace{-5pt}\textbf{Method}} & \multicolumn{3}{c}{\textbf{StarGANv2}} & \multicolumn{3}{c}{\textbf{IC-Light}} & \multicolumn{3}{c}{\textbf{CodeFormer}} & \multicolumn{3}{c}{\textbf{InfiniteYou}} & \multicolumn{3}{c}{\textbf{PuLID}} & \multicolumn{3}{c}{\textbf{FaceAdapter}} & \multirow{2}{*}{\textbf{Avg.}} \\
        \cmidrule(lr){2-4} \cmidrule(lr){5-7} \cmidrule(lr){8-10} \cmidrule(lr){11-13} \cmidrule(lr){14-16} \cmidrule(lr){17-19}
        & Acc. & P. & R. & Acc. & P. & R. & Acc. & P. & R. & Acc. & P. & R. & Acc. & P. & R. & Acc. & P. & R. & \\
        \shline
        \hspace{-5pt}\rule{0pt}{7pt}F3Net (\textit{ECCV'20}) & 41.3 & 20.7 & 6.2 & 48.9 & 47.9 & 24.8 & 71.9 & 72.8 & 69.8 & 84.9 & 78.2 & 96.8 & 85.5 & 79.5 & 95.7 & 72.6 & 77.5 & 62.8 & 67.5 \\
        \hspace{-5pt}UniFD (\textit{CVPR'23}) & 61.8 & 81.7 & 30.4 & 81.9 & 89.1 & 72.6 & 75.4 & 88.0 & 58.7 & 73.7 & 87.0 & 55.6 & 68.1 & 83.6 & 45.1 & 81.3 & 90.3 & 69.6 & 73.7 \\
        \hspace{-5pt}IID (\textit{CVPR'23}) & 41.4 & 32.8 & 16.6 & 53.3 & 54.3 & 41.0 & 79.7 & 72.9 & 94.4 & 81.8 & 73.4 & 99.6 & 81.8 & 73.3 & 99.9 & 73.7 & 68.2 & 87.1 & 68.6 \\
        \hspace{-5pt}FreqNet (\textit{AAAI'24}) & 33.1 & 11.4 & 5.0 & 73.1 & 67.6 & 88.9 & 70.3 & 66.6 & 81.5 & 72.8 & 67.4 & 88.3 & 77.4 & 69.9 & 96.2 & 67.7 & 64.9 & 75.0 & 65.7 \\
        \hspace{-5pt}ProDet (\textit{NIPS'24}) & 56.3 & 67.5 & 24.5 & 58.6 & 69.8 & 34.4 & 80.8 & 84.7 & 75.3 & 88.1 & 89.9 & 85.8 & 91.0 & 88.1 & 94.9 & 83.3 & 86.0 & 79.0 & 76.4 \\
        \hspace{-5pt}NPR (\textit{CVPR'24}) & 47.7 & 23.6 & 2.1 & 67.8 & 85.2 & 43.2 & 60.6 & 78.3 & 29.3 & 79.8 & 99.1 & 66.0 & 89.0 & 92.8 & 84.5 & 67.7 & 84.9 & 41.9 & 68.8 \\
        \hspace{-5pt}AIDE (\textit{ICLR'25}) & 56.7 & 74.6 & 20.3 & 79.2 & 91.5 & 64.3 & 86.1 & 90.8 & 80.2 & 74.2 & 89.4 & 55.0 & 62.4 & 83.9 & 30.5 & 75.7 & 90.3 & 56.7 & 72.4 \\
        \hspace{-5pt}Co-SPY (\textit{CVPR'25}) & 77.0 & 91.1 & 59.9 & 92.5 & 93.8 & 91.0 & 88.6 & 90.3 & 86.5 & 90.6 & 93.2 & 87.5 & 79.1 & 89.9 & 65.5 & 87.3 & 88.7 & 85.4 & 85.9 \\
        \hspace{-5pt}D$^3$ (\textit{CVPR'25}) & 62.4 & 82.4 & 31.5 & 71.6 & 86.3 & 51.4 & 82.9 & 90.9 & 73.1 & 80.0 & 88.7 & 68.7 & 82.4 & 89.8 & 73.1 & 73.7 & 84.1 & 57.4 & 75.5 \\
        \hspace{-5pt}Effort (\textit{ICML'25}) & 64.7 & 93.2 & 31.7 & 94.8 & 96.3 & 93.2 & 89.7 & 97.1 & 81.9 & 89.5 & 96.0 & 82.3 & 92.9 & 96.2 & 89.2 & 88.0 & 95.0 & 80.2 & \underline{86.6} \\
        \shline
        \hspace{-5pt}Qwen2.5-VL-7B & 50.5 & 64.7 & 2.2 & 56.7 & 90.8 & 15.2 & 50.7 & 70.0 & 2.4 & 53.6 & 90.3 & 8.1 & 54.5 & 90.4 & 10.1 & 51.6 & 63.6 & 4.7 & 52.9 \\
        \hspace{-5pt}InternVL3-8B & 62.9 & 87.7 & 30.0 & 54.2 & 77.1 & 12.0 & 62.9 & 86.9 & 30.5 & 63.6 & 92.1 & 29.7 & 54.8 & 80.1 & 12.7 & 67.7 & 96.3 & 35.8 & 61.0 \\
        \hspace{-5pt}MiMo-VL-7B & 48.7 & 47.7 & 26.2 & 82.6 & 76.4 & 94.3 & 76.4 & 75.1 & 78.9 & 79.7 & 76.3 & 85.7 & 78.4 & 75.8 & 83.5 & 82.8 & 79.1 & 88.3 & 74.8 \\
        \hspace{-5pt}GLM-4.1V-9BThink & 54.3 & 89.8 & 9.7 & 68.4 & 94.7 & 39.0 & 63.3 & 95.7 & 27.8 & 65.7 & 96.5 & 32.6 & 55.1 & 87.4 & 11.9 & 81.0 & 97.9 & 62.8 & 64.6 \\
        \hspace{-5pt}GPT-4o & 66.4 & 82.6 & 41.5 & 58.9 & 100.0 & 18.0 & 52.5 & 91.6 & 5.5 & 64.4 & 100.0 & 29.0 & 60.9 & 94.0 & 23.5 & 55.5 & 100.0 & 10.1 & 59.8 \\
        \hspace{-5pt}Gemini-2.5-Pro & 73.7 & 89.7 & 53.5 & 83.3 & 89.7 & 75.1 & 87.4 & 86.0 & 89.3 & 85.5 & 86.2 & 84.5 & 84.7 & 88.4 & 80.0 & 85.6 & 85.3 & 85.8 & 83.4 \\
        \rowcolor{Gray}\hspace{-7pt} \veritas $\;$(\textbf{ours}) & 90.3 & 99.5 & 81.0 & 75.7 & 99.5 & 51.5 & 97.0 & 98.6 & 95.3 & 91.8 & 98.9 & 84.5 & 95.1 & 99.5 & 90.6 & 91.7 & 98.6 & 84.6 & \textbf{90.3} \\
        \bottomrule[1pt]
        \end{tabular}
    }
    % \vspace{-0.25cm}
    \label{tab:supp_cf}
    % \vspace{-0.5cm}
\end{table}

\begin{table}[t]
\caption{Performance comparison on the \textbf{Cross-Domain (CD)} subset of HydraFake dataset. The best results are \textbf{bolded} and the second best are \underline{underlined}. We report Accuracy (Acc.), Precision (P.) and Recall (R.) and the averaged results (Avg.) are reported in Accuracy.}
    \centering
    \renewcommand\arraystretch{1.1}
    \scalebox{0.72}{
        \small
        \begin{tabular}{p{72pt}<{\raggedright}p{12pt}<{\centering}p{12pt}<{\centering}p{14pt}<{\centering}p{12pt}<{\centering}p{12pt}<{\centering}p{12pt}<{\centering}p{12pt}<{\centering}p{12pt}<{\centering}p{12pt}<{\centering}p{12pt}<{\centering}p{12pt}<{\centering}p{12pt}<{\centering}p{12pt}<{\centering}p{12pt}<{\centering}p{12pt}<{\centering}p{12pt}<{\centering}p{12pt}<{\centering}p{12pt}<{\centering}p{14pt}<{\centering}}
        \toprule[1pt]
        \multirow{2}{*}{\hspace{-5pt}\textbf{Method}} & \multicolumn{3}{c}{\textbf{DeepFaceLab}} & \multicolumn{3}{c}{\textbf{InfiniteYou-CD}} & \multicolumn{3}{c}{\textbf{Dreamina}} & \multicolumn{3}{c}{\textbf{Hailuo AI}} & \multicolumn{3}{c}{\textbf{GPT-4o}} & \multicolumn{3}{c}{\textbf{FFIW}} & \multirow{2}{*}{\textbf{Avg.}} \\
        \cmidrule(lr){2-4} \cmidrule(lr){5-7} \cmidrule(lr){8-10} \cmidrule(lr){11-13} \cmidrule(lr){14-16} \cmidrule(lr){17-19}
        & Acc. & P. & R. & Acc. & P. & R. & Acc. & P. & R. & Acc. & P. & R. & Acc. & P. & R. & Acc. & P. & R. & \\
        \shline
        \hspace{-5pt}\rule{0pt}{7pt}F3Net (\textit{ECCV'20}) & 57.7 & 54.7 & 89.1 & 78.5 & 72.6 & 91.6 & 55.6 & 55.0 & 60.9 & 68.6 & 63.3 & 88.6 & 66.2 & 63.6 & 75.5 & 66.4 & 64.8 & 71.4 & 65.5 \\
        \hspace{-5pt}UniFD (\textit{CVPR'23}) & 67.4 & 64.1 & 79.2 & 67.3 & 75.9 & 50.5 & 80.5 & 77.2 & 86.3 & 75.2 & 75.4 & 74.8 & 73.3 & 73.6 & 72.7 & 67.5 & 65.4 & 74.5 & 71.9 \\
        \hspace{-5pt}IID (\textit{CVPR'23}) & 65.2 & 65.1 & 65.5 & 69.9 & 63.0 & 95.9 & 63.8 & 58.0 & 99.6 & 63.3 & 57.6 & 100.0 & 63.8 & 58.2 & 97.8 & 64.2 & 66.6 & 56.6 & 65.0 \\
        \hspace{-5pt}FreqNet (\textit{AAAI'24}) & 50.6 & 50.3 & 98.7 & 67.0 & 62.3 & 85.8 & 62.1 & 58.4 & 83.4 & 59.3 & 56.9 & 76.8 & 58.3 & 56.3 & 73.3 & 51.2 & 50.6 & 95.0 & 58.1 \\
        \hspace{-5pt}ProDet (\textit{NIPS'24}) & 58.1 & 54.6 & 95.3 & 82.9 & 81.3 & 85.4 & 71.3 & 71.3 & 71.4 & 75.6 & 71.2 & 86.0 & 66.3 & 69.9 & 57.4 & 74.1 & 73.0 & 76.5 & 71.4 \\
        \hspace{-5pt}NPR (\textit{CVPR'24}) & 52.6 & 52.4 & 56.6 & 73.0 & 82.1 & 58.7 & 76.6 & 80.9 & 69.5 & 62.3 & 70.7 & 42.0 & 50.2 & 50.4 & 20.9 & 46.0 & 46.3 & 50.9 & 60.1 \\
        \hspace{-5pt}AIDE (\textit{ICLR'25}) & 59.7 & 57.5 & 73.9 & 67.9 & 75.2 & 53.2 & 49.7 & 49.4 & 25.2 & 58.0 & 60.6 & 45.8 & 51.9 & 53.0 & 33.3 & 59.2 & 57.3 & 71.2 & 57.7 \\
        \hspace{-5pt}Co-SPY (\textit{CVPR'25}) & 67.6 & 63.8 & 81.2 & 80.0 & 85.1 & 72.7 & 82.5 & 80.7 & 85.3 & 74.0 & 75.7 & 70.6 & 79.5 & 78.7 & 80.9 & 64.3 & 64.2 & 64.7 & \underline{74.7} \\
        \hspace{-5pt}D$^3$ (\textit{CVPR'25}) & 69.7 & 70.0 & 69.1 & 74.6 & 78.8 & 67.2 & 78.1 & 72.2 & 91.4 & 70.9 & 67.7 & 79.8 & 80.8 & 75.1 & 92.0 & 64.3 & 66.0 & 58.7 & 73.1 \\
        \hspace{-5pt}Effort (\textit{ICML'25}) & 64.8 & 59.1 & 96.3 & 82.2 & 75.4 & 95.5 & 61.5 & 57.6 & 86.5 & 66.4 & 60.1 & 97.4 & 53.8 & 52.9 & 69.3 & 74.0 & 68.3 & 89.7 & 67.1 \\
        \shline
        \hspace{-5pt}Qwen2.5-VL-7B & 50.7 & 71.7 & 2.4 & 53.6 & 92.8 & 7.9 & 80.2 & 99.4 & 60.7 & 67.5 & 100.0 & 35.1 & 52.5 & 100.0 & 5.1 & 50.5 & 56.1 & 4.5 & 59.2 \\
        \hspace{-5pt}InternVL3-8B & 54.4 & 55.5 & 44.1 & 67.1 & 83.6 & 42.7 & 77.1 & 85.4 & 65.3 & 66.5 & 78.9 & 45.0 & 47.4 & 37.5 & 7.6 & 51.8 & 52.0 & 45.1 & 60.7  \\
        \hspace{-5pt}MiMo-VL-7B & 57.7 & 56.4 & 68.6 & 75.6 & 73.7 & 79.6 & 79.4 & 73.5 & 91.9 & 70.7 & 68.3 & 76.5 & 67.7 & 67.1 & 69.5 & 54.9 & 54.8 & 56.2 & 67.7 \\
        \hspace{-5pt}GLM-4.1V-9BThink & 58.7 & 82.6 & 22.0 & 72.7 & 95.1 & 47.8 & 83.7 & 96.0 & 70.3 & 69.2 & 92.1 & 42.0 & 52.0 & 68.4 & 8.2 & 53.9 & 60.8 & 21.7 & 65.0 \\
        \hspace{-5pt}GPT-4o & 49.4 & 41.7 & 2.5 & 62.0 & 98.0 & 24.5 & 90.7 & 98.8 & 82.4 & 73.7 & 100.0 & 47.5 & 58.0 & 94.4 & 17.0 & 52.8 & 59.3 & 17.8 & 64.4 \\
        \hspace{-5pt}Gemini-2.5-Pro & 67.2 & 72.4 & 55.6 & 75.6 & 88.7 & 59.0 & 87.5 & 89.5 & 84.9 & 82.4 & 95.2 & 68.2 & 70.9 & 96.6 & 43.2 & 53.0 & 67.6 & 11.5 & 72.8 \\
        \rowcolor{Gray}\hspace{-7pt} \veritas $\;$(\textbf{ours}) & 58.6 & 54.7 & 100.0 & 84.1 & 94.1 & 72.8 & 92.3 & 90.0 & 95.1 & 90.2 & 86.5 & 95.2 & 89.2 & 86.3 & 93.2 & 78.5 & 76.1 & 83.0 & \textbf{82.2} \\
        \bottomrule[1pt]
        \end{tabular}
    }
    % \vspace{-0.25cm}
    \label{tab:supp_cd}
    % \vspace{-0.5cm}
\end{table}

The cross-domain results are shown in Table~\ref{tab:supp_cd}.
Co-SPY achieves the best results among previous detectors.
Most methods including ours achieve degraded performance.
Specifically, previous methods almost fail on all cross-domain subsets, while our method still achieves robust performance on in-the-wild forgeries (e.g., 92.3\% on Dreamina and 89.2\% on GPT-4o).
The performance on DeepFaceLab is extremely limited.
Different from cross-forgery and cross-model scenarios, the poor performance is due to low Precision.
Similar problem also exists in previous detectors.
This means those unseen low resolution real images are hard for model to distinguish.
Overall, our model strikes great improvements on cross-domain scenarios.

Besides, the zero-shot MLLMs tend to classify facial images into real photographs.
Even GPT-4o fails on many cases, exhibiting extremely low recall (i.e., less than 10\%).
Gemini-2.5-Pro demonstrates strong capacities for deepfake detection, especially on high-resolution images, even beating most fine-tuned specialized detectors.
Aggregating the above observations, we can find a intutive but interesting phenomenon: the MLLM-based detectors (especially reasoning MLLMs) are good at analyzing high-resolution images (typically over $512\times 512$) but fall short on low-resolution counterparts.
Once the MLLMs can ``see'' the image details, they are able to make accurate judgments and provide human-aligned reasoning process.
Conversely, small vision models exhibit certain advantages on low-resolution images, as such data is more suited for distribution modeling.
Therefore, a collaborative system of MLLMs and small models could be a promising future direction.

\begin{table}[t]
\caption{\setlength{\fboxsep}{1.5pt}Cross benchmark comparison. Performance (Acc.) on AIGIBench~\citep{li2025artificial} and the HydraFake-CD set. For previous methods, we implement two settings: (1) \textbf{train on FF++}~\citep{rossler2019faceforensics++} similar to previous setting, and (2) \textbf{train on HydraFake} dataset that contain multiple sources. The quantity of training samples for FF++ and HydraFake are kept consistent.
The performance of recent methods increase when trained on more diverse sources (highlighted in \colorbox{gray!15}{gray}), while similar gains are not observed for deepfake detection methods (highlighted in \colorbox{cyan!15}{blue}).}
    \centering
    \renewcommand\arraystretch{1.1}
    \scalebox{0.75}{
        \small
        \begin{tabular}{p{72pt}<{\raggedright}p{35pt}<{\centering}p{12pt}<{\centering}p{12pt}<{\centering}p{12pt}<{\centering}p{12pt}<{\centering}p{12pt}<{\centering}p{12pt}<{\centering}p{15pt}<{\centering}p{12pt}<{\centering}p{12pt}<{\centering}p{12pt}<{\centering}p{12pt}<{\centering}p{12pt}<{\centering}p{12pt}<{\centering}p{12pt}<{\centering}p{12pt}<{\centering}p{15pt}<{\centering}}
        \toprule[1pt]
        \multirow{2}{*}{\hspace{-5pt}\vspace{-2.5em}\textbf{Method}} & \multirow{2}{*}{\vspace{-2.5em}\textbf{Training}} & \multicolumn{6}{c}{\textbf{HydraFake-CD}} & \multirow{2}{*}{\vspace{-2.5em}\textbf{Avg.}} &  \multicolumn{8}{c}{\textbf{AIGIBench}} & \multirow{2}{*}{\vspace{-2.5em}\textbf{Avg.}} \\
        \cmidrule(lr){3-8} \cmidrule(lr){10-17}
        & & \rotatebox{60}{Deepface.} & \rotatebox{60}{InfiniteY.} & \rotatebox{60}{Dreamina} & \rotatebox{60}{Hailuo AI} & \rotatebox{60}{GPT-4o} & \rotatebox{60}{FFIW} &  & \rotatebox{60}{BLIP} & \rotatebox{60}{E4S} & \rotatebox{60}{InfiniteID} & \rotatebox{60}{InSwap} & \rotatebox{60}{IPAdapter} & \rotatebox{60}{R3GAN} & \rotatebox{60}{StyleSwim} & \rotatebox{60}{WFIR} &  \\
        \shline
        \hspace{-5pt}\rule{0pt}{7pt}F3Net (\textit{ECCV'20}) & FF++ & 52.9 & 56.4 & 63.8 & 76.0 & 69.5 & 52.6 & 61.9 & 65.6 & 44.1 & 51.8 & 52.5 & 72.7 & 51.1 & 47.1 & 46.1 & 53.9 \\
        \hspace{-5pt}UniFD (\textit{CVPR'23}) & FF++ & \underline{72.9} & 54.3 & 63.0 & 59.2 & 57.6 & 66.9 & 62.3 & 51.3 & 80.7 & 77.1 & 63.5 & 67.3 & 67.1 & 78.3 & 71.1 & 69.6 \\
        \hspace{-5pt}IID (\textit{CVPR'23}) & FF++ & 51.1 & 61.8 & 81.1 & 83.3 & 75.6 & 61.2 & \cellcolor{cyan!30}69.0 & 64.6 & 39.2 & 80.8 & 44.7 & 87.3 & 52.6 & 48.0 & 45.6 & \cellcolor{cyan!15}57.9 \\
        \hspace{-5pt}ProDet (\textit{NIPS'24}) & FF++ & 63.2 & 64.8 & \underline{85.6} & \underline{83.8} & 70.8 & 63.1 & \cellcolor{cyan!30}71.9 & 65.2 & 46.6 & 84.4 & 55.1 & 81.4 & 51.1 & 49.3 & 48.2 & \cellcolor{cyan!15}60.2  \\
        \hspace{-5pt}Co-SPY (\textit{CVPR'25}) & FF++ & 67.6 & 82.0 & 82.4 & 74.0 & 64.3 & 64.3 & 72.4 & 78.2 & 86.2 & 80.0 & 82.6 & 55.5 & 76.3 & \textbf{86.3} & 92.8 & 79.7 \\
        \hspace{-5pt}D$^3$ (\textit{CVPR'25}) & FF++ & 71.8 & 65.4 & 59.6 & 44.3 & 48.4 & 62.9 & \cellcolor{gray!15}58.7 & 48.7 & 80.3 & 60.0 & 68.9 & 55.5 & 60.3 & 66.3 & 59.4 & \cellcolor{gray!15}62.4 \\
        \hspace{-5pt}Effort (\textit{ICML'25}) & FF++ & \textbf{76.2} & 76.9 & 55.3 & 49.0 & 49.7 & 70.7 & \cellcolor{gray!15}63.0 & \textbf{83.8} & 85.9 & 82.2 & 77.3 & 71.3 & 81.4 & 82.3 & 83.1 & \cellcolor{gray!15}80.9 \\
        \shline
        \hspace{-5pt}\rule{0pt}{7pt}F3Net (\textit{ECCV'20}) & HydraFake & 57.7 & 78.5 & 55.6 & 68.6 & 66.2 & 66.4 & 65.5 & 72.1 & 77.7 & 69.8 & 80.4 & 62.9 & 53.5 & 44.3 & 74.5 & 66.9  \\
        \hspace{-5pt}UniFD (\textit{CVPR'23}) & HydraFake & 67.4 & 67.3 & 80.5 & 75.2 & 73.3 & 67.5 & 71.9 & 60.2 & 90.4 & 64.9 & 76.1 & 80.3 & 82.3 & 78.2 & 91.2 & 78.0 \\
        \hspace{-5pt}IID (\textit{CVPR'23}) & HydraFake & 65.2 & 69.9 & 63.8 & 63.3 & 63.8 & 64.2 & \cellcolor{cyan!15}65.0 & 78.3 & 65.8 & 76.0 & 62.7 & 77.8 & 58.3 & 48.2 & 81.3 & \cellcolor{cyan!30}68.6 \\
        \hspace{-5pt}ProDet (\textit{NIPS'24}) & HydraFake & 58.1 & \underline{82.9} & 71.3 & 75.6 & 66.3 & \underline{74.1} & \cellcolor{cyan!15}71.4 & \underline{82.4} & 85.9 & 81.1 & \underline{86.7} & 71.7 & 54.3 & 47.7 & 88.4 & \cellcolor{cyan!30}74.8 \\
        \hspace{-5pt}Co-SPY (\textit{CVPR'25}) & HydraFake & 67.6 & 80.0 & 82.5 & 74.0 & 79.5 & 64.3 & \underline{74.7} & 77.6 & 88.1 & \textbf{89.2} & 81.3 & 57.4 & 78.7 & \underline{86.2} & 94.2 & 81.6 \\
        \hspace{-5pt}D$^3$ (\textit{CVPR'25}) & HydraFake & 69.7 & 74.6 & 78.1 & 70.9 & \underline{80.8} & 64.3 & \cellcolor{gray!30}73.1 & 71.4 & 87.8 & 80.9 & 78.9 & 68.6 & 77.9 & 63.7 & 93.4 & \cellcolor{gray!30}77.8 \\
        \hspace{-5pt}Effort (\textit{ICML'25}) & HydraFake & 64.8 & 82.2 & 61.5 & 66.4 & 53.8 & 74.0 & \cellcolor{gray!30}67.1 & 82.1 & \underline{89.1} & 84.0 & 85.8 & \textbf{84.8} & \underline{87.1} & 82.3 & 82.7 & \cellcolor{gray!30}\underline{84.7} \\
        \rowcolor{Gray}\hspace{-7pt} \veritas $\;$(\textbf{ours}) & HydraFake & 58.6 & \textbf{84.1} & \textbf{92.3} & \textbf{90.2} & \textbf{89.2} & \textbf{78.5} & \textbf{82.2} & 81.9 & \textbf{93.5} & \underline{88.4} & \textbf{89.3} & \underline{81.8} & \textbf{91.3} & 85.2 & \textbf{99.8} & \textbf{88.9} \\
        \bottomrule[1pt]
        \end{tabular}
    }
    % \vspace{-0.25cm}
    \label{tab:cross_bench}
    % \vspace{-0.5cm}
\end{table}

\begin{table}[t]
\caption{Performance comparison on broader benchmarks, including LOKI~\citep{ye2024loki}, FakeClue~\citep{wen2025spot}, Forensics-Bench~\citep{wang2025forensics}, AIGIBench~\citep{li2025artificial} and Nano-banana-150K~\citep{ye2025echo}. Results of facial data in LOKI are also reported, since we target at deepfake detection.}
    \centering
    \renewcommand\arraystretch{1.1}
    \scalebox{0.75}{
        \small
        \begin{tabular}{p{72pt}<{\raggedright}p{20pt}<{\centering}p{20pt}<{\centering}p{22pt}<{\centering}p{22pt}<{\centering}p{20pt}<{\centering}p{20pt}<{\centering}p{27pt}<{\centering}p{27pt}<{\centering}p{20pt}<{\centering}p{20pt}<{\centering}p{20pt}<{\centering}p{20pt}<{\centering}}
        \toprule[1pt]
        \multirow{2}{*}{\hspace{-5pt}\textbf{Method}} & \multicolumn{2}{c}{\textbf{LOKI}} & \multicolumn{2}{c}{\textbf{LOKI (facial)}} & \multicolumn{2}{c}{\textbf{FakeClue}} & \multicolumn{2}{c}{\textbf{Forensics-Bench}} & \multicolumn{2}{c}{\textbf{AIGIBench}} & \multicolumn{2}{c}{\textbf{Nano-banana}} \\
        \cmidrule(lr){2-3} \cmidrule(lr){4-5} \cmidrule(lr){6-7} \cmidrule(lr){8-9} \cmidrule(lr){10-11} \cmidrule(lr){12-13}
        & Acc. & F1 & Acc. & F1 & Acc. & F1 & Acc. & F1 & Acc. & F1 & Acc. & F1 \\
        \shline
        \hspace{-5pt}UniFD (\textit{CVPR'23}) & 54.5 & 58.7 & 74.8 & 69.7 & 61.6 & 64.2 & 53.6 & 54.8 & 78.0 & 80.2 & 49.1 & 36.0 \\
        \hspace{-5pt}ProDet (\textit{NIPS'24}) & 53.8 & 56.6 & 63.2 & 66.4 & 62.9 & 69.8 & 65.1 & 72.0 & 74.8 & 73.9 & 64.6 & 63.8 \\
        \hspace{-5pt}Co-SPY (\textit{CVPR'25}) & 61.7 & \underline{65.8} & 79.1 & 75.6 & \underline{68.1} & \underline{72.4} & \textbf{70.8} & \textbf{76.0} & 81.6 & 84.3 & 52.0 & 40.9 \\
        \hspace{-5pt}D$^3$ (\textit{CVPR'25}) & 47.3 & 41.2 & 79.5 & 80.1 & 60.7 & 59.2 & 56.6 & 59.8 & 77.8 & 75.3 & \underline{70.7} & \underline{73.0} \\
        \hspace{-5pt}Effort (\textit{ICML'25}) & 53.8 & 50.0 & \underline{84.3} & \underline{84.6} & 65.0 & 63.2 & 57.0 & 59.4 & \underline{84.7} & \underline{87.6} & 62.7 & 52.1 \\
        \shline
        \hspace{-5pt}InternVL3-8B & 53.0 & 51.6 & 52.6 & 15.3 & 59.1 & 62.2 & 60.5 & 65.4 & 55.6 & 56.5 & 51.3 & 38.9 \\
        \hspace{-5pt}MiMo-VL-7B & \underline{65.1} & 64.3 & 69.7 & 65.0 & 67.2 & 71.6 & 63.8 & 70.5 & 62.8 & 64.3 & 60.7 & 55.6 \\
        \rowcolor{Gray}\hspace{-7pt} \veritas $\;$(\textbf{ours}) & \textbf{72.1} & \textbf{77.8} & \textbf{89.0} & \textbf{88.2} & \textbf{85.9} & \textbf{88.4} & \textbf{70.8} & \underline{74.9} & \textbf{88.9} & \textbf{90.4} & \textbf{86.3} & \textbf{89.0} \\
        \bottomrule[1pt]
        \end{tabular}
    }
    % \vspace{-0.25cm}
    \label{tab:cross_bench2}
    % \vspace{-0.5cm}
\end{table}

\subsubsection{Cross Benchmark Comparison}
\label{supp:cross_bench}
\begin{figure*}[t]
    \centering
    \includegraphics[width=0.96\linewidth]{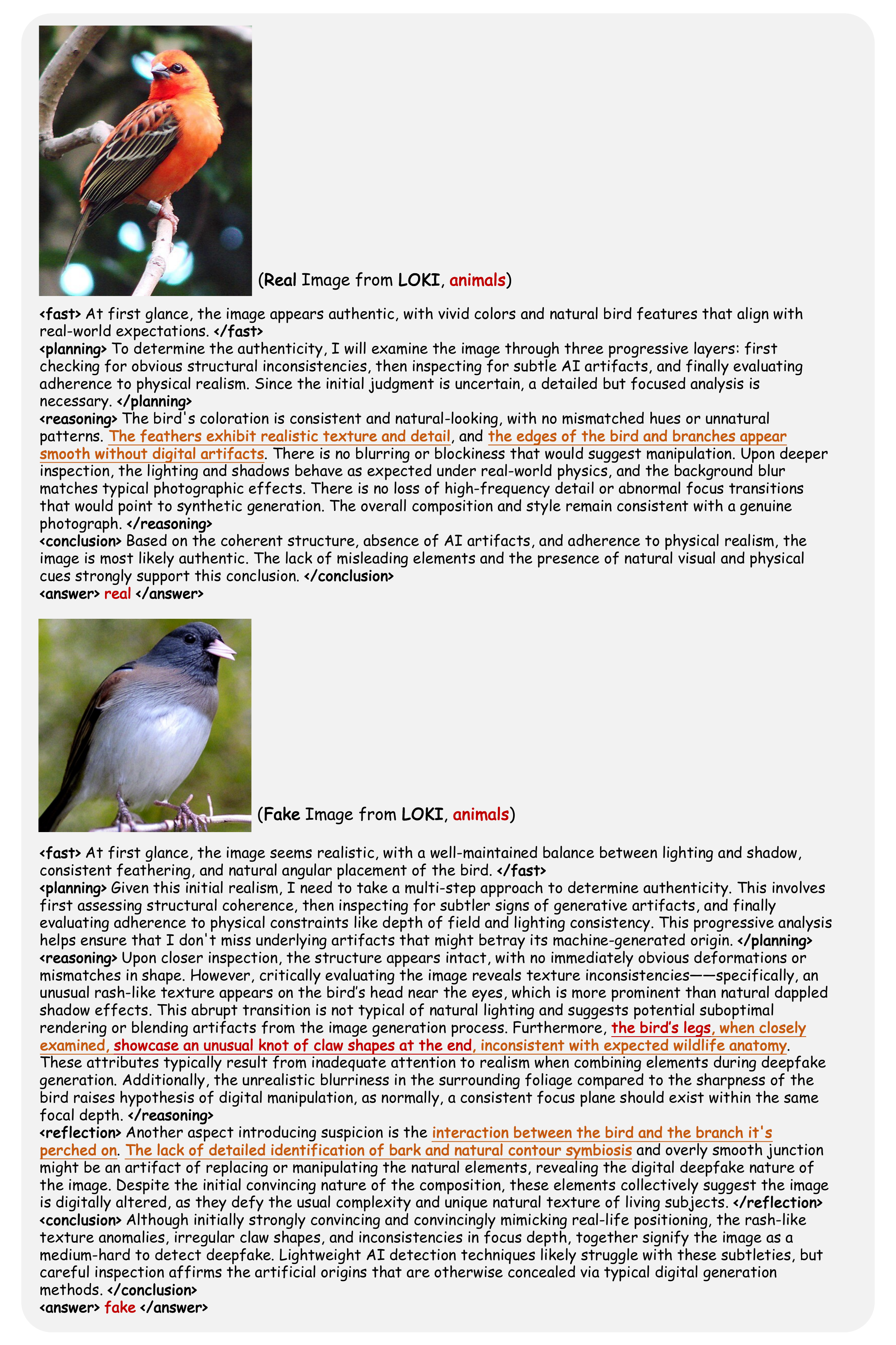}
     \caption{Reasoning output of \veritas$\,$ on AIGC images.
     }
	\label{fig:aigc1}
    \vspace{-0.2cm}
\end{figure*}

\begin{figure*}[t]
    \centering
    \includegraphics[width=0.96\linewidth]{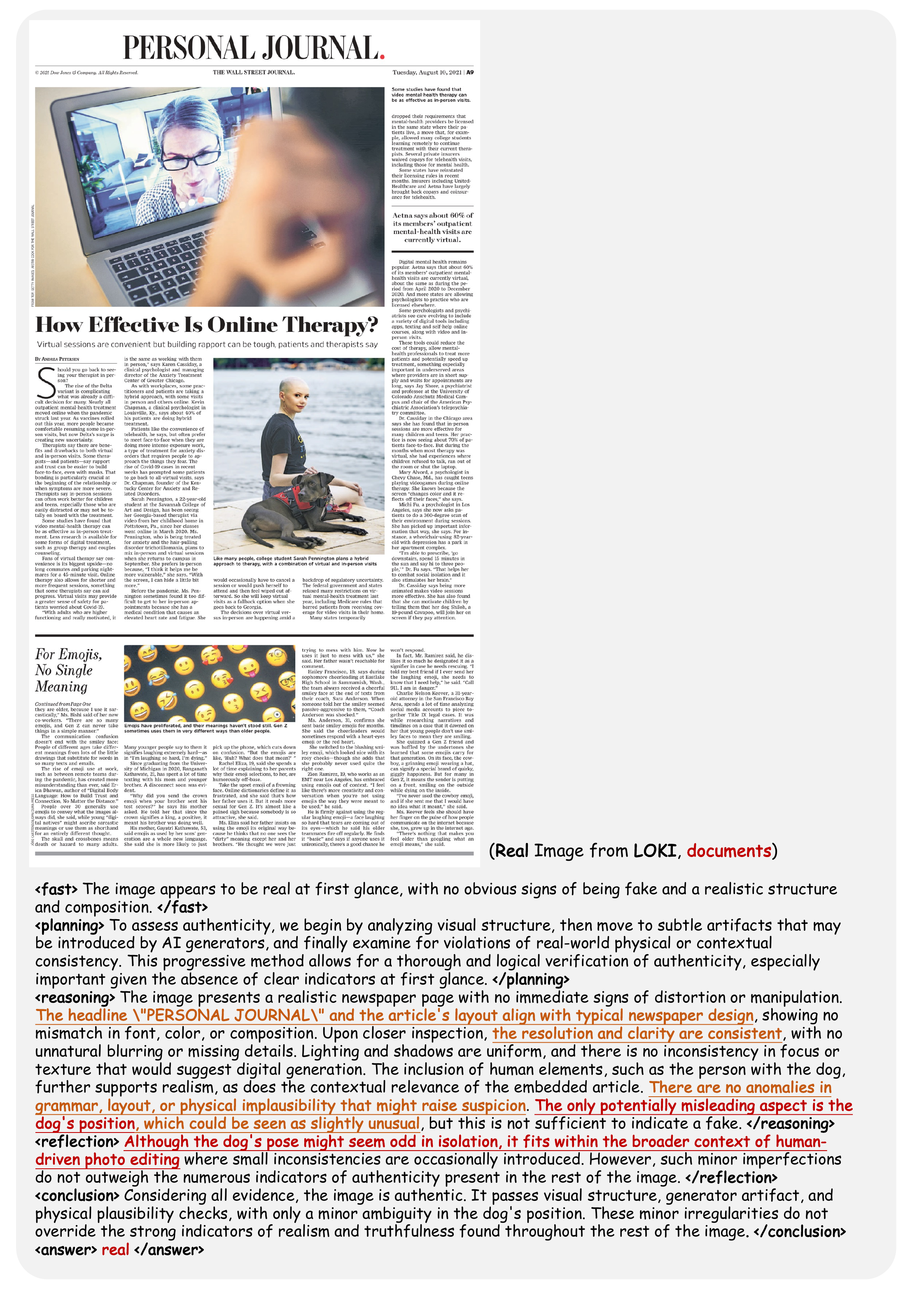}
     \caption{Reasoning output of \veritas$\,$ on AIGC images.
     }
	\label{fig:aigc2}
    \vspace{-0.2cm}
\end{figure*}

\begin{figure*}[t]
    \centering
    \includegraphics[width=0.96\linewidth]{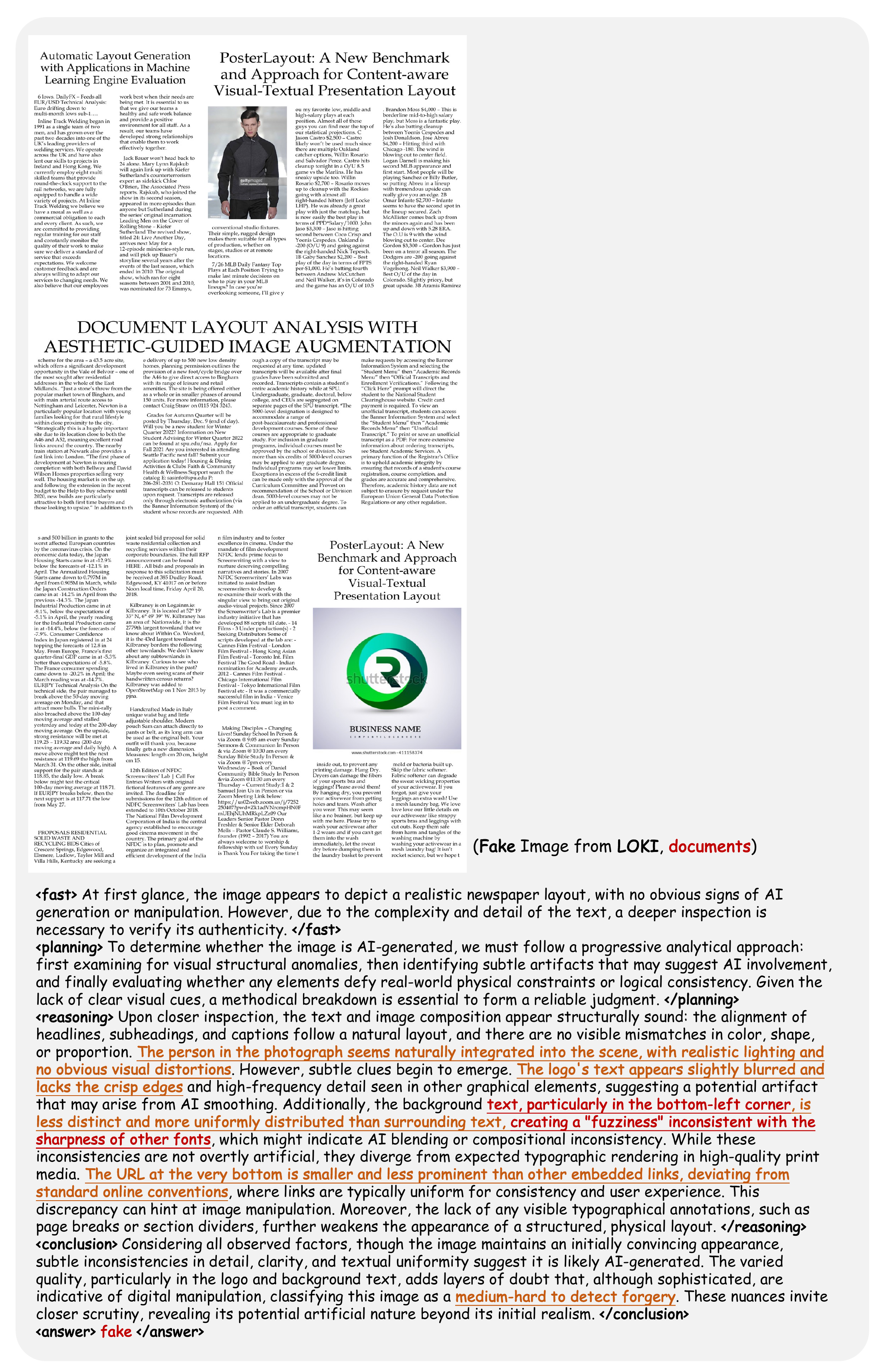}
     \caption{Reasoning output of \veritas$\,$ on AIGC images.
     }
	\label{fig:aigc2-2}
    \vspace{-0.2cm}
\end{figure*}

\begin{figure*}[t]
    \centering
    \includegraphics[width=0.86\linewidth]{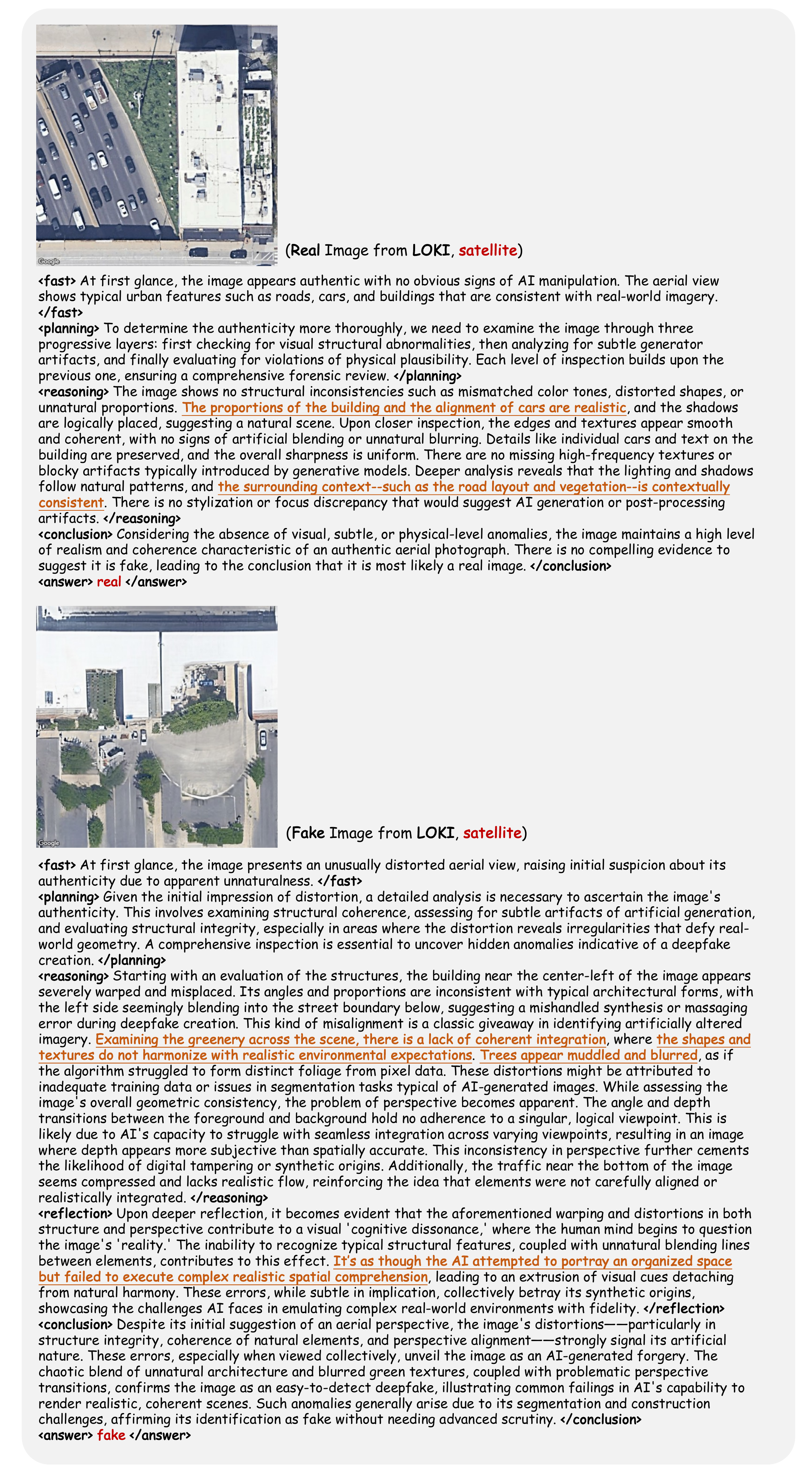}
     \caption{Reasoning output of \veritas$\,$ on AIGC images.
     }
	\label{fig:aigc3}
    \vspace{-0.2cm}
\end{figure*}

\begin{figure*}[t]
    \centering
    \includegraphics[width=0.86\linewidth]{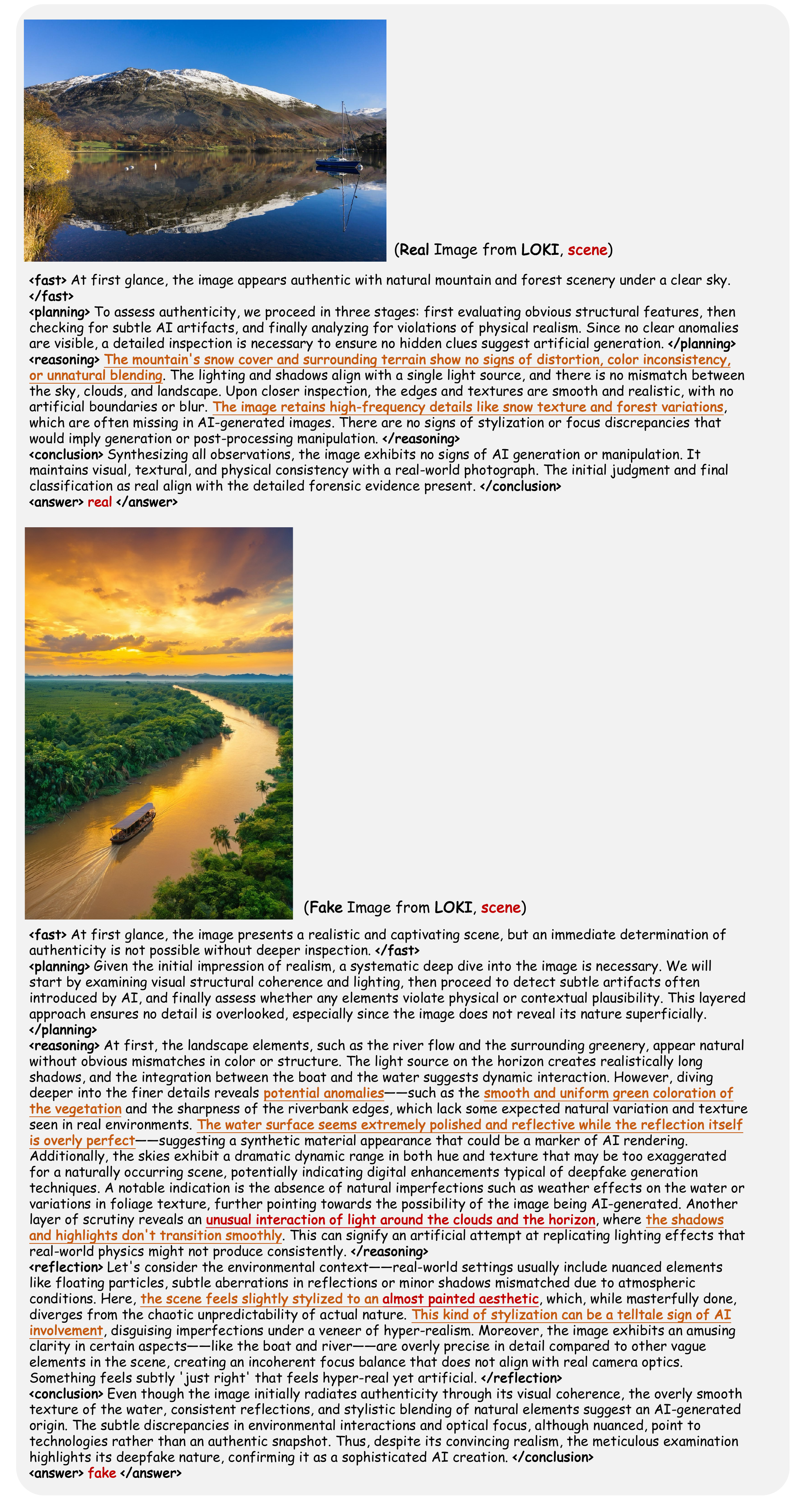}
     \caption{Reasoning output of \veritas$\,$ on AIGC images.
     }
	\label{fig:aigc4}
    \vspace{-0.2cm}
\end{figure*}

In Table~\ref{tab:cross_bench}, we provide a cross benchmark comparison.
We select the facial subsets from AIGIBench~\citep{li2025artificial}.
Note that these subsets also remain unseen in our HydraFake training set, which serves as an OOD testing of our method.
Since the real splits contain images of common objects, we substitute these images with facial images from VFHQ.
To investigate the impact of training sources, we also train existing methods on FF++~\citep{rossler2019faceforensics++} similar to previous one-to-many setting.
The quantity of training samples for FF++ and HydraFake are kept consistent (both are $48$K).
Specifically: 
\textbf{(1)} our HydraFake-CD is more challenging than AIGIBench, with the best result being 6.7\% lower (82.2\% vs. 88.9\%) and the second best result showing 10.0\% decrease (74.7\% vs. 84.7\%).
\textbf{(2)} On broader datasets, recent AIGC detection methods (e.g., Co-SPY and Effort) still demonstrates clear advantages to the methods tailored for deepfake (e.g., IID and ProDet).
This may indicate that specialized modules for facial images may struggle to generalize well to modern fully synthesized and high-fidelity deepfakes.
On these data, concurrently modeling the semantics and artifacts (like Co-SPY and Effort) may be more effective.
\textbf{(3)} Expanding the training data from FF++ to HydraFake brings promising gains for recent AIGC detection methods. For instance, D$^3$ increases from 58.7\% to 73.1\% on HydraFake.
However, similar gains are not observed for deepfake detection methods, e.g., both IID and ProDet suffer from performance drop on HydraFake when trained on more diverse sources.
This further reveal \textit{the gap} between current deepfake detection methods and practical usage.
When we have abundant training sources, \textit{the generalization performance does not scale up as expected}.
A comprehensive benchmark is necessary to measure the detectors' capacities more practically.
In Table~\ref{tab:cross_bench2}, we conduct additional evaluations on extensive benchmarks, including generic AIGC detection task.
Notably, \veritas$\,$ shows promising performance on these AIGC benchmarks, e.g., 72.1\% on LOKI and 85.9\% on FakeClue.
Note that \veritas$\,$ is only trained with facial forgery data.
Moreover, \veritas$\,$ generalizes well on the latest editing model (i.e., Nano-banana).
We also provide reasoning cases in Figure~\ref{fig:aigc1}, \ref{fig:aigc2}, \ref{fig:aigc2-2}, \ref{fig:aigc3}, \ref{fig:aigc4} to show the impressive adapation capacities of \veritas.

\subsubsection{Efficiency Comparison}
\label{supp:efficiency}
In Table~\ref{tab:efficiency}, we provide an efficiency analysis of our model.
All the data are obtained on a single PPUE GPU with the original Transformers library implementation.
We report the averaged inference time of a batch of images with the batch size set to $8$.
Specifically, we compare the efficiency of different reasoning paradigms of MLLMs.
For post-hoc explanation and flexible reasoning models, we do not perform MiPO since the human annotated data is hard to obtain.
We present a experimental prototype here to provide a understanding of the inference efficiency of our model.
Since inference efficiency is influenced by input resolutions and task difficulty, we divide samples into four parts.
Low resolutions are images with $256\times256$ size and high resolutions are $1024\times1024$.
Specifically, (1) our model achieves faster inference on low-resolution images, while becomes slower on high-resolution inputs. 
As discussed in Appendix~\ref{supp:more_results}, when the model can perceive finer details, it can perform more thorough reasoning, leading to improved accuracy and more inference time.
(2) Compared to flexible reasoning models, \veritas $\,$ incurs no significant increase in computational cost, yet achieves a 5.3\% performance gain, demonstrating the effectiveness of pattern-aware reasoning.
(3) Post-hoc explanation models exhibit little variation in efficiency across easy and hard samples, typically performing rigid, point-to-point analysis without adaptive reasoning.
(4) Without the P-GRPO to activate ``self-reflection'' and ``planning'' mechanisms, our cold-start model achieves better efficiency while still maintaining competitive performance.

\begin{table}[t]
\caption{Analysis on efficiency. We calculate the inference time (seconds) of a batch of images with the batch size set to $8$. The experiments are conducted on a single PPUE GPU with the \texttt{Transformers} library. We select samples of different resolutions and difficulty for illustration.}
    % \vspace{-0.25cm}
    \centering
    \renewcommand\arraystretch{1.1}
    \scalebox{0.9}{
        \small
        \begin{tabular}{p{75pt}<{\raggedright}p{40pt}<{\centering}p{40pt}<{\centering}p{40pt}<{\centering}p{40pt}<{\centering}p{30pt}<{\centering}}
        \multirow{2}{*}{\hspace{-5pt}\textbf{Model}} & \multicolumn{2}{c}{\textbf{Low Resolution} ($\downarrow$)} & \multicolumn{2}{c}{\textbf{High Resolution} ($\downarrow$)} & \multirow{2}{*}{\textbf{Acc.} ($\uparrow$)} \\
        \cmidrule(lr){2-3} \cmidrule(lr){4-5}
        & Easy & Hard & Easy & Hard & \\
        \shline
        \hspace{-5pt}Post-hoc Explanation & 27.08 & 28.19 & \textbf{37.64} & \textbf{36.06} & 83.4 \\
        \hspace{-5pt}Flexible Reasoning & 24.60 & 29.37 & 44.43 & 47.44 & 86.8 \\
        \hspace{-5pt}Ours (cold-start) & \textbf{19.26} & \textbf{24.94} & 40.72 & 46.40 & 89.3 \\
        \hspace{-5pt}Ours & 22.35 & 25.60 & 49.76 & 54.22 & \textbf{92.1}\\
        \rowcolor{Gray}\hspace{-5pt}\textbf{$\bm{\Delta}$ Post-hoc Exp.} & \textcolor{red}{\textbf{$\downarrow$4.73}} & \textcolor{red}{\textbf{$\downarrow$2.59}} & \textcolor{mgreen}{\textbf{$\uparrow$12.12}} & \textcolor{mgreen}{\textbf{$\uparrow$18.16}} & \textcolor{red}{\textbf{$\uparrow$8.7}} \\
        \rowcolor{Gray}\hspace{-5pt}\textbf{$\bm{\Delta}$ Flexible Reason.} & \textcolor{red}{\textbf{$\downarrow$2.25}} & \textcolor{red}{\textbf{$\downarrow$3.77}} & \textcolor{mgreen}{\textbf{$\uparrow$5.33}} & \textcolor{mgreen}{\textbf{$\uparrow$6.78}} & \textcolor{red}{\textbf{$\uparrow$5.3}} \\
        \end{tabular}
    }
    % \vspace{-0.25cm}
    \label{tab:efficiency}
    % \vspace{-0.5cm}
\end{table}

\subsubsection{Effect of Training Data in P-GRPO Stage}
\label{supp:pgrpo_data}
In Table~\ref{tab:abl_pgrpo_data}, we investigate the impact of training data in P-GRPO.
For our \veritas, we adopt balanced sampling among manipulation types, which achieves superior performance compared to randomly sampled data.
As adopted in mathematical and coding problems, the hard sampling (the samples that models fail to reach all correct answers in $8$ rollouts) achieves inferior performance in our case, but this still yields improvements over cold-start model.
Moreover, we add about $1/3$ unseen data from AIGIBench into P-GRPO stage.
Note that these data are unseen during previous training stages, but are not overlapped with the testing domain of HydraFake.
As shown in Table~\ref{tab:abl_pgrpo_data}, this yields promising improvements on cross-forgery scenarios (3.8\% over our \veritas).
From the observations, we point out that the cold-start model is a \textit{good policy model}. While in this paper we only use in-domain data during P-GRPO for fair comparisons, the users can add OOD data flexibly to elevate the detection ability, which can be achieved in two approaches:
(1) (\textbf{\textit{a cheap and scalable way}}) adopt data with binary labels and our P-GRPO for training.
(2) (\textbf{\textit{a fine-grained and controllable way}}) use the cold-start model or \veritas $\,$ to further construct a high-quality CoT dataset for customized deepfake data. This may require manual preference filtering but can further enhance the reasoning quality on target data.

\begin{table}[t]
    \scriptsize
    \centering
	\begin{minipage}{0.49\linewidth}
            \caption{Ablation studies of the hyperparameter $\beta'$ (strength of KL penalty in P-GRPO).}
            \vspace{-4pt}
    \label{tab:abl_hyper_beta}
    \centering
    \renewcommand\arraystretch{1.1}
    \scalebox{0.9}{
        \small
        \begin{tabular}{p{50pt}<{\raggedright}p{18pt}<{\centering}p{18pt}<{\centering}p{18pt}<{\centering}p{18pt}<{\centering}}
        \textbf{Value of $\beta'$} & ID & CM & CF & CD \\
        % \cmidrule(lr){2-4}
        \shline
        $\beta'\!=\!0.04$ & 96.8 & 98.4 & 89.1 & 80.2 \\
        $\beta'\!=\!0.01$ & 96.8 & 98.3 & 89.6 & 81.5 \\
        $\beta'\!=\!0.001$ & \textbf{97.3} & \textbf{98.6} & 89.3 & 81.9 \\
        $\beta'\!=\!0.0$ & \textbf{97.3} & \textbf{98.6} & \textbf{90.3} & \textbf{82.2} \\
        \end{tabular}
    }
	\end{minipage}
    \hfill
    \begin{minipage}{0.49\linewidth}
    \caption{Ablation studies of the hyperparameter $G$ (number of generations within each group).}
            \vspace{-4pt}
    \label{tab:abl_hyper_gen}
    \centering
    \renewcommand\arraystretch{1.1}
    \scalebox{0.9}{
        \small
        \begin{tabular}{p{50pt}<{\raggedright}p{18pt}<{\centering}p{18pt}<{\centering}p{18pt}<{\centering}p{18pt}<{\centering}}
        \hspace{-5pt}\textbf{Value of $G$} & ID & CM & CF & CD \\
        \shline
        $G\!=\!4$ & 97.3 & \textbf{98.6} & \textbf{90.3} & \textbf{82.2} \\
        $G\!=\!8$ & \textbf{97.4} & 98.4 & 89.8 & 82.0 \\
        $G\!=\!12$ & 96.7 & 97.6 & 88.8 & 80.5 \\
        $G\!=\!16$ & 96.9 & 93.2 & 88.9 & 80.4 \\
        \end{tabular}
    }
	\end{minipage}
\end{table}

\subsubsection{Analysis of Hyperparameters}
\label{supp:ana_hyper}
\textbf{Analysis of hyperparameter $\beta'$.}
$\beta'$ controls the strength of penalty when the outputs deviate from the reference model.
From Table~\ref{tab:abl_hyper_beta}, smaller $\beta'$ yields better performance on cross-forgery and cross-domain sets, suggesting that stronger exploration helps activate advanced reasoning behaviors of the cold-start model, improving generalization on OOD data.

\textbf{Analysis of hyperparameter $G$ (number of generated rollouts in P-GRPO).}
From Table~\ref{tab:abl_hyper_gen}, more generations within one group do not bring improvements while the training costs increase.
We attribute this to the task gap between deeepfake detection and common tasks.
While mathematical problems often admit multiple valid solution paths, deepfake detection is a fact-based classification task with a more constrained reasoning space. 
In such cases, excessive group size leads to redundant exploration.
This is also the reason we apply cold-start before the online RL stage, which ensures meaningful exploration during RL.

\subsubsection{Effect of Different Reward Model}
\label{supp:reward_model}
\begin{figure}[t]
    \scriptsize
    \centering
	\begin{minipage}{0.42\linewidth}
            \captionof{table}{Ablation studies on the training data of P-GRPO. The cold-started model serves as the baseline. We try four types of data selection. See analysis for details.}
            \vspace{-4pt}
    \label{tab:abl_pgrpo_data}
    \centering
    \renewcommand\arraystretch{1.1}
    \scalebox{0.9}{
        \small
        \begin{tabular}{p{75pt}<{\raggedright}p{14pt}<{\centering}p{14pt}<{\centering}p{14pt}<{\centering}p{14pt}<{\centering}}
        \hspace{-5pt}\textbf{Training Data} & ID & CM & CF & CD \\
        % \cmidrule(lr){2-4}
        \shline
        \hspace{-7pt}\gc{Baseline (cold-start)} & \gc{96.8} & \gc{95.8} & \gc{85.1} & \gc{79.5} \\
        \hspace{-7pt}Random (seed=42) & 96.9 & 98.2 & 88.4 & 80.1 \\
        \hspace{-7pt}Random (seed=100) & \underline{97.0} & 97.9 & 88.0 & 80.6 \\
        \rowcolor{Gray}\hspace{-9pt} Balanced sampling & \textbf{97.3} & \underline{98.6} & \underline{90.3} & \underline{82.2} \\
        \hspace{-7pt}Hard sampling & 96.9 & 98.4 & 88.6 & 81.3 \\
        \hspace{-7pt}Partially unseen & 95.7 & \textbf{99.4} & \textbf{94.1} & \textbf{82.4} \\
        \end{tabular}
    }
	\end{minipage}
    \hfill
    \begin{minipage}{0.56\linewidth}
    \includegraphics[height=3.7cm, width=0.99\linewidth]{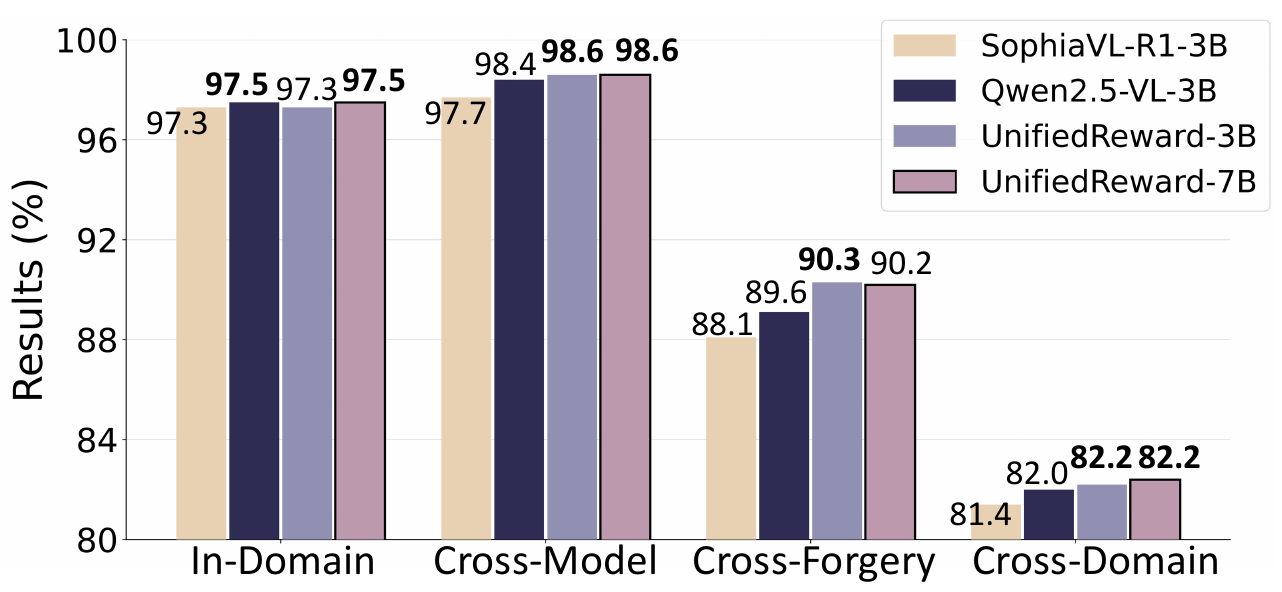}
    \vspace{-3pt}
    \caption{Ablations on the reward model $\mathcal{M}$.}
    \label{fig:abl_reward_model}
	\end{minipage}
\end{figure}

We investigate different reward models, including SophiaVL-R1-Thinking-Reward-Model-3B~\citep{fan2025sophiavl}, Qwen2.5-VL-3B, UnifiedReward-Qwen-3B and UnifiedReward-Qwen-7B.
From Figure~\ref{fig:abl_reward_model}, SophiaVL-R1-3B achieves degraded performance.
This is due to that Sophia is specifically trained to measure the quality of CoT, while the instruction following ability is limited, which is not capable of measuring the novelty of reflection content.
In contrast, Qwen2.5-VL-3B and UnifiedReward-Qwen-3B can better distinguish the reflection quality.
Further scaling up to 7B does not bring significant improvements.

\subsection{Full Prompt Templates}
\label{supp:full_prompts}
In this section, we provide all the prompt templates used in our method.
This includes the following parts:
\begin{itemize}[left=2pt]
    \item The prompts for pattern-aware SFT data annotation.
For fake images, the process contains three stages as shown in Figure~\ref{prompt:s1}, Figure~\ref{prompt:s2} and Figure~\ref{prompt:s3}.
For real images, the process contains two stages (without the anomalies detection stage) as shown in Figure~\ref{prompt:real_s1} and Figure~\ref{prompt:real_s2}.
    \item The prompts for reasoning quality evaluation, which contains score evaluation (Figure~\ref{prompt:score}) and pairwise evaluation (Figure~\ref{prompt:pair}).
    \item The prompt for generating personalization prompts, as shown in Figure~\ref{prompt:personalization}.
    \item The prompt for reflection quality reward model $\mathcal{M}$, as shown in Figure~\ref{prompt:reward}.
    \item The system prompt for our \veritas$\,$ model as shown in Figure~\ref{prompt:sys_veritas}. The system prompts for all training stages and inference stage are consistent.
    \item The prompt for zero-shot inference of MLLMs. We tested several prompts for each MLLM. Firstly we found that directly prompting them to perform pattern-aware reasoning like \veritas$\,$ fails in most cases. 
    Therefore we perform common CoT instead.
    The prompts for Qwen2.5-VL-7B, InternVL3-8B and GLM-4.1V-9B-Thinking are provided in Figure~\ref{prompt:zs_qwen}.
    For MiMo-VL-7B, we found that providing priori information is harmful to the performance, and we adopt simple system prompt instead, as shown in Figure~\ref{prompt:zs_mimo}.
    Similarly, we keep the default system prompt and only constraining the output format in user prompt for GPT-4o and Gemini-2.5-Pro, as shown in Figure~\ref{prompt:zs_gpt}.
\end{itemize}

\subsection{More Qualitative Results}
\label{supp:more_qualitative_res}
\begin{figure*}[t]
    \centering
    \includegraphics[width=0.90\linewidth]{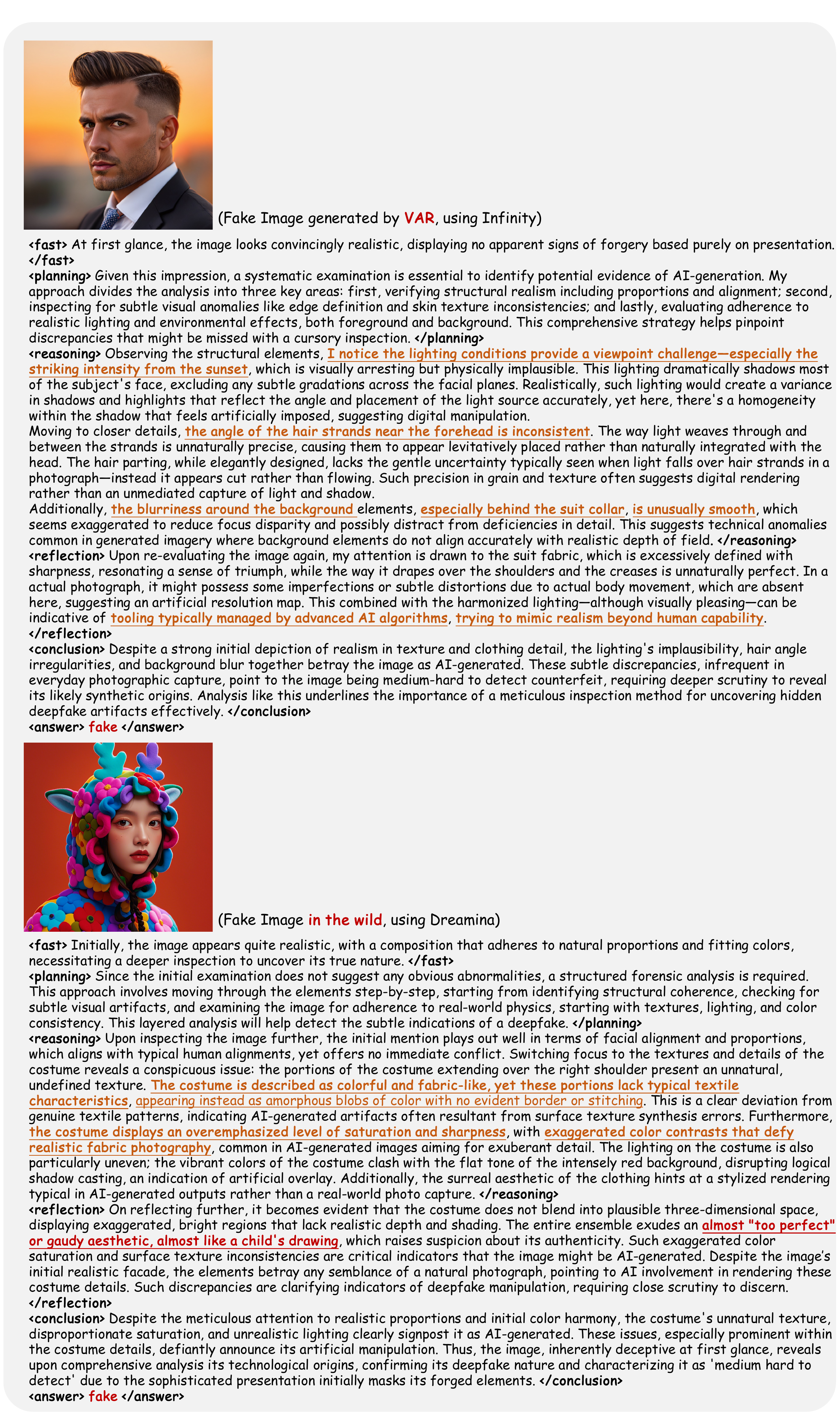}
     \caption{Reasoning output of our \veritas$\,$ model.
     }
	\label{fig:case5}
    \vspace{-0.2cm}
\end{figure*}

\begin{figure*}[t]
    \centering
    \includegraphics[width=0.90\linewidth]{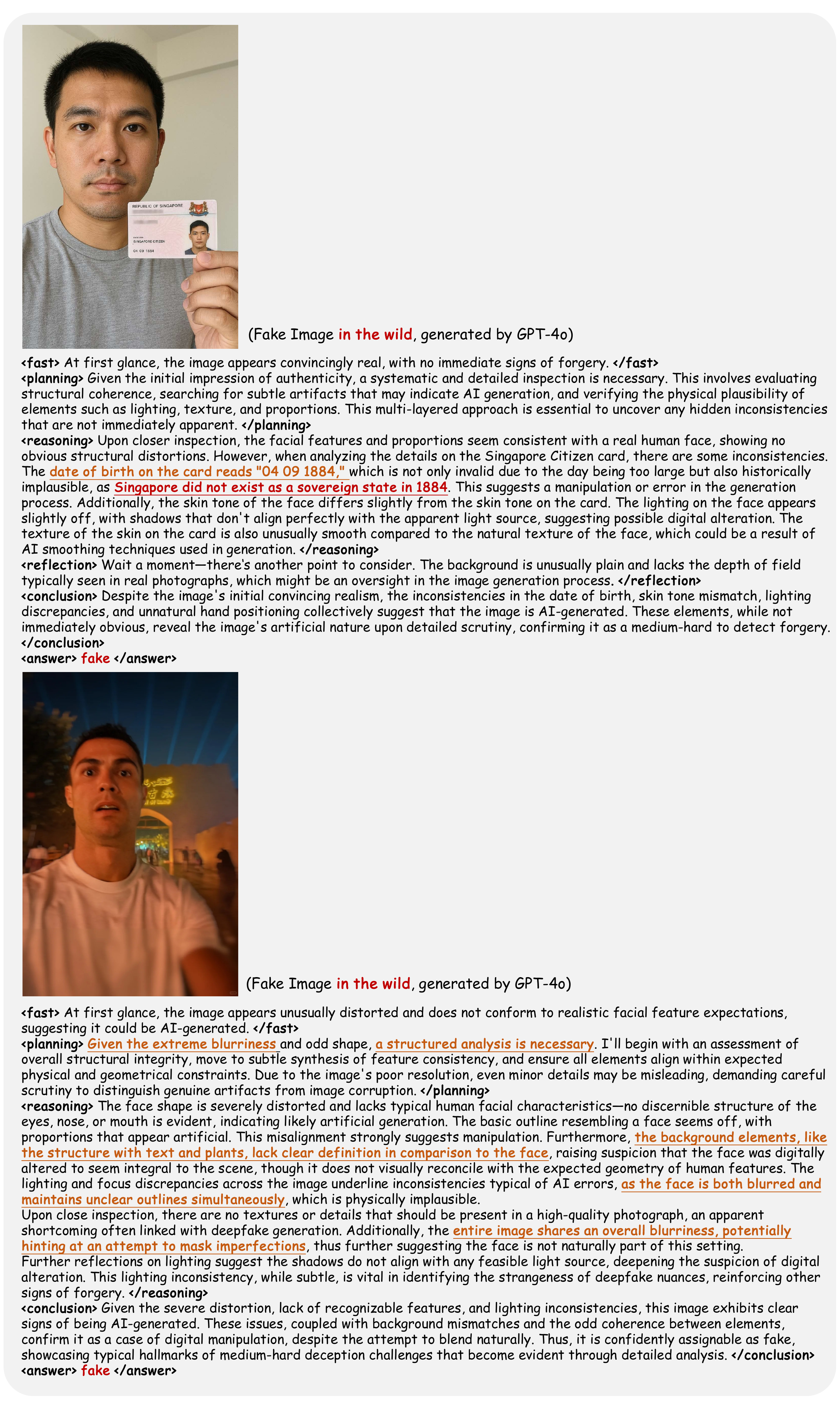}
     \caption{Reasoning output of our \veritas$\,$ model.
     }
	\label{fig:case6}
    \vspace{-0.2cm}
\end{figure*}

We provide more examples of our model's reasoning outputs.
Specifically, our model can perform adaptive pattern-aware reasoning, generating direct and concise analysis for obviously fake images.
It can also conduct thorough and holistic analysis for high-fidelity fake images.
All the examples except FF++ are from OOD scenarios.
It is worth noting that while being trained on pure in-domain facial data, \veritas$\,$ exhibits promising AIGC analysis abilities, e.g., the infeasible date of birth on ID card (Figure~\ref{fig:case6}) and over-stylized texture of fabric (Figure~\ref{fig:case5}).
Such abilities mainly emerges from the combination of MiPO and P-GRPO.
As mentioned in our main text, MiPO ensures high-quality rollouts in subsequent stage, which enables more accurate policy updates for online RL.
The effective explorations during RL facilitate the deep reasoning capacities.
Note that such observation is different from that of BusterX++~\citep{wen2025busterx++}, which found that cold-start constrains the output distribution and shrinks the OOD generalization abilities.
We suppose the discrepancy is due to the \textit{rich semantics in AI-generated content allow the MLLMs to succeed with pure RL}, since they are proficient at capturing semantic-level clues and is capable of generating high-quality rollouts at initial stage.
However, the semantics of deepfake images are extremely limited, with most anoamlies lying on low-level artifacts.
In such cases, cold-start is necessary and our work incorporates SFT and MiPO to instill the human-aligned reasoning capacities into base models.

\vspace{-0.2cm}
\subsection{More Qualitative Comparisons with Existing MLLM-based detectors}
\label{supp:more_qualitative_comp}

\begin{figure*}[t]
    \centering
    \includegraphics[width=0.96\linewidth]{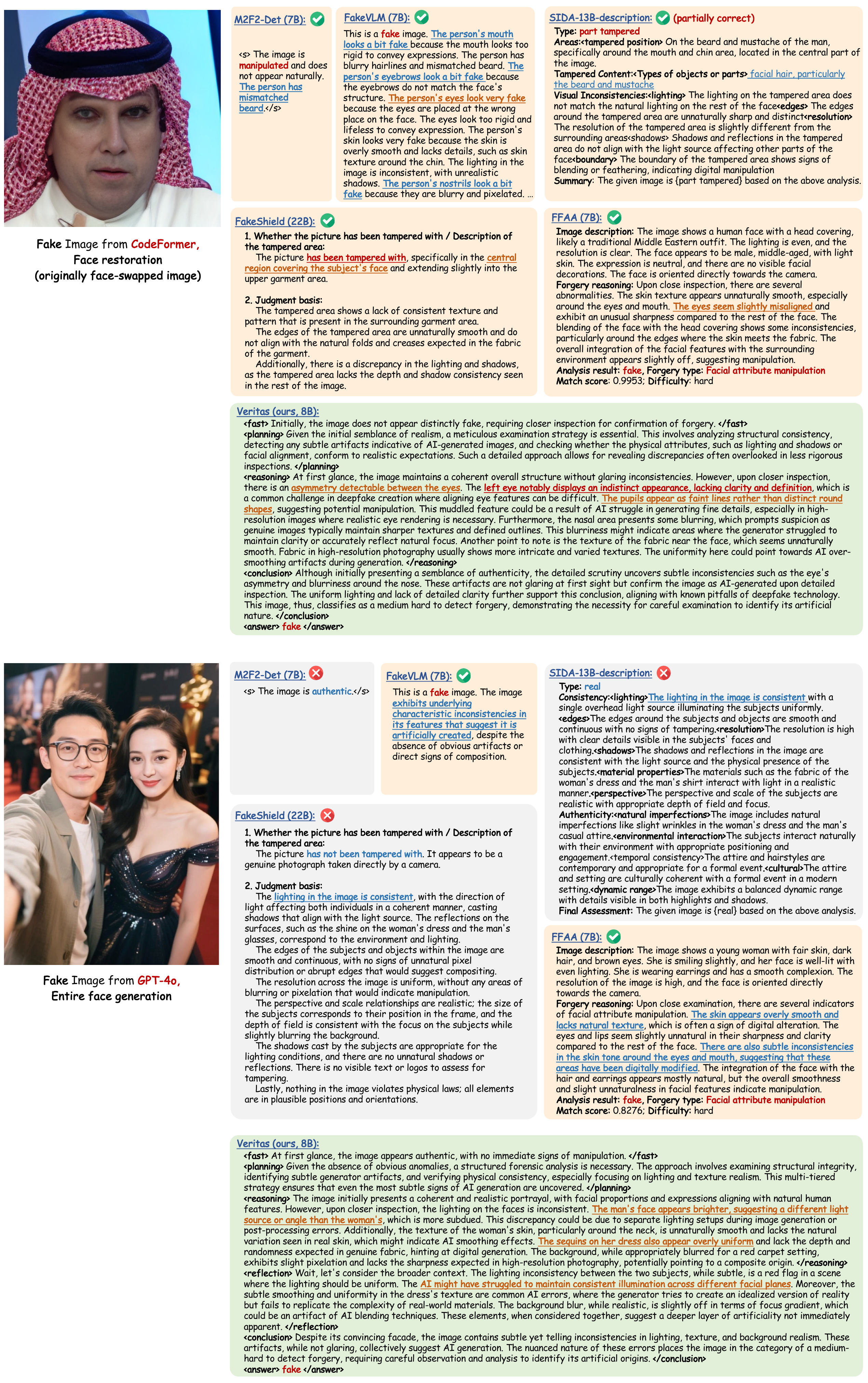}
     \caption{Explanation comparisons between \veritas$\,$ and existing MLLM-based detectors.
     }
	\label{fig:mllm_comp1}
    \vspace{-0.2cm}
\end{figure*}

\begin{figure*}[t]
    \centering
    \includegraphics[width=0.96\linewidth]{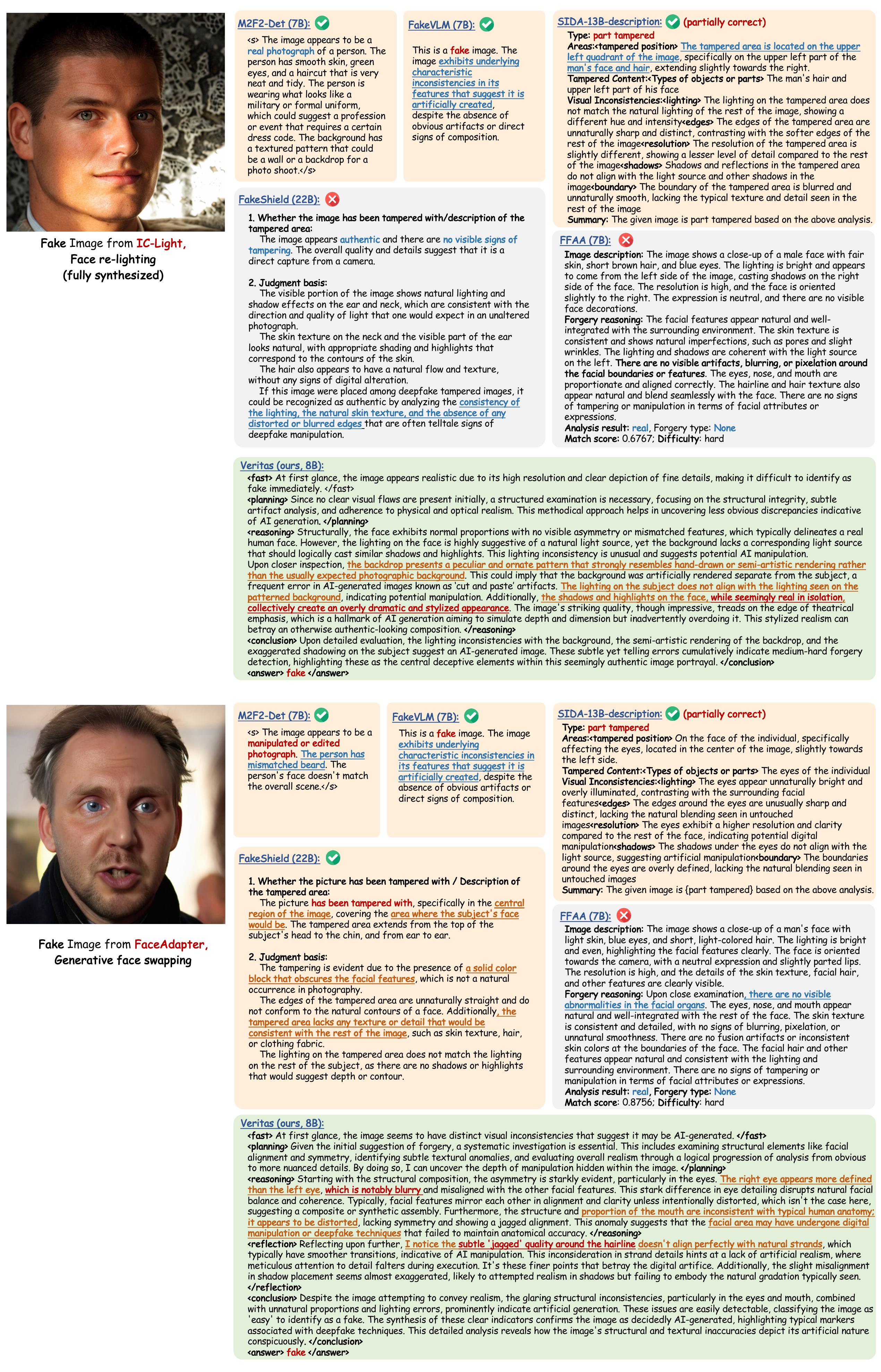}
     \caption{Explanation comparisons between \veritas$\,$ and existing MLLM-based detectors.
     }
	\label{fig:mllm_comp2}
    \vspace{-0.2cm}
\end{figure*}

\begin{figure*}[t]
    \centering
    \includegraphics[width=0.96\linewidth]{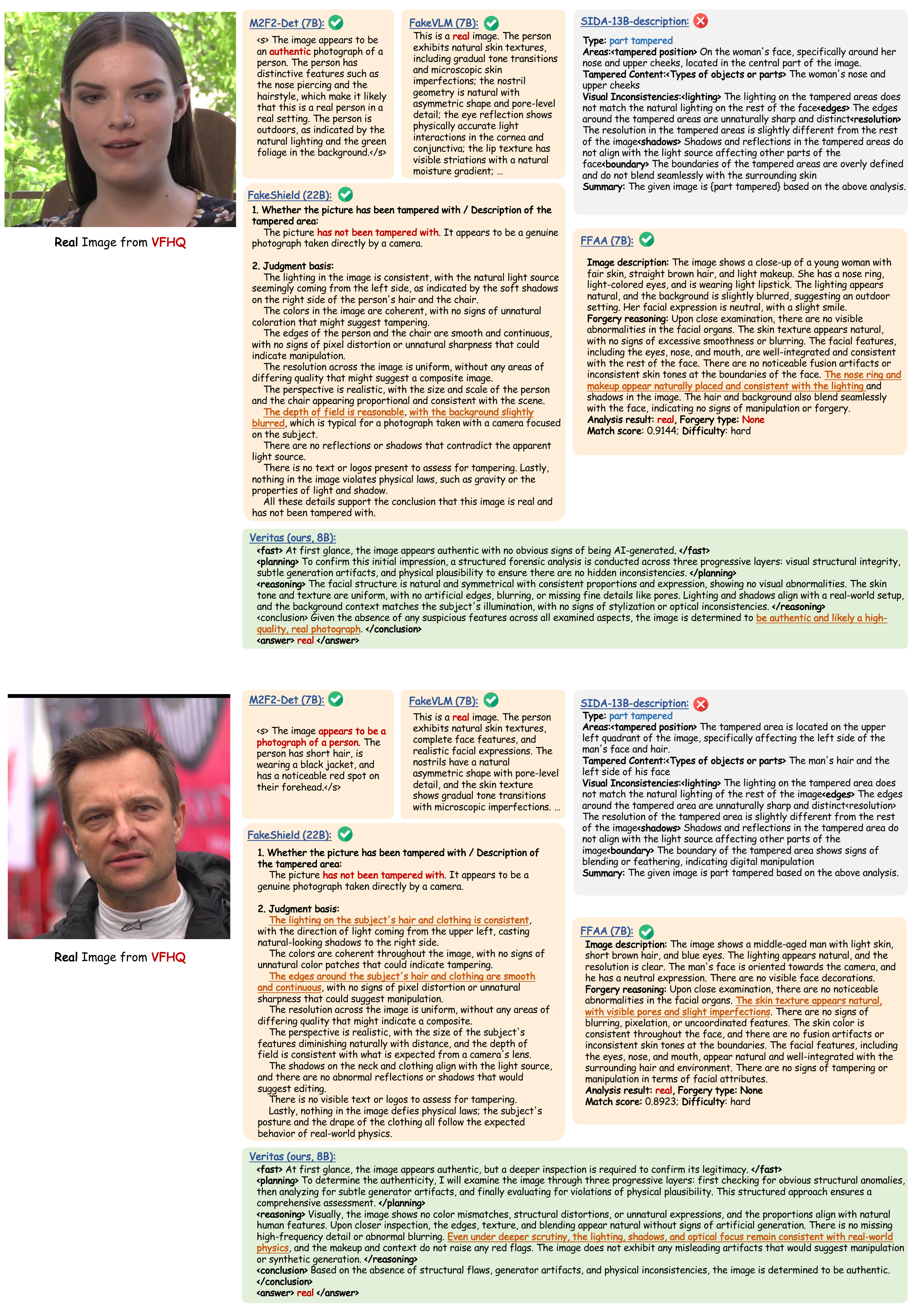}
     \caption{Explanation comparisons between \veritas$\,$ and existing MLLM-based detectors.
     }
	\label{fig:mllm_comp3}
    \vspace{-0.2cm}
\end{figure*}

We provide more reasoning comparisons between \veritas$\,$ and existing MLLM-based detectors.
Among the compared models, M2F2-Det and FFAA are specialized for deepfake detection, while other methods are generic forgery detection models.
As shown in Figure~\ref{fig:mllm_comp1}, \ref{fig:mllm_comp2}, \ref{fig:mllm_comp3},
\textbf{M2F2-Det} excels at performing faithful analyses within facial region. However, it lacks consideration of deeper dimensions, resulting in suboptimal performance on certain fully synthesized data that require considerations about overall context.
\textbf{FFAA} provides more detailed analyses. However, the logical coherence between ``description'' and ``reasoning'' part is weak, and the ``reasoning'' part lacks in-depth understanding.
\textbf{FakeShield} falls short on analyzing fully synthesized facial data, but it demonstrates certain advantages in local artifact analysis (Figure~\ref{fig:mllm_comp2} lower), since it is specifically trained for IMDL tasks.
\textbf{SIDA-13B-description} provides generally high-quality explanations. 
However, it has a tendency to classify real facial images as fake.
\textbf{FakeVLM} provides low-quality explanations regarding facial forgeries despite its high detection accuracy, e.g., \textit{most cases are explained as ``The image exhibits underlying characteristic inconsistencies in its features that suggest it is artificially created''}. Such vague and template-like explanantions are likely due to its large-scale SFT training nature.
In contrast, \veritas$\,$ generates holistic and faithful reasoning process.

\vspace{-0.2cm}
\subsection{Failure Analysis of \veritas}
\label{supp:failure_ana}
\begin{figure*}[t]
    \centering
    \vspace{-0.5cm}
    \includegraphics[width=0.92\linewidth]{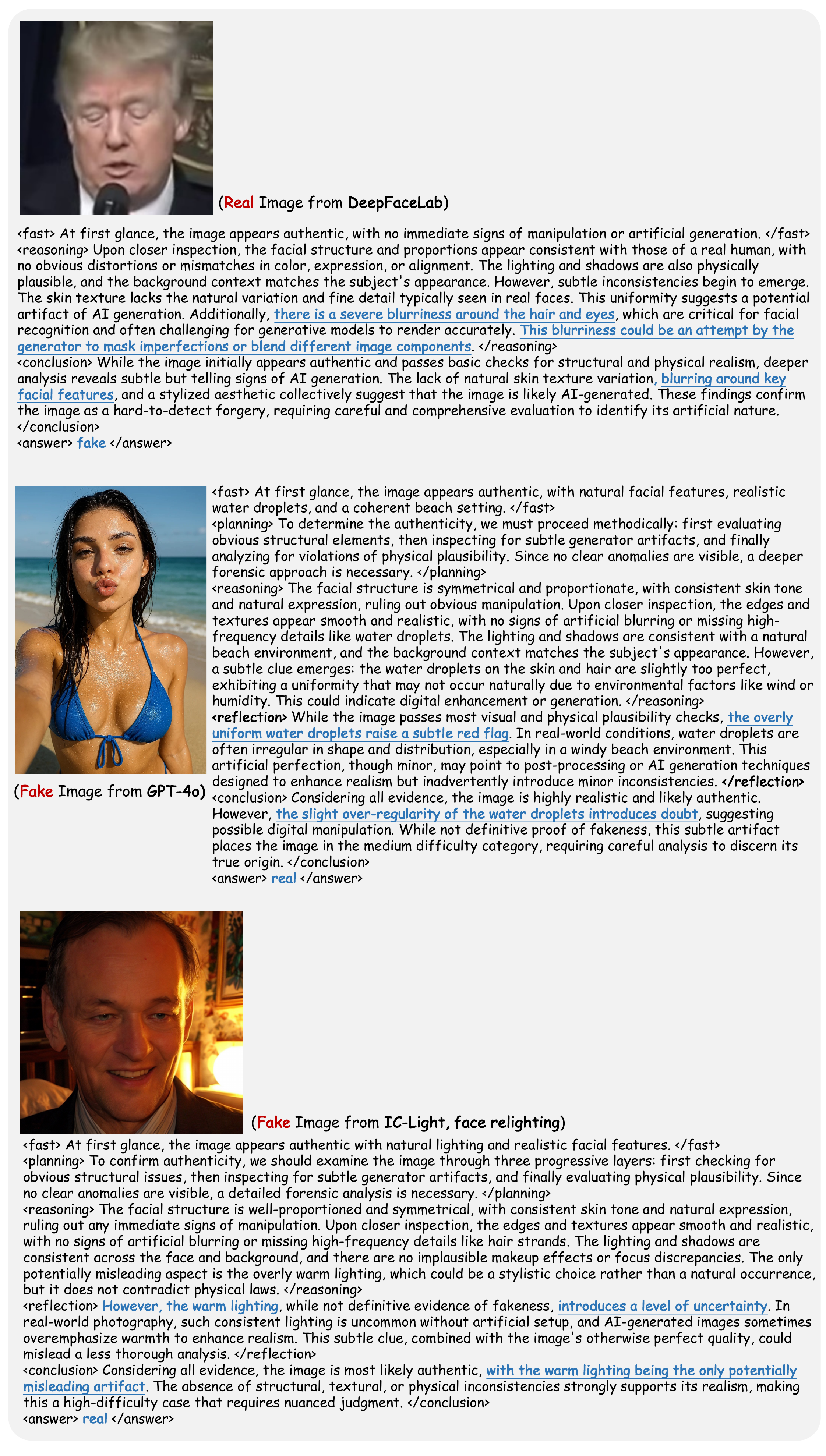}
     \caption{Failure cases of \veritas.
     }
	\label{fig:failures}
    \vspace{-0.2cm}
\end{figure*}

For \textbf{real} images, the failures mainly clustered at low-resolution data.
As shown in Figure~\ref{fig:failures} upper, these data are generally in low quality, where the unexpected artifacts such as localized blurriness would affect the model's judgement.
For \textbf{fake} images, failures mainly occur on totally unseen forgery types such as face relighting.
However, although the final answer is incorrect, \veritas$\,$ still figure out suspicious clues, e.g., ``overly uniform water droplets raise \textbf{red flag}'' and ``the warm lighting introduces \textbf{uncertainty}'' in Figure~\ref{fig:failures}.
This providing valuable insights that could be used for further scrutiny or future improvements.

\subsection{The Use of Large Language Models}
We used LLMs for grammatical refinement and language polishing of the paper, aiming to improve the clarity and readability.
Some MLLMs are used for the annotation of reasoning data, which is a common practice.
Besides, the LLMs are not involved in research design or idea generation.

\subsection{Ethics Statement}
\label{supp:ethics_state}
All real facial data used in this work are from publicly available academic datasets.
The fake images include those from public benchmarks and those generated by our team using generative models or face-swapping techniques.
The latter were created only from public or synthetic data, with no unauthorized use of personal images.
Our work focuses on improving deepfake detection to combat misinformation, and all data are used strictly for non-commercial, academic purposes.

\subsection{Limitations and Future Work}
\label{supp:future_work}
While HydraFake involves multi-level evaluations, it is limited to the image modality.
With recent advances in video generation models, extracting frames from videos and detecting manipulations solely based on spatial artifacts is challenging.
Moreover, as analyzed in Appendix~\ref{supp:more_results}, our \veritas $\,$ model still exhibits shortages on low-quality subsets such as DeepFaceLab and FFIW as the reasoning requires more visual details.
Therefore, we figure out the \textbf{future directions}:
\textbf{(1)} A collaborative system of MLLMs and small vision models, since the MLLM-based detectors (especially reasoning MLLMs) are good at analyzing high-resolution images while small vision models exhibit certain advantages on low-resolution counterparts.
This has been explored by a recent work~\citep{chen2024x2}, while how to develop a more adaptive or agent-like system is still an interesting problem.
\textbf{(2)} A unified image-video deepfake benchmark. Recent video generation models are capable of creating high-fidelity talking faces and hand-face interactions (e.g., touching eyebrows or nose), posing new challenges to facial security systems.
Due to high frame-level realism, traditional frame-based detectors often fail.
Consequently, there is a growing need for unified detection frameworks capable of handling both image and video inputs, as well as rigorous benchmarks to facilitate the development of robust detection methods.

\begin{figure*}[t]
    \centering
    \includegraphics[width=0.96\linewidth]{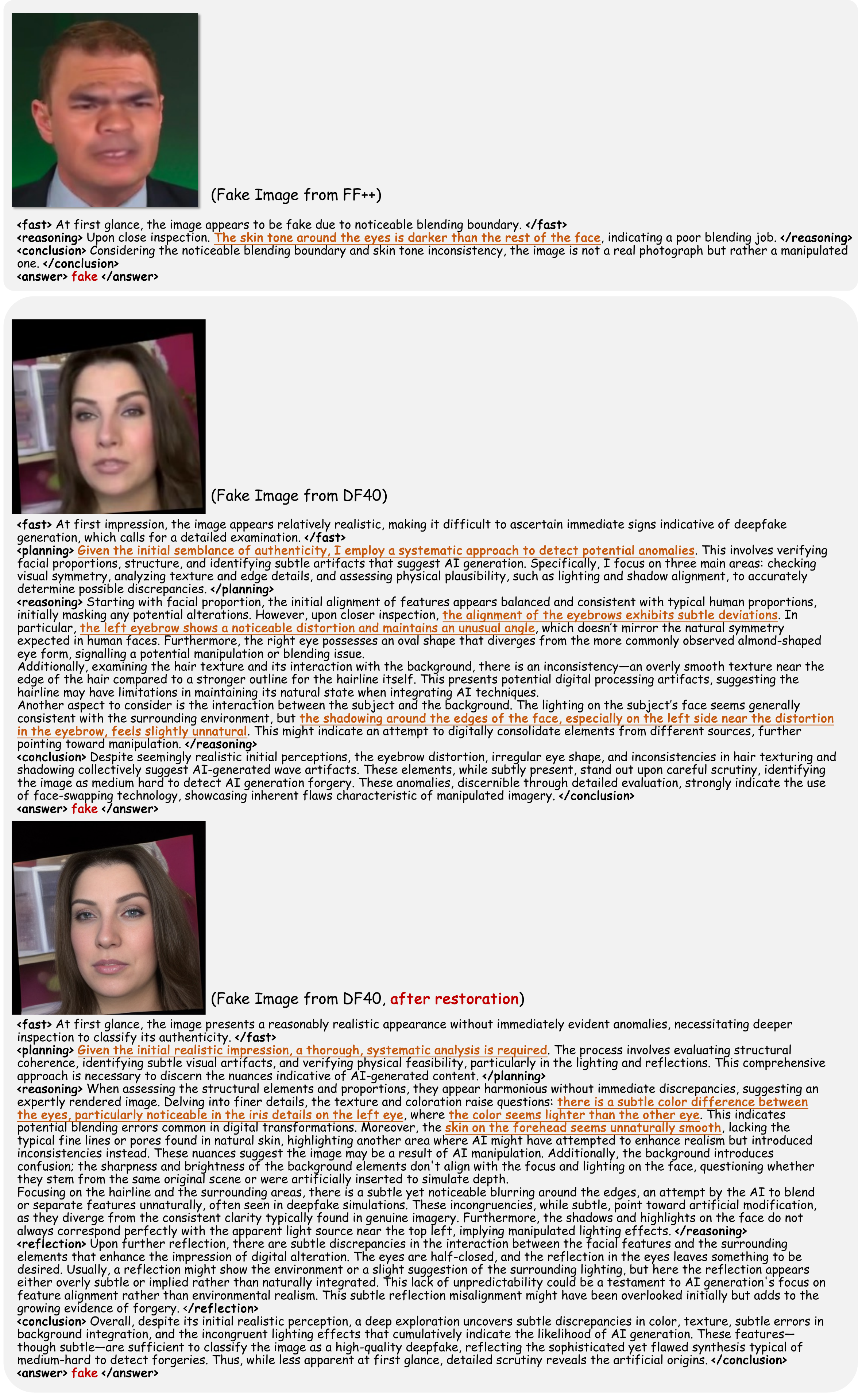}
     \caption{Reasoning output of our \veritas$\,$ model.
     }
	\label{fig:case1}
    \vspace{-0.2cm}
\end{figure*}

\begin{figure*}[t]
    \centering
    \includegraphics[width=0.98\linewidth]{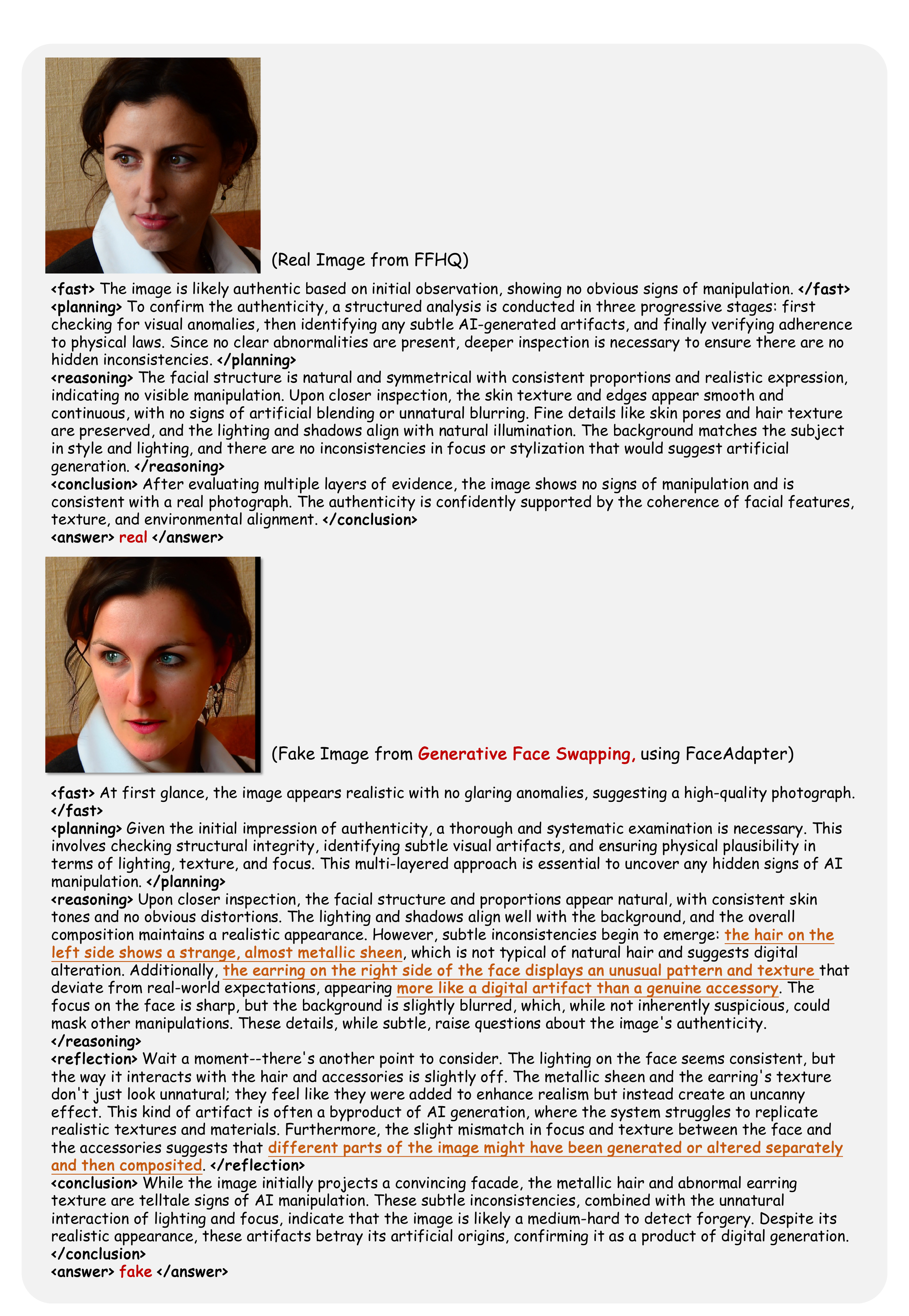}
     \caption{Reasoning output of our \veritas$\,$ model.
     }
	\label{fig:case2}
    \vspace{-0.2cm}
\end{figure*}

\begin{figure*}[t]
    \centering
    \includegraphics[width=0.98\linewidth]{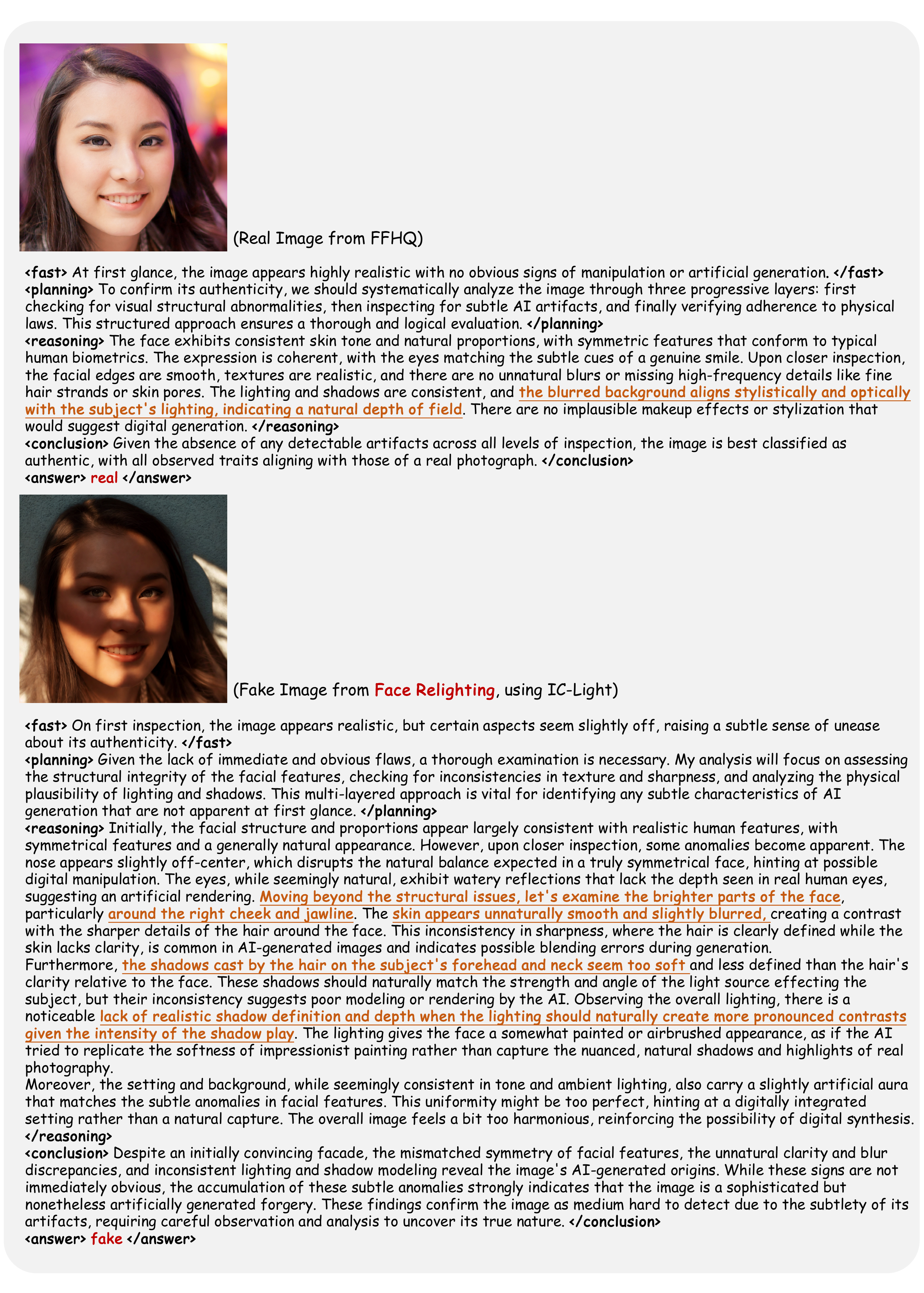}
     \caption{Reasoning output of our \veritas$\,$ model.
     }
	\label{fig:case3}
    \vspace{-0.2cm}
\end{figure*}

\begin{figure*}[t]
    \centering
    \includegraphics[width=0.98\linewidth]{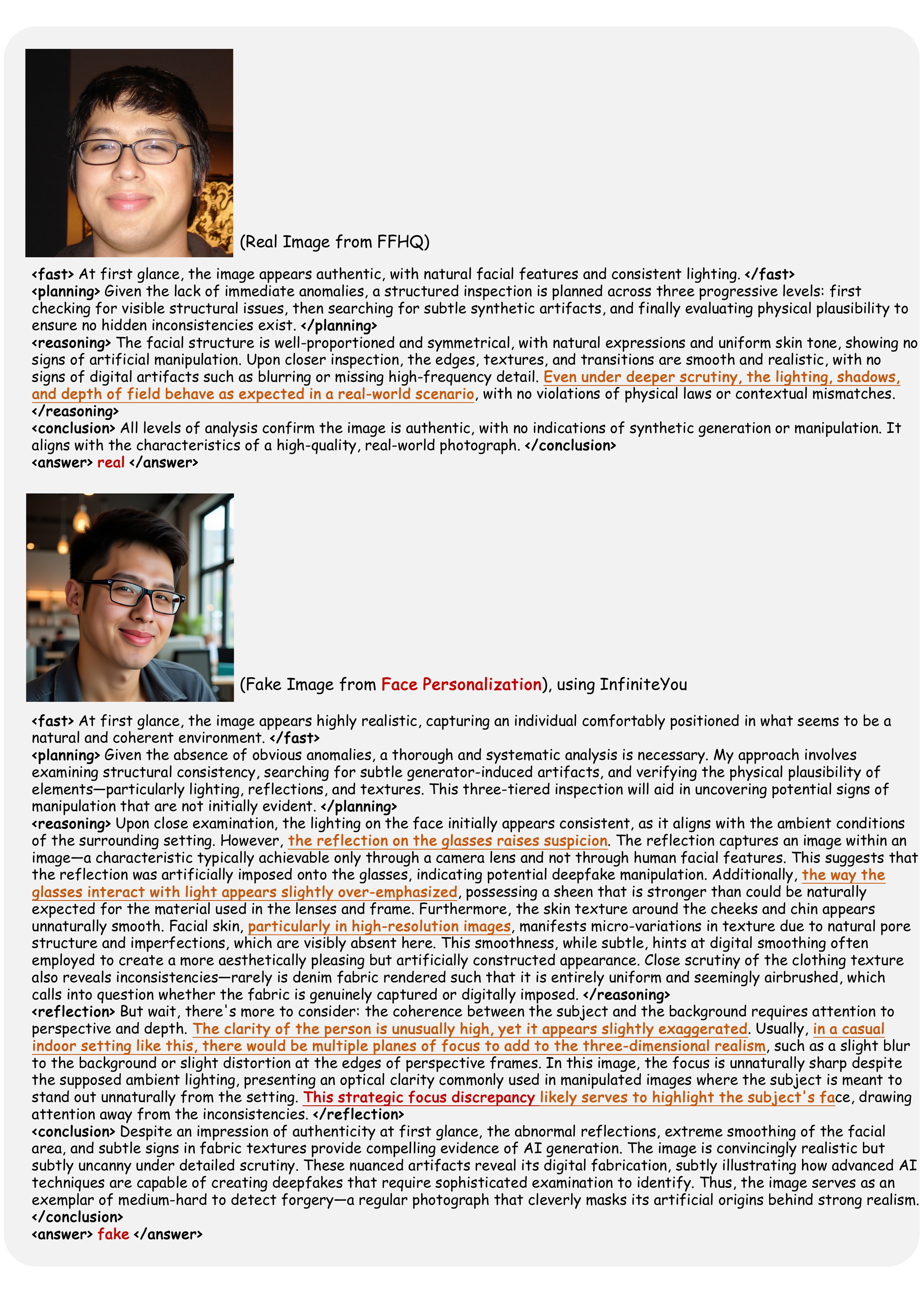}
     \caption{Reasoning output of our \veritas$\,$ model.
     }
	\label{fig:case4}
    \vspace{-0.2cm}
\end{figure*}

% rebuttal

\begin{figure*}[t]
    \centering
    \includegraphics[width=0.96\linewidth]{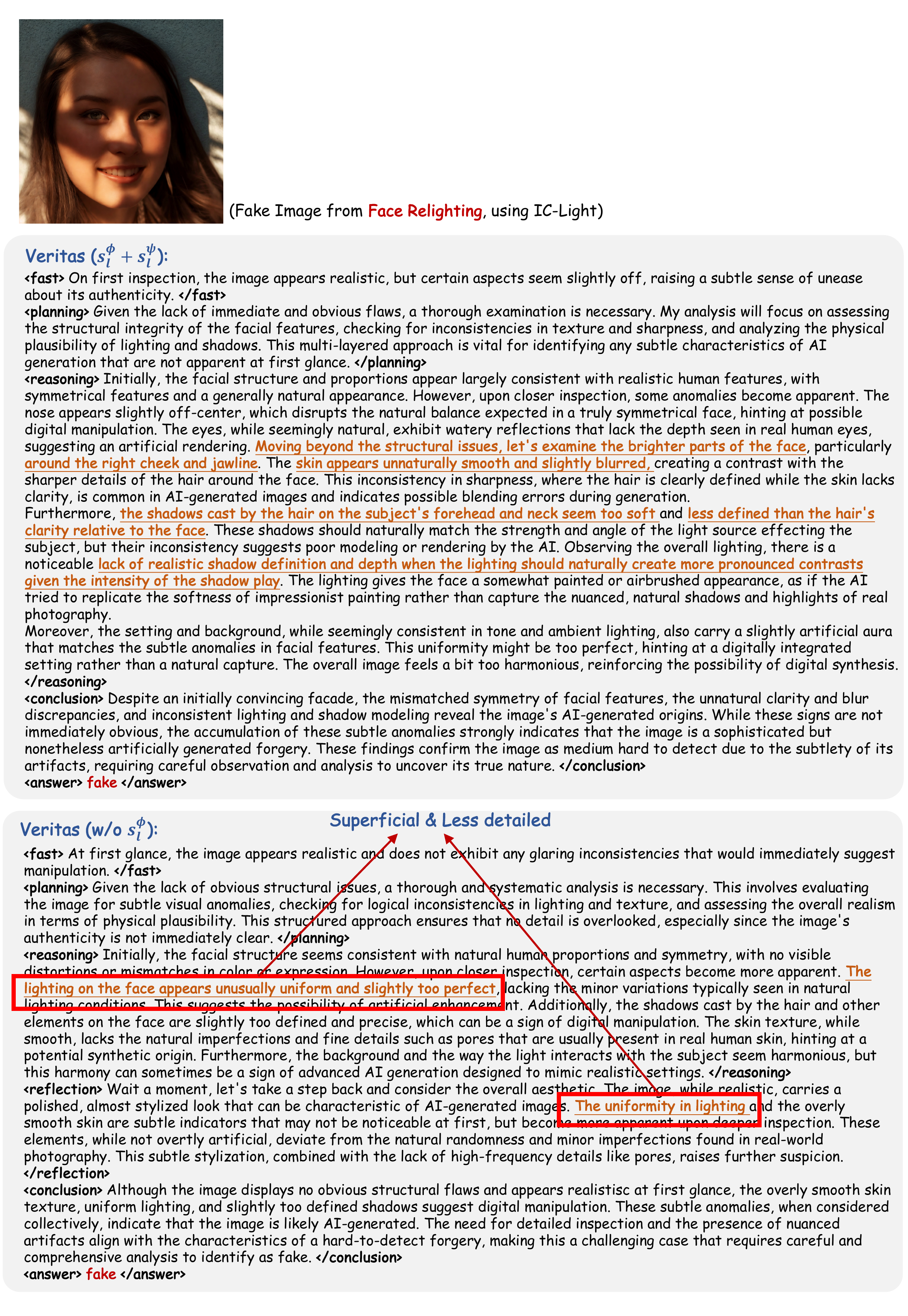}
     \caption{Qualitative comparisons of the effects of non-preference $\bm{s}_l^{\phi}$ in MiPO.
     }
	\label{fig:mipo_comp}
    \vspace{-0.2cm}
\end{figure*}

\begin{figure*}[ht]
\centering
\begin{tcolorbox}[listing only, 
                  listing style={basicstyle=\ttfamily\footnotesize},
                  colback=gray!10, 
                  title=Pattern-aware SFT Data Annotation (Stage-1 Prompt for Fake Images)]
\verb|{real image}| \verb|{fake image}|\\
You are given two images:\\
1. A real face image captured by real-world camera.\\
2. A fake face image generated by AI, using \verb|{Forgery Type and Explanation}|.
\\
\\
Please compare them visually and select \textbf{two most noticeable abnormalities} that best indicate the second image is fake.
\\
\\
Choose from the following list:\\
\textbf{1. Color Difference} \\
Inconsistent skin tone between facial regions (e.g., skin tone mismatches in face-swapped regions).\\
\textbf{2. Structure Abnormal}\\
The structures of facial features (e.g., eyes, nose, eyebrows, mouth) are distorted or asymmetric, which do not conform to the biometric patterns found in real human faces.\\
\textbf{3. Edge Abnormal}\\
Artificially sharp or jagged facial contours/edges, which are inconsistent with natural soft transitions.\\
\textbf{4. Expression Abnormal}\\
Unnatural or distorted facial expression which are not common in real human faces (e.g., mismatched smile-eye movements).\\
\textbf{5. Facial Proportion Abnormal}\\
Unusual proportion of facial components (e.g., abnormally wide eye distance or overly large forehead).\\
\textbf{6. Texture Abnormal}\\
Overly smooth/rough or discontinuous skin textures.\\
\textbf{7. Blending Boundary}\\
In blending techniques, the facial skin exhibits hard transitions or obvious block boundary.\\
\textbf{8. Unnatural Blur}\\
Abnormal local/block blurriness on facial regions or facial contour.\\
\textbf{9. High-Frequency Detail Missing}\\
Although the image is in high-resolution and looks realistic, some human biological details (e.g., pores and fine hair) are missing.\\
\textbf{10. Non-physically Plausible Lighting and shadow}\\
Inconsistent light/shadow directions across facial regions or light/shadow that violate physical illumination laws.\\
\textbf{11. Contextual Element Mismatch}\\
Incongruous background or mismatched style/lighting between face and background.\\
\textbf{12. Makeup Implausibility}\\
Conflicting specular highlights and makeup distribution (e.g., highlight on non-glossy skin areas).\\
\textbf{13. Holistic Stylization}\\
In advanced entire face generation, the image looks realistic, but is stylized overall (e.g., hyper-realistic digital art).\\
\textbf{14. Abnormal Optical Focus Discrepancy}\\
Entirely generated faces sometimes fail to simulate natural camera optics, exhibiting simultaneous sharpness in geometrically incompatible depth planes or abrupt defocus transitions violating optical distance gradients.\\
\\
Return only your selected clues, exactly as listed, separated by `` , ''. \\
Please strictly follow the format.
\end{tcolorbox}

\caption{Prompt for fake images SFT data annotation (Stage 1: anomalies identification).}
\label{prompt:s1}
\end{figure*}

\begin{figure*}[ht]
\centering
\begin{tcolorbox}[listing only, 
                  listing options={basicstyle=\ttfamily\footnotesize},
                  colback=gray!10, 
                  title=Pattern-aware SFT Data Annotation (Stage-2 Prompt for Fake Images)]
\verb|{image}|\\
\#\#\# Background Information\\
A fake face image generated by AI has multiple possible artifacts. Generally, these can be divided into three categories.\\
1. \textbf{Visual Structural Abnormalities}: Obvious abnormalities that can be clearly perceived visually, typically including the following types:\\
    \mbox{}\hspace{2em}1.1 Color Difference\\
    \mbox{}\hspace{2em}Inconsistent skin tone between facial regions (e.g., skin tone mismatches in face-swapped regions).\\
    \mbox{}\hspace{2em}...\\
2. \textbf{Subtle Artifacts from Generator}: Subtle artifacts introduced by generators that require more careful observation, typically including the following types:\\
    \mbox{}\hspace{2em}2.1 Edge Abnormal\\
    \mbox{}\hspace{2em}Artificially sharp or jagged facial contours/edges, which are inconsistent with natural soft transitions.\\
    \mbox{}\hspace{2em}...\\
3. \textbf{Violation of Physical Laws}: Implicit artifacts that require deeper observation and thinking, connecting visual clues with common knowledge. This typically includes the following types:\\
    \mbox{}\hspace{2em}3.1 Non-physically Plausible Lighting and shadow\\
    \mbox{}\hspace{2em}Inconsistent light/shadow directions across facial regions or light/shadow that violate physical illumination laws.\\
    \mbox{}\hspace{2em}...\\
\\
\#\#\# Preliminary Observation\\
The given image is a fake image generated by \verb|{Forgery Type and Explanation}|.\\
After careful inspection, we conclude two most noticeable abnormalities that indicate the image is fake:\\
\verb|{Stage-1 Results}|\\
\\
\#\#\# Task Definition\\
Your task is to examine the given image, then:\\
1. Give your initial judgement of the authenticity of the image:
\begin{itemize}
    \item Do not use any priori information provided above 
    \item If you think it is hard to make a judgment, you can point this out faithfully 
\end{itemize}
2. Extract \textbf{meticulous visual facts} that mainly conform to above two abnormalities. Specifically:
\begin{itemize}
    \item Take a careful examination of the given image.
    \item Perform step-by-step forensics analysis according to the above three progressive categories.
\end{itemize}
3. Draw a comprehensive conclusion based on your findings.\\
\\
Keep your answer detailed and factual.
    
\end{tcolorbox}

\caption{Prompt for fake images SFT data annotation (Stage 2: visual facts forensics). 
The omitted parts in ``Background Information'' are consistent with the artifacts list in Figure~\ref{prompt:s1}.}
\label{prompt:s2}
\end{figure*}

\begin{figure*}[ht]
\centering
\begin{tcolorbox}[listing only, 
                  listing options={basicstyle=\ttfamily\footnotesize},
                  colback=gray!10, 
                  title=Pattern-aware SFT Data Annotation (Stage-3 Prompt for Fake Images)]
Your task is to convert the given information into \textbf{logical chain-of-thought (CoT)}. 
The length and complexity should be conditioned on the given information.\\
\\
\#\#\# Extracted Evidence\\
The following information is the explanation to the artifacts of a fake image generated by AI.
Specifically, we partition the explanation into three parts:\\
1. \textbf{Initial Judgement}\\
We have required previous model truthfully give the judgment at a first glance. If the image has obvious artifacts, it will generate certain judgment. Otherwise, it will need further inspection.\\
The initial judgement of current sample is as follows:\\
\verb|{Initial Judgement from Stage-2}|
\\
2. \textbf{Detailed Evidence}\\
We cluster the possible artifacts into three progressive groups and then require previous model to conduct point-by-point forensic analysis. The extracted evidence of current sample is as follows:\\
\verb|{Forensics Analysis from Stage-2}|
\\
3. \textbf{Conclusion}\\
Generally, for some ambiguous samples, we need to make a comprehensive judgement based on different aspects. Therefore, we require previous model to draw a comprehensive conclusion based on the extracted evidence.\\
The conclusion of current sample is as follows:\\
\verb|{Conclusion from Stage-2}|
\\

\#\#\# Task Definition\\
Your task is to convert the given information into logical Chain-of-Thought (CoT).\\
You are not given the image, and you should keep faithful to the above information.\\
You can \textbf{flexibly} decide whether to perform Long-CoT or Short-CoT, based on the sample's difficulty.
    \begin{itemize}
        \item Long-CoT: Hard samples need to generate comprehensive and \textbf{logical} reasoning content. Often follow a \textbf{structured pattern}: Fast Judgement (\verb|<fast>|) - Problem Planning (\verb|<planning>|) - Evidence Collection (\verb|<reasoning>|) - Conclusion (\verb|<conclusion>|)
        \item Short-CoT: Medium and easy samples need to generate brief yet \textbf{critical} reasoning content. Often follow a \textbf{structured pattern}: Fast Judgement (\verb|<fast>|) - Evidence Collection (\verb|<reasoning>|) - Conclusion (\verb|<conclusion>|)
    \end{itemize}
The following guidance is only useful when you need it:
    \begin{itemize}
        \item For ``Problem Planning'', you should analyze the current state and draw a progressive and reasonable plan
        \item For ``Evidence Collection'', you should convert the given evidence into logical and coherent content. Do not mechanically perform step-by-step analysis using conjunctions like ``first'' and ``next''. Instead, make your reasoning smooth and natural
        \item If you suppose the current sample is extremely hard, you can insert ``Self-Reflection'' pattern before ``Conclusion'':
            You can smartly move some hard-to-detect artifacts from ``Evidence Collection'' into this part. Use natural conjunctions like ``However'', ``But wait'', etc. Enclose in \verb|<reflection>| tags.
            The reflective content is not a ``restatement'' but discovering something new that you have not considered before, which should be coherent with your previous reasoning content
        \item For ``Conclusion'', you should draw a comprehensive conclusion finally
    \end{itemize}
\end{tcolorbox}

\caption{Prompt for fake images SFT data annotation (Stage 3: thinking patterns injection).}
\label{prompt:s3}
\end{figure*}

\begin{figure*}[ht]
\vspace{-12pt}
\centering
\begin{tcolorbox}[listing only, 
                  listing style={basicstyle=\ttfamily\footnotesize},
                  colback=gray!10, 
                  title=Pattern-aware SFT Data Annotation (Stage-1 Prompt for Real Images)]
\verb|{image}|\\
\#\#\# Background Information\\
The authentic images can be manipulated by AI for improper use.\\
A fake face image generated by AI may suffer from some of the following artifacts. Generally, these can be divided into three categories.\\
1. \textbf{Visual Structural Abnormalities}: Obvious abnormalities that can be clearly perceived visually, typically including the following types:\\
    \mbox{}\hspace{2em}1.1 Color Difference\\
    \mbox{}\hspace{2em}Inconsistent skin tone between facial regions (e.g., skin tone mismatches in face-swapped regions).\\
    \mbox{}\hspace{2em}...\\
2. \textbf{Subtle Artifacts from Generator}: Subtle artifacts introduced by generators that require more careful observation, typically including the following types:\\
    \mbox{}\hspace{2em}2.1 Edge Abnormal\\
    \mbox{}\hspace{2em}Artificially sharp or jagged facial contours/edges, which are inconsistent with natural soft transitions.\\
    \mbox{}\hspace{2em}...\\
3. \textbf{Violation of Physical Laws}: Implicit artifacts that require deeper observation and thinking, connecting visual clues with common knowledge. This typically includes the following types:\\
    \mbox{}\hspace{2em}3.1 Non-physically Plausible Lighting and shadow\\
    \mbox{}\hspace{2em}Inconsistent light/shadow directions across facial regions or light/shadow that violate physical illumination laws.\\
    \mbox{}\hspace{2em}...\\
\\
\#\#\# Preliminary Observation\\
The given image is an authentic image captured by real-world camera.\\
Besides, we divide the authentic samples into three difficulty levels: easy, medium and difficult.
In general, easy samples have high visual clarity and are extremely realistic.\\
Samples of medium difficulty are in lower quality, but there are still strong evidence indicating its authenticity.\\
High-difficulty samples are in low quality and may contain some misleading artifacts, requiring careful thinking and comprehensive judgment.\\
The currently given image is considered as \verb|{difficulty}|.\\
\\
\#\#\# Task Definition\\
Your task is to examine the given image, then:\\
1. Give your initial judgement of the authenticity of the image:
\begin{itemize}
    \item Do not use any priori information provided above
    \item If you think it is hard to make a judgement, you can point this out faithfully
\end{itemize}
2. Provide an explanation that can distinguish the given real image from fakeness. Specifically:
\begin{itemize}
    \item Take a careful examination of the given image
    \item Perform step-by-step reasoning according to the above three progressive categories
    \item If there exits some \textbf{misleading artifacts} in the given image, you can point them out truthfully
\end{itemize}
3. Draw a comprehensive conclusion based on your reasoning\\
Keep your answer detailed and factual.
\end{tcolorbox}

\caption{Prompt for real images SFT data annotation (Stage 1: visual facts forensics). We divide the difficulty of real images upon datasets. FFHQ and CelebAHQ are considered as simple for their clear visual details. FaceForensics++ and CelebA are classified as medium for their miss of visual details. LFW is considered as hard for its low resolutions and unexpected noises.
The omitted parts in ``Background Information'' are consistent with the artifacts list in Figure~\ref{prompt:s1}.}
\label{prompt:real_s1}
\end{figure*}

\begin{figure*}[ht]
\vspace{-30pt}
\centering
\begin{tcolorbox}[listing only, 
                  listing style={basicstyle=\ttfamily\footnotesize},
                  colback=gray!10, 
                  title=Pattern-aware SFT Data Annotation (Stage-2 Prompt for Real Images)]
\verb|{image}|\\
Your task is to convert the given information into \textbf{logical chain-of-thought (CoT)}. 
The length and complexity of the reasoning chain should be conditioned on the given information.\\
\\
\#\#\# Extracted Evidence\\
The following information is a detailed explanation that distinguishes a given real image from fakeness.\\
Specifically, we partition the explanation into three parts:\\
1. \textbf{Initial Judgement}\\
We have required previous model truthfully give the judgement at a first glance. If the image has obvious artifacts, it will generate certain judgement. Otherwise, it will need further inspection.
The initial judgement of current sample is as follows:
\verb|{Initial Judgement from Stage-1}|\\
2. \textbf{Detailed Evidence}\\
We cluster the possible artifacts into three **progressive** groups and then require previous model to conduct point-by-point forensic analysis. The extracted evidence of current sample is as follows:\\
\verb|{Forensics Analysis from Stage-1}|\\
3. \textbf{Conclusion}
Generally, for some ambiguous samples, we need to make a comprehensive judgement based on different aspects. Therefore, we require previous model to draw a comprehensive conclusion based on the extracted evidence.
The conclusion of current sample is as follows:
\verb|{Conclusion from Stage-1}|\\
\\
\#\#\# Difficulty Information\\
To enable better control of the length of the reasoning chain, we divide the authentic samples into three difficulty levels: easy, medium and hard.\\
In general, easy samples have high visual clarity and are extremely realistic.
Medium samples are in lower quality, but there are still strong evidence indicating its authenticity.
Hard samples are in extremely low quality and may contain some misleading artifacts, requiring careful thinking and comprehensive judgment.
The currently given image is roughly classified as \verb|{difficulty}|.\\
\\
\#\#\# Task Definition\\
You are not given the image, and you should keep faithful to the above information.
You can \textbf{flexibly} decide whether to perform Long-CoT or Short-CoT, based on the sample's difficulty.
    \begin{itemize}
        \item Long-CoT: Hard samples need to generate comprehensive and \textbf{logical} reasoning content. Often follow a \textbf{structured pattern}: Fast Judgement (\verb|<fast>|) - Problem Planning (\verb|<planning>|) - Evidence Collection (\verb|<reasoning>|) - Conclusion (\verb|<conclusion>|)
        \item Short-CoT: Medium and easy samples need to generate brief yet \textbf{critical} reasoning content. Often follow a \textbf{structured pattern}: Fast Judgement (\verb|<fast>|) - Evidence Collection (\verb|<reasoning>|) - Conclusion (\verb|<conclusion>|)
    \end{itemize}
The following guidance is only useful when you need it:
    \begin{itemize}
        \item For ``Problem Planning'', you should analyze the current state and draw a progressive and reasonable plan
        \item For ``Evidence Collection'', you should convert the given evidence into logical and coherent content. Do not mechanically perform step-by-step analysis using conjunctions like ``first'' and ``next''. Instead, make your reasoning smooth and natural
        \item If there are any misleading artifacts in the provided information, should put them into ``Reflection'' pattern before ``Conclusion'', using natural conjunctions like ``However'', ``Although'', etc. Enclose in \verb|<reflection>| tags. ONLY insert ``Reflection'' pattern when there are \textbf{known misleading artifacts}.
        \item For ``Conclusion'', you should draw a comprehensive conclusion finally
    \end{itemize}
\end{tcolorbox}
\caption{Prompt for real images SFT data annotation (Stage 2: thinking patterns injection).}
\label{prompt:real_s2}
\end{figure*}

\begin{figure*}[ht]
\centering
\begin{tcolorbox}[listing only, 
                  listing options={basicstyle=\ttfamily\footnotesize},
                  colback=gray!10, 
                  title=Reasoning quality (Score evaluation)]
You are a helpful assistant proficient in analyzing vision reasoning problems.\\
\\
\#\# Instruction:
Please examine the provided image attentively and serve as an unbiased judge in assessing the quality of the response from an AI assistants regarding the instruction. You will receive a single response from the assistant to user's instruction.\\
\\
\#\# Noticement:
Your assessment should identify whether the assistant effectively adheres to the user's instructions and addresses the user's inquiry.\\
In your evaluation, weigh factors such as preciseness, comprehensiveness, clarity, creativity, and the granularity of the responses.\\
Do not allow the length of the responses to influence your evaluation.\\
Do not favor certain names or positions of the assistants. Be as objective as possible.\\
\\
\#\# Criteria:\\
Use scores to show the quality of the response. Here is the detailed scoring rubric for evaluating the quality of responses from AI assistants:\\
\textbf{Incorrect (1)}: The answer is incorrect (e.g., predicting real for the fake image), the score should be 1.\\
\textbf{Poor (2)}: The response fails to address the query effectively. Although the answer is correct, the reasoning process is vague and repetitive. Creativity and granularity are absent or poorly executed.\\
\textbf{Average (3)}: The response shows superficial analysis of the given image. It lacks depth in creativity and granularity. The reasoning process is not detailed.\\
\textbf{Good (4)}: The response is well-aligned with the given image, demonstrating a high degree of preciseness and comprehensiveness. It shows creativity and a nuanced understanding of the topic, with a detailed granularity that enhances the response quality.\\
\textbf{Excellent (5)}: The response shows great preciseness, comprehensiveness, creativity, and granularity. It provides an insightful, detailed, and thorough answer, indicating a deep and nuanced understanding of the user's inquiry.\\
\\
\#\# Desired Output Format:\\
Present your verdict in a JSON format, with the key 'analysis' for a short reason of your judgement and the key 'judgment' to indicate your decision: use "[[1]]", "[[2]]", "[[3]]", "[[4]]", "[[5]]" to indicate your evaluate score.\\
\\
\#\# Question:\\
\verb|{image}| This is a \verb|{Ground Truth}| image. The following are the user's query and the model's output.\\
\lbrack The Start of User Instruction\rbrack\\
Please determine the authenticity of this image.\\
\lbrack The End of User Instruction\rbrack\\
\lbrack The Start of Assistant's Answer\rbrack\\
\verb|{Model's Reasoning Output}|\\
\lbrack The End of Assistant's Answer\rbrack
\end{tcolorbox}

\caption{Prompt for reasoning quality evaluation (Score evaluation).}
\label{prompt:score}
\end{figure*}

\begin{figure*}[ht]
\centering
\begin{tcolorbox}[listing only, 
                  listing options={basicstyle=\ttfamily\footnotesize},
                  colback=gray!10, 
                  title=Reasoning quality (Pairwise evaluation)]
You are a helpful assistant proficient in analyzing vision reasoning problems.\\
\\
\#\# Instruction:\\
Please examine the provided image attentively and serve as an unbiased judge in assessing the quality of responses from two AI assistants regarding the user's question shown beneath the image.\\
\\
\#\# Noticement:\\
Your assessment should identify the assistant that more effectively adheres to the user's instruction and provides more detailed, more precise and high-quality reasoning.\\
In your evaluation, weigh factors such as preciseness, comprehensiveness, clarity, creativity, and the granularity of the responses.\\
Avoid any position biases and ensure that the order in which the responses were presented does not influence your decision.\\
Do not allow the length of the responses to influence your evaluation.\\
Do not favor certain names of the assistants. Be as objective as possible.\\
\\
\#\# Desired Output Format:\\
Present your verdict in a JSON format, with the key 'analysis' for a short reason of your judgement and the key 'judgment' to indicate your decision: use "[[A]]" if assistant A prevails, "[[B]]" if assistant B does, and "[[C]]" for a tie.\\
\\
\#\# Question:\\
\verb|{image}| This is a \verb|{Ground Truth}| image. The following are the user's query and the model's output.\\
\lbrack The Start of User Instruction\rbrack\\
Please determine the authenticity of this image.\\
\lbrack The End of User Instruction\rbrack\\
\lbrack The Start of Assistant A's Answer\rbrack\\
\verb|{Model A's Reasoning Output}|\\
\lbrack The End of Assistant A's Answer\rbrack\\
\lbrack The Start of Assistant B's Answer\rbrack\\
\verb|{Model B's Reasoning Output}|\\
\lbrack The End of Assistant B's Answer\rbrack
\end{tcolorbox}

\caption{Prompt for reasoning quality evaluation (Pairwise evaluation).}
\label{prompt:pair}
\end{figure*}

\begin{figure*}[ht]
\centering
\begin{tcolorbox}[listing only, 
                  listing options={basicstyle=\ttfamily\footnotesize},
                  colback=gray!10, 
                  title=Generation of Personalization Prompts]
\verb|{image}|\\
Your task is to create customized prompts for the input image to fool deepfake detectors. \\
\\
\#\#\# Requirements\\
1. Tailor the prompt based on the specific input image. Change the context. For example, ``Old man with beard'', ``a chef in a bustling kitchen, exuding expertise and dedication'', ``beautiful bride, traditional, attire, floral braid, sequin headdress, orchid backdrop, pastels'', etc.\\
2. Keep the prompt concise and effective.\\
3. Avoid using any obscure words.
\end{tcolorbox}

\caption{Input for generating customized personalization prompts.}
\label{prompt:personalization}
\end{figure*}

\begin{figure*}[ht]
\centering
\begin{tcolorbox}[listing only, 
                  listing options={basicstyle=\ttfamily\footnotesize},
                  colback=gray!10, 
                  title=Prompt for Reflection Quality Reward Model]
\verb|{image}|\\
You are provided with an image and a question for this image. The provided response is the reasoning process of determining its authenticity.\\
You should re-examine the image carefully, and then review the self-reflection content enclosed in the \verb|<reflection>| \verb|</reflection>| tags:\\
1. Is the reflection content redundant with the reasoning content? Is the reflection content just a restatement or conclusion of previous reasoning? The reflection should introduce new insights rather than restatement.\\
2. The reflection should not be vague statements such as "too perfect" or "lack of imperfections". Instead, it should be specific and detailed.\\
\\
From 0 to 100, how do you rate for the reflection quality?\\
Be strict, give low score if it is not aligned with the above principles.\\
Provide a few lines for explanation and the rate number at last after "Final Score:".\\
\\
Your task is provided as follows:\\
\\
Question: \lbrack \verb|{Question}|\rbrack \\
Response: \lbrack \verb|{Reasoning Output}|\rbrack
\end{tcolorbox}

\caption{Prompt for Reflection Quality Reward model (UnifiedReward-Qwen-3B).}
\label{prompt:reward}
\end{figure*}

\begin{figure*}[ht]
\centering
\begin{tcolorbox}[listing only, 
                  listing options={basicstyle=\ttfamily\footnotesize},
                  colback=gray!10, 
                  title=Input Prompt for Veritas (All stages)]
\#\#\# \verb|System|:\\
You are an image authenticity expert. Your task is to determine the authenticity of the given facial image.\\
\\
Firstly, give an overall judgement to the authenticity of the image, enclosed in \verb|<fast>| \verb|</fast>| tags.\\
Then, make a careful and structured thinking before reaching an answer. Based on your thinking, draw a comprehensive conclusion. Enclose the corresponding part in different tags, e.g., \verb|<planning>| or \verb|<reasoning>| or \verb|<reflection>| or \verb|<conclusion>|.\\
Finally, give the final answer with ``real'' or ``fake'', enclosed in \verb|<answer>| \verb|</answer>| tags.\\
\\
\#\#\# \verb|User|:\\
\verb|{image}| Please determine the authenticity of this image.
\end{tcolorbox}

\caption{Input prompt for Veritas. The prompts for all training stages are consistent.}
\label{prompt:sys_veritas}
\end{figure*}

\begin{figure*}[ht]
\centering
\begin{tcolorbox}[listing only, 
                  listing options={basicstyle=\ttfamily\footnotesize},
                  colback=gray!10, 
                  title=Input Prompt for Qwen2.5-VL-7B InternVL3-8B and GLM-4.1V-9B-Thinking]
\#\#\# \verb|System|:\\
You are given an facial image. Please analyze the provided facial image and determine whether it is authentic or fake based on the following classification criteria:\\
Real Captured Facial image\\
    - Images captured using a real camera or device without any alternations or manipulation.\\   
Fake Facial Image\\
    - Images generated or manipulated using digital technologies, such as deepfakes, face swapping, face reenactment, photo editing software, entire face synthesis, etc.\\
Output the thinking process in \verb|{<think>}| \verb|{</think>}| and final answer (``real'' or ``fake'') in \verb|{<answer>}| \verb|{</answer>}| tags, i.e., the output answer format should be as follows:\\
\verb|{<thinking>}| your thinking process here \verb|{</thinking>}| \verb|{<answer>}| your judgement here \verb|{</answer>}|
Please strictly follow the format.\\
\\
\#\#\# \verb|User|:\\
\verb|{image}| Please determine the authenticity of this image.
\end{tcolorbox}

\caption{Input prompt for Qwen2.5-VL-7B, InternVL3-8B and GLM-4.1V-9B-Thinking.}
\label{prompt:zs_qwen}
\end{figure*}

\begin{figure*}[ht]
\centering
\begin{tcolorbox}[listing only, 
                  listing options={basicstyle=\ttfamily\footnotesize},
                  colback=gray!10, 
                  title=Input Prompt for MiMo-VL-7B]
\#\#\# \verb|System|:\\
You are Qwen, created by Alibaba Cloud. You are a helpful assistant.\\
\\
\#\#\# \verb|User|:\\
\verb|{image}| Please determine the authenticity of this image. Output your final answer (``real'' or ``fake'') in \verb|<answer>| \verb|</answer>| tags
\end{tcolorbox}

\caption{Input prompt for MiMo-VL-7B. We found that providing priori knowledge does no good for MiMo-VL-7B, hence we remove any priori in system prompt.}
\label{prompt:zs_mimo}
\end{figure*}

\begin{figure*}[ht]
\centering
\begin{tcolorbox}[listing only, 
                  listing options={basicstyle=\ttfamily\footnotesize},
                  colback=gray!10, 
                  title=Input Prompt for GPT-4o and Gemini-2.5-Pro]
\#\#\# \verb|System|:\\
You are an image authenticity expert. Your task is to determine the authenticity of the given facial image.\\
\\
Firstly, give an overall judgement to the authenticity of the image, enclosed in \verb|<fast>| \verb|</fast>| tags.\\
Then, make a careful and structured thinking before reaching an answer. Based on your thinking, draw a comprehensive conclusion. Enclose the corresponding part in different tags, e.g., \verb|<planning>| or \verb|<reasoning>| or \verb|<reflection>| or \verb|<conclusion>|.\\
Finally, give the final answer with ``real'' or ``fake'', enclosed in \verb|<answer>| \verb|</answer>| tags.\\
\\
\#\#\# \verb|User|:\\
\verb|{image}| Please determine the authenticity of this image.
\end{tcolorbox}

\caption{Input prompt for GPT-4o and Gemini-2.5-Pro. Similar to MiMo-VL-7B, we found that providing priori knowledge is not helpful. We keep the default system prompt and only customize user prompt by constraining the output format.}
\label{prompt:zs_gpt}
\end{figure*}

% 1. R2-Q2, 更多benchmark实验：表1 + aigc分析图1
% 2. R3-Q1：图2 failure case
% 3. R3-Q2: 关于patterns细粒度消融：表2
% *（4. R3-Q3: deepfake推理合理性，case分析图)
% 5. R3-Q6: 关于mipo细粒度消融：表3 + mipo case分析图3
% 6. R4: 与更多方法的比较：单独开一个表4 + 多方法hydrafake case比较图4
% 7. R4: AIGC扩展性。图1前述已有，  表5性能

\end{document}